%% file: main.tex
\documentclass[11pt]{article}
\usepackage{arxiv}
\input{preamble}

\renewcommand{\shorttitle}{Shared Encoding and Conditional Signed Readout in Vision and Language Models}
\title{How Do Deep Networks Encode and Selectively Use Information? Shared Encoding and Conditional Signed Readout in Vision and Language Models}

\author{Kaixiang Shu\\
{\normalfont Independent Researcher}\\
\texttt{614729197@qq.com}}
\date{}

\begin{document}
\maketitle

\begin{abstract}
How can trained deep networks retain contextual information across internal representations while using it selectively in a decision? We study this question through controlled background-texture edits in vision classifiers and entity-binding edits in frozen language models. Across both modalities, factor perturbations preserve recognizable structure across pre-fixed, mutually disjoint blocks of hidden units: visual channel blocks match their complements more closely for the same image than for other images under the same texture edit, while language blocks transfer a relational direction fitted under one query role to rank relational states within held-out name pairs under the other role. We then decompose the local output change into signed channel and pathway contributions. In a convolutional classifier, net channel-level reinforcement ($R^{(512)}>1$) coexists on individual images with cancellation between positive- and negative-weight groups ($R^{(2)}<1$). Holding weights, the paired relation edit, and red/blue answer words fixed, switching prompt presentation reverses the mean finite selection effect in Qwen2.5. Two held-out tests validate this mechanism. Attenuating development-selected antagonistic units amplifies edit effects on unseen inputs; in a vision transformer, attenuating the same frozen units in one forward pass increases the mean magnitude of the background-edit effect on one class margin and reduces it on another, with the opposite pattern for a second group (four-cell paired interaction $+78.30\times 10^{-5}$ logit). Finally, combining conditional channel variance with fixed downstream readout weights predicts single-channel pruning recovery across ten cuts. These results show how shared encoding and conditional signed readout can coexist in vision and language models.
\end{abstract}


\input{sections/01_introduction}
\input{sections/02_related_work}
\input{sections/03_framework}
\input{sections/04a_protocol_encoding_channels}
\input{sections/04b_paths_conditions}
\input{sections/04c_gating_pruning}
\input{sections/05_discussion}
\input{sections/06_conclusion}

\FloatBarrier
\bibliographystyle{plainnat}
\bibliography{references}

\FloatBarrier
\clearpage
\pdfbookmark[0]{Supplementary Material}{supplementary-material}
\renewcommand{\shorttitle}{Supplementary Material: Shared Encoding and Conditional Signed Readout}
\setcounter{section}{0}
\setcounter{figure}{0}
\setcounter{table}{0}
\setcounter{equation}{0}
\renewcommand{\thesection}{S\arabic{section}}
\renewcommand{\thesubsection}{\thesection.\arabic{subsection}}
\renewcommand{\thefigure}{S\arabic{figure}}
\renewcommand{\thetable}{S\arabic{table}}
\renewcommand{\theequation}{S\arabic{equation}}
\renewcommand{\theHsection}{supp.\arabic{section}}
\renewcommand{\theHsubsection}{supp.\arabic{section}.\arabic{subsection}}
\renewcommand{\theHfigure}{supp.\arabic{figure}}
\renewcommand{\theHtable}{supp.\arabic{table}}
\renewcommand{\theHequation}{supp.\arabic{equation}}
\captionsetup{font=small,labelfont=bf,skip=4pt}
\setlength{\floatsep}{7pt plus 2pt minus 2pt}
\setlength{\textfloatsep}{9pt plus 2pt minus 2pt}
\setlength{\intextsep}{8pt plus 2pt minus 2pt}
\raggedbottom

\begin{center}
  \rule{\linewidth}{1.6pt}\\[0.55em]
  {\LARGE\scshape Supplementary Material}\\[0.45em]
  \rule{\linewidth}{0.7pt}
\end{center}
\vspace{0.5em}

\noindent\emph{Overview.} Sections~\ref{app:notation}--\ref{app:thm4} provide the unified conflict-free notation table and complete mathematical proofs, derivations, and analytical models supporting \cref{sec:framework}. Sections~\ref{supp:s1}--\ref{supp:s9} document checkpoint provenance, cohort-reuse disclosures, numerical precision audits, and full extended tables and figures supporting \cref{sec:experiments,sec:discussion}. All analysis code, frozen confirmation summaries, and figure-generation scripts are publicly available at \url{https://github.com/k-Jayus/How-Do-Deep-Networks-Encode-and-Selectively-Use-Information-}.

\input{supplementary/S01_notation}
\input{supplementary/S02_thm1_proofs}
\input{supplementary/S03_thm3_commutation}
\input{supplementary/S04_gram_decomps}
\input{supplementary/S05_fwl_toy_audit}
\FloatBarrier
\input{supplementary/S06_to_S10_experiments}
\FloatBarrier
\input{supplementary/S11_to_S14_experiments}

\end{document}

%% file: preamble.tex
\usepackage[utf8]{inputenc}
\usepackage[T1]{fontenc}
\usepackage{amsmath,amssymb,amsthm,mathtools}
\usepackage{booktabs}
\usepackage{longtable,tabularx,array,multirow}
\usepackage[numbers,sort&compress]{natbib}
\usepackage{graphicx}
\usepackage{float}
\usepackage{caption}
\usepackage{microtype}
\usepackage{etoolbox}
\usepackage{enumitem}
\usepackage{placeins}
\usepackage{needspace}
\usepackage{xcolor}
\usepackage{xurl}
\usepackage{hyperref}
\usepackage[nameinlink,noabbrev,capitalise]{cleveref}

\AtBeginEnvironment{thebibliography}{\raggedright}
\renewcommand{\theHequation}{\thesection.\arabic{equation}}

\definecolor{paperInk}{HTML}{383C3A}
\definecolor{paperSlate}{HTML}{556B77}
\definecolor{paperMauve}{HTML}{7A6775}
\definecolor{paperSage}{HTML}{657660}
\definecolor{paperRose}{HTML}{9E5E54}
\definecolor{paperLightBg}{HTML}{F7F8F8}

\hypersetup{
  colorlinks=true,
  linkcolor=paperSlate,
  filecolor=paperSlate,
  citecolor=paperRose,
  urlcolor=paperSlate,
  pdftitle={How Do Deep Networks Encode and Selectively Use Information? Shared Encoding and Conditional Signed Readout in Vision and Language Models},
  pdfauthor={Kaixiang Shu}
}

\AtBeginDocument{\setlength{\headheight}{26pt}}

\setlist{itemsep=0.25em,topsep=0.35em}
\makeatletter
\patchcmd{\@maketitle}
  {\vskip 0.4in \@minus 0.1in \center{\@date}   \vskip 0.2in}
  {\ifx\@date\@empty\vskip .14in\else\vskip .2in\center{\@date}\vskip .14in\fi}{}{}
\makeatother

\pretocmd{\section}{\Needspace{7\baselineskip}}{}{}
\pretocmd{\subsection}{\Needspace{6\baselineskip}}{}{}
\pretocmd{\subsubsection}{\Needspace{5\baselineskip}}{}{}

\graphicspath{{figures/main/}{figures/supplement/}}

\theoremstyle{plain}
\newtheorem{axiom}{Axiom}
\newtheorem{theorem}{Theorem}

\theoremstyle{definition}
\newtheorem{definition}{Definition}
\newtheorem{example}{Example}

\theoremstyle{remark}

\newcommand{\Rbb}{\mathbb{R}}
\newcommand{\Ebb}{\mathbb{E}}
\newcommand{\Xcal}{\mathcal{X}}
\newcommand{\Hcal}{\mathcal{H}}
\newcommand{\Qcal}{\mathcal{Q}}
\newcommand{\Pcal}{\mathcal{P}}
\newcommand{\Gcal}{\mathcal{G}}
\newcommand{\Aop}{\mathcal{A}}
\newcommand{\Wop}{\mathcal{W}}
\newcommand{\Rint}{\mathcal{R}_{\mathrm{int}}}
\newcommand{\Eincoh}{E_{\mathrm{incoh}}}
\newcommand{\Hout}{\mathbf{H}_{\mathrm{out}}}
\newcommand{\Dcomm}{\Delta_{\mathrm{comm}}}
\newcommand{\TER}{\operatorname{TER}}
\newcommand{\EVR}{\operatorname{EVR}}

\newcommand{\diag}{\operatorname{diag}}
\newcommand{\sign}{\operatorname{sign}}
\newcommand{\inner}[2]{\left\langle #1,\,#2 \right\rangle}
\newcommand{\norm}[1]{\left\lVert #1 \right\rVert}
\newcommand{\abs}[1]{\left\lvert #1 \right\rvert}

%% file: sections/01_introduction.tex
\section{Introduction}
\label{sec:intro}

A central scientific question in the study of trained deep neural networks is how internal representations can broadly preserve input information while outputs remain highly selective for the current task. Conventional intuition often pictures inference as progressive upstream filtering: early layers extract rich structure, deeper layers discard task-irrelevant context, and the final layer reads off a clean, isolated target variable. Yet empirical probes and source-space visualizations across both vision and language repeatedly show that deep hidden states retain far more contextual and relational structure than any single output decision appears to use \citep{belinkov2022probing,ravichander2021probing,xiao2021noise,shu2026adjoint,shu2026shared}. How do vision and language models organize input information internally, and how do they turn broadly preserved representations into task-selective outputs?

Two concrete cross-modal observations sharpen this question into a testable mechanistic problem. In deep image classifiers, Local Adjoint Correctors (LACs) \citep{shu2026adjoint} revealed that channels carrying positive class-head weights and channels carrying negative class-head weights both retain extensive scene and background structure, yet their signed combination concentrates energy sharply onto the foreground object; follow-up closed-form source-grounded inversion with shared-axis residualization \citep{shu2026source} further located candidate region-sensitive channel subsets. In frozen autoregressive language models performing multi-entity relational binding \citep{shu2026shared,feng2024binding}, a transferable relational component remains identifiable across distinct query roles inside intermediate hidden representations, yet holding model weights and the relational edit fixed while changing the prompt presentation or candidate answer contrast modulates the edit's decision gain and can reverse its sign. Viewed side by side, these two findings pose a single architectural question spanning both halves of inference: how do deep networks \emph{encode} a factor across internal units, and how do downstream computations \emph{selectively use} it?

Two familiar abstractions in deep learning leave this coexistence underspecified. On the encoding side, the classical notion of a distributed representation \citep{rumelhart1986learning} permits different units to carry complementary puzzle fragments that become recognizable only when assembled jointly. What our visual discovery initially suggested as ``holographic'' encoding \citep{shu2026adjoint}---and what we operationalize natively across modalities as \emph{shared encoding}---is more specific: the response to a controlled input factor leaves recognizable structure across multiple pre-fixed, mutually disjoint blocks of coordinates. On the readout side, even when the same factor remains recognizable across disjoint blocks, selectivity need not rely on upstream erasure. Downstream channels and nonlinear pathways can combine signed contributions according to the current readout, allowing broadly retained information to be suppressed in one task, amplified in another, or inverted across query conditions.

\begin{figure}[t]
  \centering
  \includegraphics[width=\textwidth]{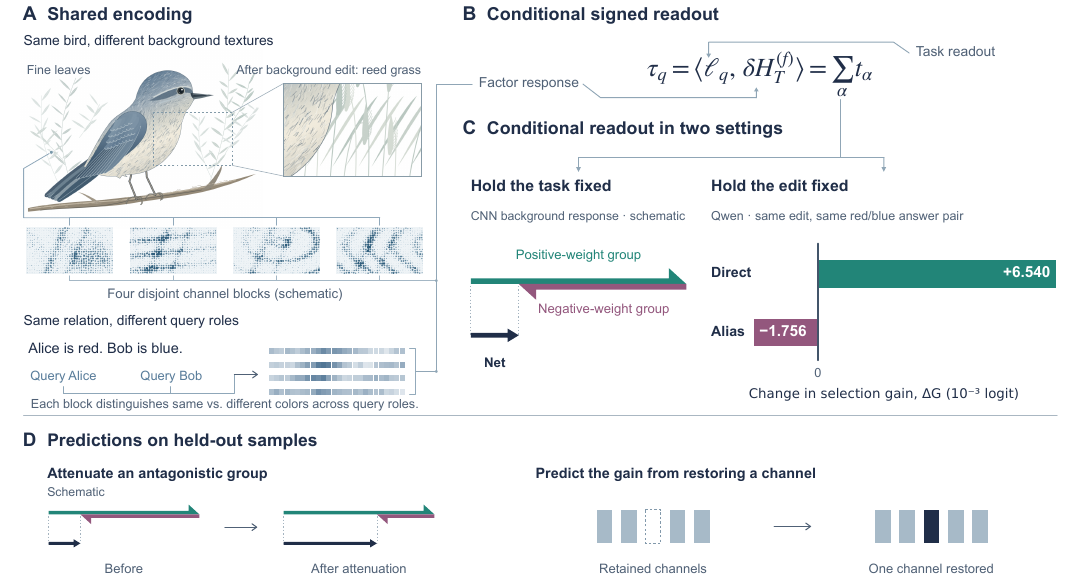}
  \caption{\textbf{Shared encoding and conditional signed readout.} \textbf{A,} Shared encoding across modalities: in vision (top), editing background texture on the same bird leaves recognizable structure across four pre-fixed, mutually disjoint channel blocks (schematic); in language (bottom), a multi-entity relational state remains distinguishable across four disjoint hidden-coordinate blocks under distinct query roles. \textbf{B,} At an internal cut $T$, the conditional signed-readout law $\tau_q = \langle \ell_q, \delta H_T^{(f)} \rangle = \sum_\alpha t_\alpha$ pairs the primal factor response $\delta H_T^{(f)}$ with the task-conditioned dual readout covector $\ell_q$. \textbf{C,} Two complementary readout settings probe the same law: holding the task fixed (left, CNN background schematic) shows positive-weight (teal) and negative-weight (plum) groups partially cancelling into a smaller net effect, whereas holding the edit and red/blue answer pair fixed in frozen Qwen2.5 (right) shows prompt switching reversing the measured selection gain from Direct ($+6.540$) to Alias ($-1.756 \times 10^{-3}$ logit). \textbf{D,} Two out-of-sample predictive tests validate the mechanism: attenuating a development-selected antagonistic group increases the net edit effect on held-out inputs (left, schematic), and combining conditional channel variance with fixed-readout coupling predicts the gain from restoring one omitted channel in structured pruning (right).}
  \label{fig:fig1}
\end{figure}

To formalize both stages on equal footing, we introduce a unified primal--dual framework for trained-network inference (\cref{fig:fig1}). At any internal cut $T$, an upstream map $f_{\mathrm{up}}$ encodes a controlled factor perturbation $\delta H_T^{(f)}$ in the primal representation space $\Hcal_T$, while a downstream map $f_{\mathrm{down}}(\cdot;\,x,q)$ defines a task-conditioned dual readout covector $\ell_q(x) = \nabla_{H_T} m_q(x) \in \Hcal_T^*$. Decomposing $\ell_q(x)$ across orthogonal unit partitions and tangent-space computational pathways yields a single local signed-readout law: the net directional effect is the signed sum $\tau_q(x,f) = \inner{\ell_q(x)}{\delta H_T^{(f)}} = \sum_\alpha t_\alpha$, whose magnitude factors exactly as $|\tau_q| = \sqrt{\Eincoh}\,\sqrt{\Rint}$, separating incoherent response energy $\Eincoh = \sum_\alpha t_\alpha^2$ from a scale-invariant interference ratio $\Rint = \tau_q^2 / \Eincoh$. Under this law, the visual puzzle of static class-head cancellation \citep{shu2026adjoint} and the language puzzle of prompt-conditioned gain modulation and sign reversal \citep{shu2026shared} become two complementary projections of the same rule: holding the task fixed unpacks how individual channels and pathways reinforce or cancel within one readout, whereas holding the factor fixed and varying the readout condition reveals how the same encoded information acquires different magnitudes or opposite signs across tasks. This resolves our opening question in a single testable statement: \emph{deep networks can jointly retain input information across disjoint representational units while combining those units' contributions with readout-dependent magnitudes and signs---enabling broadly preserved representations to yield highly selective outputs.}

Establishing a mechanistic account requires separating hypothesis discovery from causal proof. Throughout this work, source-coordinate inversion tools (LAC and closed-form source-grounded inversion \citep{shu2026adjoint,shu2026source}) serve strictly to surface candidate spatial and channel structures in readable input coordinates; every mechanistic claim is adjudicated natively in hidden tangent spaces and via held-out forward interventions, with the mathematical and empirical boundaries of source inversion explicitly bounded. Crucially, a genuine working principle must do more than rationalize descriptive decompositions post hoc---it must survive risky out-of-sample predictions when the network is actively perturbed. We subject the framework to two complementary predictive tests across large independent confirmation cohorts:
\begin{enumerate}
  \item \emph{Prospective antagonistic gating and same-forward-pass readout flipping.} If a strongly responding group of units genuinely cancels other contributions under a target readout, then locking its membership on a development split and attenuating its edit response by $10\%$ on unseen inputs should \emph{increase} the factor's net decision impact. We confirm this counterintuitive prediction across attention heads and last-layer MLPs in frozen Qwen2.5-1.5B and Llama-3.2-1B and in original ImageNet-pretrained ViT-B/16, where replacing contiguous channel-index blocks with scattered functional units resolves strong held-out antagonism. Most incisively, in ViT-B/16 (Block~10, 264 held-out images), reading two class margins from the same gated output shows that attenuating one frozen 384-unit group increases the mean background-edit effect on Margin~A ($+19.25 \times 10^{-5}$ logit) while decreasing it on Margin~B ($-10.90 \times 10^{-5}$ logit), whereas a second frozen group exhibits the opposite pattern ($-7.64$ versus $+40.50 \times 10^{-5}$ logit), yielding a four-cell paired group-by-readout interaction of $+78.30 \times 10^{-5}$ logit. Whether a group of units promotes or antagonizes a factor is therefore defined only jointly with the target readout.
  \item \emph{Quantitative prediction of channel recovery in structured pruning.} Turning the intervention from attenuation to restoration, we project the same signed-readout law onto the conditional innovation direction of a removed channel under a fixed linear successor, deriving an exact three-factor Frisch--Waugh--Lovell (FWL) recovery identity connecting conditional residual variance and fixed downstream readout coupling \citep{shu2026pruning}. On 4,096 independent ImageNet images spanning 10 cuts across ResNet-50, ViT-S/16, and Swin-T, incorporating fixed-readout coupling elevates held-out candidate recovery rank correlations to $0.865\text{--}1.000$ ($0.975\text{--}0.999$ across the six Vision Transformer cuts) and corrects actual top-channel selection in Swin-T.
\end{enumerate}

Equally important to a mechanistic theory are its boundaries. Our large-sample experiments reveal where simplified slogans fail: architectural labels alone do not dictate cancellation (Swin-T exhibits constructive texture interference, $R = 1.113$, and in ViT-B/16 both measured texture and random directions cancel strongly with no resolved excess cancellation for texture over random, $+0.006\,[-0.004, +0.017]$); grouping units by coarse structural pathways or contiguous indices dilutes observable antagonism; under input blur in Pets ResNet-18, relative background cancellation strengthens ($R^{(2)}$ drops from $0.906$ to $0.619$) even as absolute coherent background energy \emph{rises} ($0.00351 \to 0.00559$) and classification accuracy collapses ($87.6\% \to 25.9\%$); local single-cut FWL recovery does not by itself explain whole-network multi-layer pruning differences across architectures; and closed-form source inversion cannot substitute for native dual gradients when ranking signed input edits.

\paragraph{Summary of contributions.}
\begin{itemize}
  \item \textbf{Unified primal--dual theory (\cref{sec:framework}, Supplementary Sections~\ref{app:notation}--\ref{app:thm4}):} We formalize shared encoding on pre-fixed disjoint blocks and conditional signed readout under a single scale-invariant synthesis law, deriving an exact two-scale channel bridge, a path-gradient Gram integral for isotropic Gaussian random baselines, prospective gating criteria, an exact FWL channel-recovery identity, and a commutation-defect boundary for source-coordinate inversion.
  \item \textbf{Cross-modal native verification of encoding and readout (\cref{sec:exp_encoding,sec:exp_channels,sec:exp_paths,sec:exp_conditions}):} Across large held-out vision (1,024 CUB and 444 Pets images) and language (256 zero-overlap name pairs) cohorts, we independently establish disjoint-block shared encoding, demonstrate same-image coexistence of $R^{(512)} > 1$ and $R^{(2)} < 1$ (248/444 Pets images), map multi-pathway Gram geometry and coarse-to-fine grouping dilution across six architectures, and document conditional amplitude/composition shifts in vision alongside prompt-induced sign reversal in Qwen2.5 and vocabulary-dependent modulation in Llama-3.2.
  \item \textbf{Dual out-of-sample causal and quantitative predictions (\cref{sec:exp_gating,sec:exp_pruning}):} We validate prospective antagonistic gating across attention heads and functional MLP groups in vision and language---culminating in a same-unit, same-forward-pass cross-readout sign flip in ViT-B/16---and confirm quantitative single-channel recovery prediction across 10 pruning cuts on 4,096 ImageNet images.
  \item \textbf{Explicit mechanistic and observational boundaries (\cref{sec:discussion}, Supplementary Sections~\ref{supp:s1}--\ref{supp:s9}):} We document grouping-scale dependence, relative-versus-absolute decoupling under image corruption, single-cut versus whole-network pruning gaps, and cross-operator source-inversion limits, supported by full cohort-reuse and numerical-exactness audits.
\end{itemize}

%% file: sections/02_related_work.tex
\section{Related Work}
\label{sec:related}

Our framework connects four research traditions that have largely developed in parallel: mechanistic studies of distributed representations, feature inversion and attribution, contextual routing and background selectivity, and structured network pruning.

\subsection{Distributed Representations, Superposition, and Mechanistic Interpretability}
\label{sec:related_interp}

Since the foundational framing of distributed representations \citep{rumelhart1986learning}, neural networks have been understood to encode information across populations of units rather than strictly localized ``grandmother'' neurons. Recent mechanistic interpretability has sharpened this picture through the linear representation hypothesis \citep{park2024linear,hernandez2024linearity}, polysemanticity and superposition in overcomplete feature spaces \citep{elhage2022superposition}, concept activation vectors and network dissection \citep{bau2017network,kim2018interpretability}, and circuit-level decomposition via causal mediation, activation patching, and path patching \citep{vig2020investigating,geiger2021causal,meng2022locating,wang2023ioi,conmy2023automated,goldowskydill2023pathpatching,zhang2024patching}. Probing studies have likewise highlighted a persistent gap between whether a property is decodable from hidden states and whether the model actually relies on it for the task \citep{belinkov2022probing,ravichander2021probing}.

Our work advances this literature in two specific respects. First, on the encoding side, we distinguish generic distributed coding---where distinct subsets may carry non-overlapping pieces of a factor---from \emph{shared encoding} across pre-fixed disjoint blocks, verified natively via paired instance matching in vision and independent cross-role decoding in language. Second, on the readout side, rather than enumerating isolated task-specific circuits or assuming fixed functional labels for individual components, we formulate a single signed-readout synthesis law spanning channels and nonlinear paths, and prove directly via same-forward-pass cross-readout gating that whether a frozen set of units promotes or antagonizes a factor is a joint property of the unit group and the readout.

\subsection{Feature Inversion, Attribution, and Observation Boundaries}
\label{sec:related_inversion}

A large literature seeks to make hidden representations interpretable in input coordinates via gradient saliency \citep{simonyan2013deep,selvaraju2017grad,sundararajan2017axiomatic,ancona2018towards}, optimization-based preimage inversion and feature visualization \citep{zeiler2014visualizing,mahendran2015understanding,dosovitskiy2016inverting,olah2017feature,nash2019inverting}, embedding inversion \citep{morris2023text}, and closed-form adjoint or source-grounded estimators \citep{shu2026adjoint,shu2026source}. At the same time, sanity checks \citep{adebayo2018sanity}, critiques of attention weights as explanations \citep{jain2019attention,wiegreffe2019attention}, and the pattern-versus-filter distinction in linear decoding \citep{haufe2014interpretation,wiener1949extrapolation} demonstrate that readable input-space patterns cannot be equated automatically with causal sensitivity.

We incorporate this methodological insight directly into our theoretical architecture. Local Adjoint Correctors (LACs) \citep{shu2026adjoint} and source-grounded inversion with shared-component residualization \citep{shu2026source} played a genuine discovery role in surfacing candidate channel-group cancellation and spatial region preferences. However, \cref{ax:inversion_boundary} and \cref{thm:commutation} formally quarantine source inversion as a hypothesis-discovery instrument: ideal Wiener--Haufe commutation holds as a sufficient condition across sample-independent linear chains, whereas nonlinear networks incur an empirical commutation defect that decouples whole-field inversion from native signed decision gradients. All causal and mechanistic verdicts in this study therefore rest solely on native tangent-space measurements and held-out forward interventions.

\subsection{Visual Background Selectivity and Contextual Routing in Language Models}
\label{sec:related_routing}

In computer vision, extensive evidence shows that deep classifiers remain sensitive to scene backgrounds and spurious correlations even while achieving high foreground accuracy \citep{xiao2021noise,geirhos2020shortcut}. In autoregressive language models, entity-binding and factual-recall analyses show that relational structure is transported across token positions while late attention heads and MLPs compete, compensate, or actively suppress earlier signals---as documented in the Hydra self-repair effect \citep{mcgrath2023hydra}, copy suppression \citep{mcdougall2024copy}, conditional co-ablation \citep{gong2026coablation}, and factual recall circuits \citep{geva2023dissecting,feng2024binding}. Parallel studies of activation steering and inference-time intervention \citep{li2023iti,turner2023activation,zou2023representation} increasingly find that fixed steering vectors exhibit strong context dependence, unreliability, or even sign inversion across prompts \citep{tan2024steering,braun2025unreliability,torop2026inverted,li2026vectorfields}.

Our primal--dual framework brings these visual and linguistic phenomena under one roof. Static positive/negative head-weight cancellation in CNNs, competing residual/attention/MLP paths in Vision Transformers, and prompt-conditioned modulation or sign flips in frozen LLMs emerge as specializations of the same signed-readout law under progressively richer cut degrees of freedom. Moreover, our coarse-to-fine grouping experiments clarify why antagonistic structure is frequently missed when units are aggregated by architectural pathway or contiguous channel index.

\subsection{Structured Pruning, Functional Redundancy, and Local Recovery}
\label{sec:related_pruning}

Structured and unstructured network compression traditionally rank units either by activation/weight magnitude and layer-adaptive heuristics \citep{lee2021layer,sun2024simple}, by second-order loss sensitivity \citep{lecun1989optimal,hassibi1993optimal}, or by local layer-wise reconstruction and least-squares weight folding \citep{frantar2023sparsegpt,shu2026pruning}. Unlabeled variance or covariance criteria ask how much of a unit's representation cannot be reconstructed from retained units, whereas task-aware criteria weight perturbations by downstream sensitivity.

By projecting our signed-readout law onto a candidate unit's conditional innovation direction under a fixed linear successor, we obtain an exact three-factor Frisch--Waugh--Lovell identity \citep{frisch1933partial,lovell1963seasonal} that resolves an open question from our companion pruning study \citep{shu2026pruning}: knowing a channel's conditional residual variance relative to retained units is not sufficient to rank its recovery value without also accounting for how the fixed downstream linear readout sums its signed contributions (conditional readout). At the same time, our direct-slicing diagnostics across ViT-S/16 and Swin-T show cleanly where single-cut local linearization stops and where whole-network multi-layer pruning remains an open problem.

%% file: sections/03_framework.tex
\section{A Unified Primal--Dual Framework}
\label{sec:framework}

We formalize inference in a trained, parameter-frozen differentiable network as a two-stage primal--dual computation. To keep the main text streamlined and directly tied to testable predictions, we state the core definitions, main law, testable consequences, and observational boundaries here, deferring full proofs, exact analytical examples, and integral derivations to Supplementary Sections~\ref{app:notation}--\ref{app:thm4}.

\subsection{Two-Stage Inference Cut and Primal--Dual Separation}
\label{sec:fw_primal_dual}

Fix an input instance $x \in \Xcal$, an internal representation cut $T$, and a task or query contrast $q \in \Qcal$. We decompose the forward map at cut $T$ into an upstream representation stage $f_{\mathrm{up}}$ and a downstream vector-valued output stage $F_{\mathrm{down}}(\cdot;\, x, q) \in \Rbb^{d_{\mathrm{out}}}$ (class or token logits), projected by a contrast vector $u_{\mathrm{out}}(q) \in \Rbb^{d_{\mathrm{out}}}$ into a scalar output margin $m_q(x) := u_{\mathrm{out}}(q)^\top F_{\mathrm{down}}(H_T(x);\, x, q) \in \Rbb$:
\begin{equation}
  x \in \Xcal \;\xrightarrow{\;f_{\mathrm{up}}\;\text{(encoding)}\;}\; H_T(x) \in \Hcal_T \;\xrightarrow{\;F_{\mathrm{down}}(\cdot;\,x,q)\;}\; \Rbb^{d_{\mathrm{out}}} \;\xrightarrow{\;u_{\mathrm{out}}(q)^\top\;}\; m_q(x) \in \Rbb.
  \label{eq:two_stage}
\end{equation}
In vision classifiers, $f_{\mathrm{up}}$ is independent of the class contrast $q$ chosen at the head; in autoregressive language models, query tokens appear in $x$, yet intermediate hidden states at cut $T$ can still retain relational structure shared across distinct query roles \citep{shu2026shared}. A controlled factor edit $f$ induces a primal perturbation $\delta H_T^{(f)} \in \Hcal_T$ (either pushed forward from an input edit $\delta x$ via $\delta H_T^{(f)} = J_{\mathrm{up},T}(x)\,\delta x$ or injected directly at cut $T$). Meanwhile, the downstream network defines the dual readout covector
\begin{equation}
  \ell_q(x) \;:=\; \nabla_{H_T} m_q(x) \;=\; J_{F_{\mathrm{down}}}(x;\,q)^*\, u_{\mathrm{out}}(q) \;\in\; \Hcal_T^*,
  \label{eq:dual_covector}
\end{equation}
where $J_{F_{\mathrm{down}}}(x;\,q)$ is the Jacobian of $F_{\mathrm{down}}$ with respect to $H_T$ and $^*$ denotes the adjoint transpose.

\begin{axiom}[Strict Primal--Dual Separation]
\label{ax:primal_dual}
At any cut $T$, activations $H_T(x)$ and factor perturbations $\delta H_T^{(f)}$ belong to the primal space $\Hcal_T$, whereas downstream readout sensitivities $\ell_q(x)$ belong to the dual space $\Hcal_T^*$. Under diagonal coordinate rescaling $H_{T,k} \mapsto d_k H_{T,k}$ ($d_k \neq 0$), primal perturbations scale by $d_k$ while both dual gradients $\ell_{q,k}$ and linear state estimators $\Aop_x$ scale by $d_k^{-1}$; dual covectors therefore cannot be fed directly into primal state estimators without covariance metric conversion. Moreover, whenever cut $T$ forms a complete vertex cut intercepting all computational paths from $x$ to $m_q$, the input gradient satisfies $\nabla_x m_q(x) = J_{\mathrm{up}}(x)^* \ell_q(x)$ and gives the exact first-order directional derivative $\inner{\nabla_x m_q(x)}{\delta x}$ (if $T$ lies on a partial branch, $\nabla_x m_q$ contains an additional bypass term; \cref{app:thm1}).
\end{axiom}

\subsection{Pillar~1: Operationalizing Shared Encoding Across Disjoint Blocks}
\label{sec:fw_encoding}

To separate shared encoding from generic distributed representations---where complementary subsets might each hold unusable fragments---we partition $\Hcal_T$ using a \emph{pre-fixed}, task-independent family of mutually orthogonal coordinate-block projections $\{\Pi_m\}_{m=1}^M$ satisfying $\Pi_m \Pi_{m'} = \delta_{mm'}\Pi_m$ and $\sum_{m=1}^M \Pi_m = I_{\Hcal_T}$ (in our experiments, $M=4$ non-overlapping channel or hidden-dimension blocks).

\begin{definition}[Disjoint-Block Shared Encoding]
\label{def:shared_encoding}
A representation cut $T$ exhibits \emph{shared encoding} for a controlled factor $f$ if the factor's perturbation response $\delta H_T^{(f)}$ retains recognizable structure across each pre-fixed disjoint block $\Pi_m \Hcal_T$ ($m=1,\dots,M$). Because controlled edits differ structurally across vision and language, we operationalize this requirement natively in two domain-appropriate ways:
\begin{enumerate}
  \item \emph{Paired instance matching in vision:} After subtracting the development-set mean response for a given texture contrast, each disjoint block $\Pi_m$ uses exclusively its own raw channels---and its complement $(I-\Pi_m)$ independently uses exclusively its own channels---to estimate coordinates in a pre-fixed low-dimensional development subspace; the resulting coordinate estimates align significantly more closely for the same image than for mismatched images edited by the same texture (testing shared instance-level structure rather than single-block image identity decoding).
  \item \emph{Independent cross-role decoding in language:} Using a relational direction estimated under one query role on development names, each disjoint block $\Pi_m$ independently discriminates the relational factor on held-out name pairs evaluated under a different query role.
\end{enumerate}
\end{definition}

Two methodological safeguards follow immediately from \cref{def:shared_encoding}. First, subtracting the development-set mean response for each texture contrast removes the common texture-pair mean offset while retaining instance-specific variation across images (though amplitude differences along a shared axis can still remain). Second, the partition $\{\Pi_m\}$ must be fixed prior to readout evaluation: selecting blocks post hoc by task performance would conflate encoding with readout.

\subsection{Pillar~2: The Conditional Signed-Readout Law}
\label{sec:fw_readout}

Given a factor perturbation $\delta H_T^{(f)}$ at cut $T$, we decompose $\Hcal_T$ by a complete orthogonal unit or group partition $\{\Pi_k\}_{k=1}^K$ ($\sum_k \Pi_k = I_{\Hcal_T}$) and decompose the downstream sensitivity across a complete tangent-space inclusion--exclusion set of computational paths $\ell_q(x) = \sum_{\pi \in \Pcal} \ell_q^{(\pi)}(x)$ (for instance, bypass, attention-only, MLP-only, and mixed paths $D, A, M, J$ in Transformers, or residual/convolutional paths $SS, SF, FS, FF$ in ResNets; exact expressions in \cref{app:thm1}). Indexing active components by $\Omega(T) = \{(k,\pi)\}$, each component carries a signed directional contribution
\begin{equation}
  t_{k,\pi}(x, q;\, f) \;:=\; \inner{\ell_q^{(\pi)}(x)}{\Pi_k\,\delta H_T^{(f)}}.
  \label{eq:component_contrib}
\end{equation}

\begin{theorem}[Conditional Signed-Readout Synthesis Law]
\label{thm:synthesis}
For any differentiable cut $T$, complete orthogonal partition $\{\Pi_k\}$, and complete tangent-space path decomposition $\Pcal$, the first-order directional effect $\tau_q(x,f)$ of factor $f$ on margin $m_q(x)$ obeys
\begin{equation}
  \tau_q(x, f) \;=\; \inner{\ell_q(x)}{\delta H_T^{(f)}} \;=\; \sum_{\alpha \in \Omega(T)} t_\alpha(x, q;\, f).
  \label{eq:signed_sum}
\end{equation}
Define the readout-weighted incoherent energy $\Eincoh(f;\,x,q) := \sum_{\alpha \in \Omega(T)} t_\alpha(x,q;f)^2$. Whenever $\Eincoh > 0$, the dimensionless \emph{phase-coherence interference ratio}
\begin{equation}
  \Rint^{\Omega(T)}(f;\, x, q) \;:=\; \frac{\Big(\sum_{\alpha \in \Omega(T)} t_\alpha\Big)^2}{\sum_{\alpha \in \Omega(T)} t_\alpha^2} \;=\; 1 + \sum_{\alpha < \beta} \frac{2\, t_\alpha t_\beta}{\Eincoh}
  \label{eq:interference_ratio}
\end{equation}
satisfies the exact amplitude--composition factorization
\begin{equation}
  \abs{\tau_q(x, f)} \;\equiv\; \sqrt{\Eincoh(f;\, x, q)}\,\sqrt{\Rint^{\Omega(T)}(f;\, x, q)}.
  \label{eq:factorization}
\end{equation}
Moreover: (i) every term $t_\alpha$, $\Eincoh$, and $\Rint^{\Omega(T)}$ is strictly invariant under any block-invertible change of basis $M = \bigoplus_k M_k$ commuting with $\{\Pi_k\}$ (including arbitrary nonzero coordinate rescalings and sign flips) and requires no division by baseline channel norms; and (ii) assigning independent Rademacher signs $s_\alpha \in \{-1,+1\}$ to the components yields the exact expectation baseline $\Ebb_s[\Rint^{\Omega(T)}] = 1$ whenever $\Eincoh > 0$ (while any single sign realization may deviate from~$1$).
\end{theorem}

The two quantities in \cref{thm:synthesis} answer distinct questions. The signed sum $\tau_q$ gives the factor's net first-order effect on the current output, whereas $\Rint^{\Omega(T)}$ measures how strongly the specified partition reinforces ($\Rint > 1$) or cancels ($\Rint < 1$) contributions along that readout. We also distinguish \emph{structural Type~0} vanishing---where every individual component vanishes identically for perturbations in a subspace $\mathcal{S}$ ($\forall \alpha \in \Omega(T),\, t_\alpha \equiv 0$, so $\Eincoh \equiv 0$ and $\Rint$ is undefined, as for non-\texttt{[CLS]} terminal patches in a pure \texttt{[CLS]}-head ViT)---from \emph{exact destructive cancellation} ($\Eincoh > 0$, $\tau_q = 0$, $\Rint = 0$).

Crucially, seemingly different readout phenomena across architectures are specializations of the single covector $\ell_q(x) = J_{F_{\mathrm{down}}}(x;q)^* u_{\mathrm{out}}(q)$ under progressively tighter structural constraints (\cref{tab:cut_taxonomy}). Static head-weight cancellation in a CNN (\cref{sec:exp_channels}), nonlinear multi-path synthesis in vision backbones (\cref{sec:exp_paths}), and prompt-conditioned modulation or sign reversal in frozen LLMs (\cref{sec:exp_conditions}) all obey \cref{eq:signed_sum,eq:factorization}.

\begin{table}[t]
\centering
\caption{\textbf{Cut specialization hierarchy of the conditional signed-readout law.} Distinct empirical behaviors across vision and language architectures arise as constrained projections of the single dual readout covector $\ell_q(x) = J_{F_{\mathrm{down}}}(x;q)^* u_{\mathrm{out}}(q)$.}
\label{tab:cut_taxonomy}
\small
\begin{tabularx}{\textwidth}{@{}l >{\raggedright\arraybackslash}p{4.7cm} l >{\raggedright\arraybackslash}X@{}}
\toprule
\textbf{Cut regime} & \textbf{Constraint on $\ell_q(x)$} & \textbf{Active axes} & \textbf{Representative models and behavior} \\
\midrule
Type~III (full dynamic) & Unconstrained: both $J_{F_{\mathrm{down}}}(x;q)$ and $u_{\mathrm{out}}(q)$ vary with context and query & $k \times p \times \pi \times q$ & Frozen LLMs (Qwen2.5, Llama-3.2): competing paths ($D,A,M,J$), prompt-conditioned gain modulation and sign flips (\cref{sec:exp_paths,sec:exp_conditions}). \\
Type~II (nonlinear static-$q$) & Fixed readout contrast $q$, instance-dependent nonlinear $J_{F_{\mathrm{down}}}(x)$ and multiple paths $\pi$ & $k \times p \times \pi$ & Intermediate CNN and ViT/Swin cuts: backbone-dependent path Gram geometry, class-pair reweighting, functional MLP gating (\cref{sec:exp_paths,sec:exp_conditions,sec:exp_gating}). \\
Type~I (fixed linear head) & Constant downstream Jacobian $J_{F_{\mathrm{down}}} \equiv W$ and single path $|\Pcal|=1$ & $k$ (units/groups) & Final CNN head (Pets ResNet-18 two-scale synthesis, \cref{sec:exp_channels}) and post-GELU \texttt{fc2} pruning cuts (\cref{sec:exp_pruning}). \\
Type~0 (structural zero) & $\forall \alpha \in \Omega(T),\, t_\alpha \equiv 0$ on subspace $\mathcal{S}$ ($\Eincoh \equiv 0$, $\Rint$ undefined) & None on $\mathcal{S}$ & Pure ViT-B/16 final-layer non-\texttt{[CLS]} patches under a \texttt{[CLS]}-only head (\cref{sec:exp_paths}). \\
\bottomrule
\end{tabularx}
\end{table}

\subsection{Four Testable Mechanistic Consequences}
\label{sec:fw_consequences}

Four quantitative predictions follow directly from \cref{thm:synthesis} and drive our experimental design in \cref{sec:experiments}.

\paragraph{1. Exact two-scale bridge and grouping-scale dilution.}
Given any mutually disjoint coarse grouping $\Gcal = \{g_1,\dots,g_M\}$ partitioning the fine components $\Omega(T)$ with coarse group sums $U_m := \sum_{\alpha \in g_m} t_\alpha$, let $R_{\mathrm{within}} := \sum_m \sum_{\alpha \neq \beta \in g_m} t_\alpha t_\beta / \Eincoh$ and $R_{\mathrm{between}} := \sum_{m \neq m'} U_m U_{m'} / \Eincoh$. Whenever $1 + R_{\mathrm{within}} > 0$, the fine and coarse interference ratios satisfy the exact identity (proof in \cref{app:thm3})
\begin{equation}
  \Rint^{\mathrm{fine}} \;=\; 1 + R_{\mathrm{within}} + R_{\mathrm{between}}, \qquad
  \Rint^{\mathrm{group}} \;:=\; \frac{\big(\sum_m U_m\big)^2}{\sum_m U_m^2} \;\equiv\; \frac{\Rint^{\mathrm{fine}}}{1 + R_{\mathrm{within}}} \;=\; 1 + \frac{R_{\mathrm{between}}}{1 + R_{\mathrm{within}}}.
  \label{eq:two_scale}
\end{equation}
Equation~\eqref{eq:two_scale} yields two immediate consequences. First, whenever within-group synergy exceeds the magnitude of between-group cancellation ($R_{\mathrm{within}} > |R_{\mathrm{between}}|$ with $R_{\mathrm{between}} < 0$), fine-scale net reinforcement ($\Rint^{\mathrm{fine}} > 1$) and coarse-scale net cancellation ($\Rint^{\mathrm{group}} < 1$) strictly coexist on the \emph{same} perturbation (\cref{sec:exp_channels}). Second, collapsing opposing units into a single coarse pathway bin or contiguous coordinate block absorbs their negative cross-terms into the coarse denominator, diluting observable cancellation or antagonism (\cref{sec:exp_paths,sec:exp_gating}).

\paragraph{2. Path-gradient Gram geometry for isotropic Gaussian random baselines.}
At a multi-path cut with path readout gradients $a_i := J_i^* u_q$ ($i=1,\dots,P$), define the $P \times P$ path-gradient Gram matrix $G_{ij} := a_i^\top a_j$ with eigendecomposition $G = U \diag(\lambda) U^\top$. For an isotropic Gaussian (equivalently, uniform spherical) random perturbation $v_{\mathrm{rand}}$, the path response vector is Gaussian up to scale, yielding an exact one-dimensional quadrature formula for the direction-wise expected interference ratio (\cref{app:thm3}):
\begin{equation}
  \Ebb\!\left[\Rint(v_{\mathrm{rand}})\right] \;=\; \sum_{j=1}^P (\mathbf{1}^\top U_{:j})^2 \int_0^\infty \frac{\lambda_j}{1 + 2t\lambda_j}\prod_{k=1}^P (1 + 2t\lambda_k)^{-1/2}\,dt.
  \label{eq:gram_integral}
\end{equation}
The Gram matrix thus specifies the exact population baseline under isotropic Gaussian (or spherical uniform) perturbations, whereas the interference ratio of a specific texture or relational factor additionally depends on how that factor projects onto the path gradients $\{a_i\}$ (\cref{sec:exp_paths}).

\paragraph{3. Prospective gating under real forward interventions.}
To test signed cancellation causally rather than descriptively, consider attenuating a candidate unit group $G$ by a small factor $\lambda \in (0,1)$ applied exclusively to the edit increment $\delta h_G := h_G(x_{\mathrm{edit}}) - h_G(x_{\mathrm{clean}})$ while holding the clean operating point fixed (physical output scaling is evaluated as a companion control in \cref{supp:s6}). Define the held-out paired gating endpoint
\begin{equation}
  H_G(\lambda) \;:=\; \abs{\Delta m_G(\lambda)} - \abs{\Delta m(0)},
  \label{eq:gating_H}
\end{equation}
where $\Delta m_G(\lambda)$ is the edited-minus-clean margin change under attenuation $\lambda$. In a linearized single-step model ($\tau_G(\lambda) = \tau(0) - \lambda t_G$), attenuating a \emph{promoting} group (same sign as $\tau(0)$) preserves the original margin sign whenever $0 < \lambda < |\tau(0)|/|t_G|$ and strictly reduces $|\tau_G(\lambda)|$ whenever $0 < \lambda < 2|\tau(0)|/|t_G|$ (\cref{app:thm4}). In real nonlinear networks, we use the development-set local gating slope $\kappa_G := \partial |\Delta m_G(\lambda)|/\partial\lambda|_{\lambda=0}$ to select and freeze candidate promoting and antagonistic groups, predict that at a frozen small dose ($\lambda_0 = 0.1$) attenuating an antagonistic group yields $H_G(\lambda_0) > 0$ (weakening the group \emph{amplifies} the factor's decision impact) while attenuating a promoting group yields $H_G(\lambda_0) < 0$, and further predict that the same frozen group can shift the mean $H_G$ in opposite directions across two readouts $q_A, q_B$ during the same forward pass. While the local sum \cref{eq:signed_sum} holds by definition for any complete decomposition, whether development-selected groups predict the sign and magnitude of finite nonlinear gating on held-out inputs is a directly falsifiable empirical prediction (\cref{sec:exp_gating}).

\paragraph{4. Fixed-readout channel recovery in structured pruning (FWL identity).}
Finally, consider a Type~I cut with centered calibration activations $A \in \Rbb^{N \times C}$, a fixed linear successor $Y = A W^\top \in \Rbb^{N \times d_{\mathrm{out}}}$, a positive semidefinite output metric $\Hout \succeq 0$, and optimal least-squares weight folding $W'_{\mathrm{keep}} = W_S + W_R B_S^\top$ ($B_S = A_S^+ A_R$) onto a retained subset $S \subset \{1,\dots,C\}$ with reconstruction loss $L(S) := \frac{1}{N}\|(I - P_S)Y\|_{F,\Hout}^2$ \citep{shu2026pruning}. Restoring a single omitted unit $j \notin S$ injects the conditional residual $r_{j|S} := (I - P_S)a_j$ with conditional residual variance $v_j := \frac{1}{N}\|r_{j|S}\|_2^2$ relative to $S$ and induces the conditional innovation vector $d_{j|S} := A^\top r_{j|S} / \|r_{j|S}\|_2^2$ (satisfying $(d_{j|S})_j = 1$). Writing $z_k := W \Pi_k d_{j|S}$ for the block-wise readout contributions, the signed-readout law yields an exact recovery identity (\cref{app:thm4}):

\begin{theorem}[FWL Three-Factor Channel-Recovery Identity]
\label{thm:fwl}
Under a fixed linear successor $Y = A W^\top$, positive semidefinite metric $\Hout \succeq 0$, and optimal least-squares weight folding, whenever $v_j > 0$ and $\sum_k \|z_k\|_{\Hout}^2 > 0$, restoring candidate channel $j \notin S$ reduces output loss by
\begin{equation}
  L(S) - L(S \cup \{j\}) \;\equiv\; \underbrace{v_j}_{\substack{\text{conditional residual}\\\text{variance}}} \;\cdot\; \underbrace{\left(\sum_k \norm{z_k}_{\Hout}^2\right)}_{\substack{\text{incoherent readout}\\\text{energy}}} \;\cdot\; \underbrace{\Rint^{\Hout}(d_{j|S})}_{\substack{\text{signed readout synthesis}\\\text{along } d_{j|S}}}.
  \label{eq:fwl_identity}
\end{equation}
\end{theorem}

Theorem~\ref{thm:fwl} holds directly from the linear successor and least-squares folding geometry (without requiring the cross-block recognizability condition of \cref{def:shared_encoding}): predicting how much local output error a restored channel recovers requires \emph{both} its conditional residual variance relative to retained units ($v_j$) and how the fixed downstream linear readout combines its signed contributions (\cref{sec:exp_pruning}).

\subsection{Observational Scope of Source-Coordinate Inversion}
\label{sec:fw_inversion}

Because our visual hypothesis originated from Local Adjoint Correctors and closed-form source-grounded inversion \citep{shu2026adjoint,shu2026source}, we explicitly bound what source-coordinate estimators can and cannot establish.

\begin{axiom}[Boundary Between Source-Geometry Observation and Native Causality]
\label{ax:inversion_boundary}
A source-grounded linear estimator $\Aop_x(U)$ reflects the joint geometry of $(H_T(x), J_{\mathrm{up}}(x), \Theta_{\mathrm{cal}})$ and serves to expose candidate shared or channel-specific structures in readable input coordinates (e.g., via shared-axis residualization $r_q = u_q - a_q b$ in \citet{shu2026source}). However, a single-line counterexample demonstrates that Jacobian-conditioned pullbacks cannot establish functional carrying on their own: for $h(x) = \|x\|_2$ ($x \neq 0$), which discards all $(d-1)$ angular degrees of freedom, $J_x = x^\top/\|x\|_2$ satisfies $J_x^\top c = c\,x/\|x\|_2$ for \emph{any} positive seed $c > 0$. Consequently, every causal or functional claim must be adjudicated by native tangent-space measurements and held-out forward interventions rather than source-space visual similarity.
\end{axiom}

\begin{theorem}[Sufficient Linear Commutation and Nonlinear Commutation Defect]
\label{thm:commutation}
As a sufficient condition for exact commutation, let $H_u = J_v H_v$ be a sample-independent linear map with $\Sigma_u = J_v \Sigma_v J_v^*$. Then the exact linear MMSE (Wiener) state estimator $\Wop_{v\leftarrow u} = \Sigma_v J_v^*(J_v \Sigma_v J_v^*)^+$ commutes identically with dual backpropagation (\cref{app:thm2}):
\begin{equation}
  \Wop_{v\leftarrow u}(\Sigma_u \ell_u) \;\equiv\; \Sigma_v (J_v^* \ell_u) \;=\; \Sigma_v \ell_v.
  \label{eq:wiener_commute}
\end{equation}
In deep nonlinear networks, this identity incurs an empirical upstream commutation defect
\begin{equation}
  \Dcomm(x;\, q) \;:=\; \Aop_x\!\big(\Sigma_T^{\mathrm{global}}\ell_q(x)\big) - \Sigma_X J_{\mathrm{up}}(x)^*\ell_q(x),
  \label{eq:comm_defect}
\end{equation}
combining finite-sample estimation, global-versus-local statistics mismatch, ridge regularization, and approximation error; empirically, we summarize this vector discrepancy per image via the scalar \emph{normalized commutation defect} $\|\Dcomm(x;q)\|_2 / \|\Psi(x)\|_2$, where $\Psi(x) := \Sigma_X \nabla_x m_q(x)$. Signed input-direction prediction is therefore governed natively by $\nabla_x m_q(x)$ (and its covariance-weighted primal counterpart $\Psi(x)$), while \cref{thm:commutation} calibrates the nonlinear operating boundary of closed-form source inversion (\cref{sec:disc_inversion}, \cref{supp:s9}).
\end{theorem}

%% file: sections/04a_protocol_encoding_channels.tex
\section{Experimental Verification}
\label{sec:experiments}

We evaluate the primal--dual framework across seven vision and language architectures using large, pre-separated confirmation cohorts. The experimental sequence mirrors the structure of the paper: \cref{sec:exp_protocol} establishes native operational protocols, cohort provenance, and numerical audits; \cref{sec:exp_encoding} tests \emph{shared encoding} natively across disjoint blocks; \cref{sec:exp_channels,sec:exp_paths,sec:exp_conditions} test \emph{conditional signed readout} under fixed-task decomposition and varying-task switches; and \cref{sec:exp_gating,sec:exp_pruning} subject the framework to prospective causal gating and quantitative pruning-recovery predictions.

\subsection{Experimental Protocols, Cohorts, and Exactness Audits}
\label{sec:exp_protocol}

All mechanistic claims are tested directly on parameter-frozen pretrained or task-trained networks without relying on feature inverters. \Cref{tab:protocols} summarizes the eight native experimental protocols, their fixed versus varied quantities, and their independent statistical units; a consolidated confirmation ledger across all endpoints and scope boundaries is archived in \cref{tab:master_summary} (\cref{supp:s1}).

\paragraph{Cohort separation, reuse disclosures, and statistical conventions.}
Independent resampling units are always whole source images (vision) or complete entity name pairs (language); multiple donor edits per image, multiple prompts per name pair, and multiple readouts on the same forward pass are aggregated within each unit before paired bootstrap resampling ($10{,}000$ draws by default; $20{,}000$ draws for language attention-head gating and CUB regional signed Pearson correlation; $4{,}000$ draws for ImageNet pruning; and $1{,}000$ draws for Pets dose-level derivative-to-finite Spearman calibration). Unless explicitly marked as multiplicity-adjusted ($97.5\%$, $98.75\%$, or $99.375\%$), brackets denote two-sided $95\%$ paired bootstrap confidence intervals. To prevent inflated sample counts, we separate initial exploratory cohorts from final confirmation cohorts throughout and disclose every cross-section reuse explicitly (full audit in \cref{supp:s1}):
\begin{itemize}
  \item \textbf{CUB-200-2011 \citep{wah2011caltech}:} $1{,}024$ independent confirmation images are shared across ResNet-18, ResNet-50, Swin-T, and ViT-B/16 encoding, pathway, and readout analyses ($1{,}020$ satisfy the spatial-region eligibility filter in \cref{supp:s9}), alongside a separate $512$-image development set and an independent $264$-image reserve used for ViT-B/16 functional-group holdout confirmation (the same $264$ images were also used in a CUB-finetuned ViT-B/32 comparison in \cref{supp:s6}).
  \item \textbf{Oxford-IIIT Pets \citep{parkhi2012cats}:} Two independent fresh cohorts of $128$ and $444$ source images ($572$ total, each paired with four exclusive donor images; $442$ eligible for the weak warm/cool tint probe), kept strictly distinct from the historical $16$ + $48$ = $64$-image exploratory archive.
  \item \textbf{Frozen LLMs (Qwen2.5-1.5B \citep{qwen2024technical} and Llama-3.2-1B \citep{meta2024llama32}):} $256$ zero-overlap confirmation name pairs ($512$ distinct names) are shared by both models for cross-role encoding, four-path decomposition, and attention-head gating; $52$ pairs passing single-token tokenization in both tokenizers and both word positions form the $2\times 2$ factorial cohort; and a separate set of $256$ additional zero-overlap name pairs ($512$ distinct names) provides independent confirmation for last-layer MLP functional grouping and coarse-to-fine pathway splitting.
  \item \textbf{ImageNet-1K \citep{deng2009imagenet} structured pruning:} $4{,}096$ confirmation images spanning all $1{,}000$ classes are shared across all $10$ cuts of ResNet-50, ViT-S/16, and Swin-T ($257$ images overlap with the historical $3{,}200$-image whole-network evaluation from \citet{shu2026pruning}, reported separately in \cref{tab:s7b_whole}).
\end{itemize}

\paragraph{Numerical closure and physical-surgery audits.}
Before evaluating mechanistic hypotheses, we verified that every decomposed quantity matches exact forward execution on the final confirmation cohorts (\cref{tab:s1_audit}). Tangent-space path sums close to within $5.26\times 10^{-7}$ in CUB ResNet-18 \texttt{layer3} ($1{,}024$ images; $4.20\times 10^{-8}$ in \texttt{layer4.0}); diagonal coordinate rescaling obeys \cref{ax:primal_dual} to $2.32\times 10^{-6}$ across the $444$ Pets images; fast FWL scoring matches explicit least-squares refitting to $\le 9.43\times 10^{-18}$ on early/middle Vision Transformer cuts ($9.11 / 8.81\times 10^{-18}$ at the middle cuts; $\le 1.005\times 10^{-14}$ across all six cuts on $4{,}096$ images); finite-difference versus directional-derivative sign agreement reaches $98.27\%$ (early) and $99.44\%$ (late) in CUB ResNet-18 ($1{,}024$ images) and $99.988\%/99.994\%$ (main/half dose) in ViT-B/16 ($1{,}024$ images); full-network hook audits across $21{,}504$ language MLP gating conditions match batched evaluations identically; and across all $36$ physical-pruning surgery configurations $\times 4{,}096$ images ($147{,}456$ pairs), every pair passes the class-margin tolerance (the $732$ float32 local-MSE exceptions under late Swin post-GELU direct slicing drop to a maximum discrepancy of $6.22\times 10^{-15}$ when replayed in float64).

\begin{table}[t]
\centering
\caption{\textbf{Operational protocols of the eight native experiments.} Each test freezes model parameters and isolates a controlled perturbation or readout change, summarized at the level of independent source images or complete entity name pairs.}
\label{tab:protocols}
\scriptsize
\begin{tabularx}{\textwidth}{@{}l >{\raggedright\arraybackslash}p{3.0cm} >{\raggedright\arraybackslash}p{4.3cm} >{\raggedright\arraybackslash}X@{}}
\toprule
\textbf{Experiment (\S)} & \textbf{Frozen models \& cuts} & \textbf{Fixed vs.\ varied quantities} & \textbf{Native endpoint \& unit} \\
\midrule
Shared encoding, vision (\ref{sec:exp_encoding}) & CUB ResNet-18 (\texttt{layer3}, \texttt{layer4.0}), ViT-B/16 (Block~6); Pets ResNet-18 & Fixed texture pairs \& dev 16D basis; compare same-image block-to-complement vs.\ mismatched same-texture image & Mean-centered cosine advantage across 4 disjoint blocks (1,024 CUB); weak tint probe (442 Pets); source image \\
Shared encoding, language (\ref{sec:exp_encoding}) & Qwen2.5-1.5B (Block~14), Llama-3.2-1B (Block~8) & Frozen dev relation direction; evaluate across query roles in 4 disjoint hidden-dim blocks & Within-name-pair cross-role ranking AUC \& cosine (256 name pairs); name pair \\
Channel synthesis (\ref{sec:exp_channels}) & Pets ResNet-18 final 512 channels & Fixed source image, clean top-1 class, 4 donor BG/FG edits; head sign partition vs.\ random partitions & Signed contributions, two-scale $R^{(2)}$ \& $R^{(512)}$, finite forward rank corr (444 + 128 images); source image \\
Pathway geometry (\ref{sec:exp_paths}) & ResNet-18/50, Swin-T, ViT-B/16 (CUB); Qwen2.5, Llama-3.2 & Fixed input \& readout; vision: subspace-projected, unit-normalized channel vector from background edits (broadcast across spatial/token sites) vs.\ matched random channel vectors; LLM: pair-specific interaction projected onto frozen dev subspace (query-token attention output); coarse vs.\ functional MLP split & Directional $R$, Gram integral, 6 normalized cross-terms, clamped forward, coarse$-$fine $\Delta R$; image / name pair \\
Readout switch, vision (\ref{sec:exp_conditions}) & CUB ResNet-18 (\texttt{layer3}) & Fixed clean image, subspace-projected unit-normalized background channel vector broadcast at \texttt{layer3}, dose, and identical 8 bird classes; re-pair within vs.\ across habitats & Unit-dose $|\Delta m|$ DiD (full/half dose) \& $|\tau|=\sqrt{E}\sqrt{R}$ decomposition (1,024 images); source image \\
Readout switch, language (\ref{sec:exp_conditions}) & Qwen2.5 (Block~14), Llama-3.2 (Block~8) & Frozen cross-name relation axis ($0.5ab\,v$, query-token attention output); $2\times 2$ presentation (Direct/Alias) $\times$ answer pair (RB/OP) & Signed finite selection contrast $\Delta G$ ($10^{-3}$ logit, 52 dual-tokenizer pairs); name pair \\
Prospective gating (\ref{sec:exp_gating}) & Qwen2.5, Llama-3.2 (heads \& last MLP); ViT-B/16 (Block~6, 10) & Dev-selected frozen units; $10\%$ edit-increment attenuation ($\lambda_0=0.1$, clean point fixed); ViT same-forward readout flip & Held-out $H_G = |\Delta m_G| - |\Delta m(0)|$ vs.\ contiguous \& strength-matched controls; image / name pair \\
Pruning recovery (\ref{sec:exp_pruning}) & ResNet-50 (4 bottlenecks), ViT-S/16 \& Swin-T (early/mid/late MLP) & Frozen retained set $S$, OLS folding, linear successor; restore each candidate channel one by one & Linear-successor MSE drop, candidate Spearman $\rho$, single-block margin MAE (4,096 images); source image \\
\bottomrule
\end{tabularx}
\end{table}

\subsection{Shared Encoding Across Disjoint Representation Blocks}
\label{sec:exp_encoding}

We first test Pillar~1 (\cref{def:shared_encoding}): whether a controlled input factor leaves recognizable structure across pre-fixed, mutually disjoint coordinate blocks in native representation space, independent of any feature inverter.

\begin{figure}[t]
  \centering
  \includegraphics[width=\textwidth]{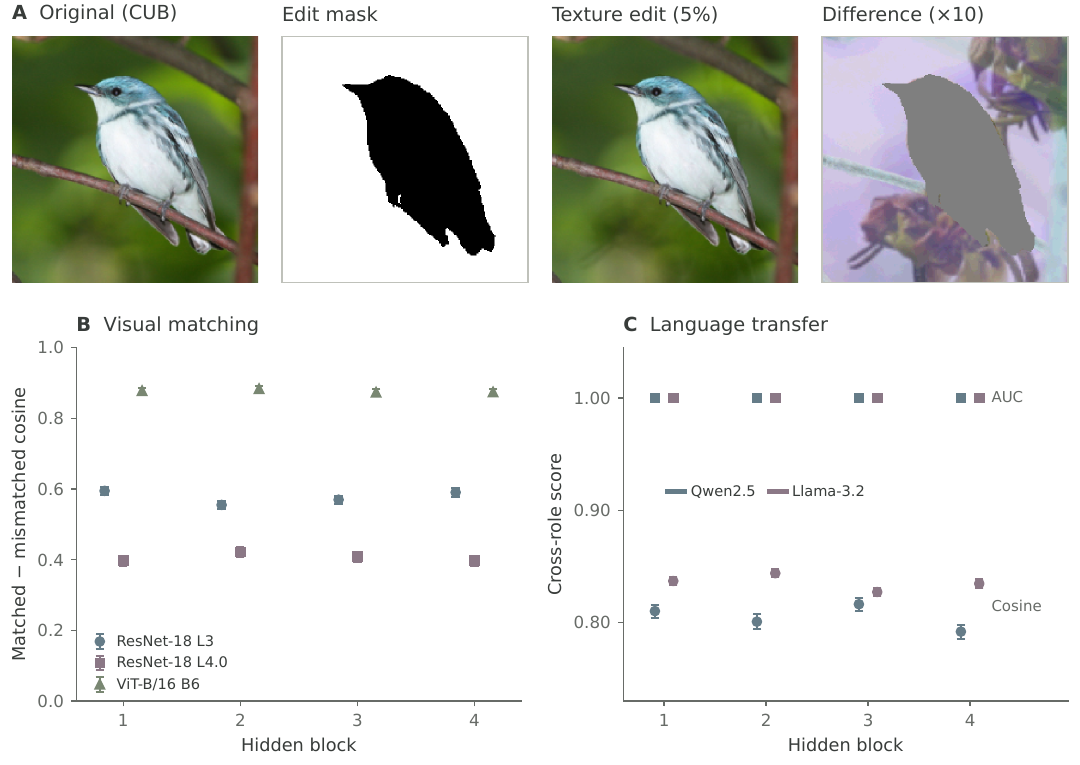}
  \caption{\textbf{Shared encoding across disjoint representation blocks in vision and language models.} \textbf{A,} Controlled visual background-texture edit on CUB-200-2011: original bird image, editable background mask (white: editable background; black: untouched foreground), $5\%$ natural-texture background edit, and $10\times$ amplified signed pixel difference centered at mid-grey (zero change across the entire bird silhouette; Pets LAC discovery fields appear in \cref{fig:s2b}). \textbf{B,} Native visual paired matching on $1{,}024$ held-out CUB images: after subtracting the development-set mean response for each texture contrast, each disjoint channel block and its complement independently estimate $16$D coordinates using exclusively their own raw channels (pre-fixed $16$D subspace covering $42.14\%$ of response energy at ResNet-18 \texttt{layer3}, $14.09\%$ at \texttt{layer4.0}, and $49.26\%$ at ViT-B/16 Block~6); same-image alignment strongly exceeds mismatched images with the same texture (bars: $99.375\%$ source-image bootstrap CIs). \textbf{C,} Native language cross-role transfer on $256$ held-out name pairs per model: all four disjoint hidden blocks in frozen Qwen2.5-1.5B (Block~14) and Llama-3.2-1B (Block~8) achieve within-name-pair, cross-role ranking $\mathrm{AUC} = 1.000$ (squares, ranking the two $a=b$ vs.\ two $a\neq b$ states within each pair) and high continuous cosine alignment (circles; bars: $95\%$ name-pair CIs).}
  \label{fig:fig2}
\end{figure}

\paragraph{Visual discovery anchor and why per-texture cross-image mean directions are insufficient.}
Single-channel LAC inversion of Pets ResNet-18 reveals extensive scene and background structure inside individual deep channels (\cref{fig:s2b}, top). To measure background encoding natively without inversion, we construct controlled background edits on CUB-200-2011 by blending natural textures strictly outside the bird segmentation mask at $5\%$ mixing weight, leaving every foreground pixel unchanged (\cref{fig:fig2}A). A natural first hypothesis is that shared encoding merely reflects a fixed cross-image mean response for each texture contrast. Estimating one frozen cross-image mean direction per texture pair (four directions across the four texture pairs) at ResNet-18 terminal global average pooling (GAP) and evaluating on $1{,}024$ held-out CUB images refutes this simple picture: while matched texture-pair alignment exceeds mismatched texture pairs ($0.0891$ versus $0.0020$, advantage $+0.0870\,[0.0840, 0.0901]$), each texture pair's cross-image mean direction explains on average only $1.52\%$ of that pair's held-out GAP response energy and predicts the sign of the class-margin change on only $54.86\%$ of trials. By contrast, the per-image native directional derivative achieves $73.47\%$ sign agreement, dropping to $50.50\%$ when mismatched to a different texture pair and to $49.77\%$ when paired with a reassigned readout (\cref{fig:s2a}).

\paragraph{Instance-grounded visual matching across disjoint blocks.}
Rather than being captured by the low-energy per-texture-pair cross-image mean directions at terminal GAP ($1.52\%$), the visual background response preserves coherent instance-specific structure across disjoint channel blocks. On the $1{,}024$ held-out CUB images, after \emph{subtracting the development-set mean response for each texture contrast}, each of four pre-fixed disjoint channel blocks uses \emph{exclusively its own raw channels} (via pseudoinverse onto the block-restricted $16$-dimensional development basis, with $16$ chosen as an analysis dimension rather than claiming intrinsic dimensionality) to estimate $16$-dimensional coordinates, while the complementary channels independently estimate coordinates using exclusively their own raw channels; we then test whether same-image block-to-complement cosine alignment exceeds mismatched images edited with the exact same texture (\cref{fig:fig2}B). Every block passes with large positive cosine advantages at all three tested visual cuts, with each cut's subspace energy coverage reported separately alongside the $1.52\%$ mean energy share of the per-texture-pair terminal GAP directions:
\begin{itemize}
  \item \textbf{ResNet-18 \texttt{layer3}:} same-minus-mismatched cosine advantages $+0.555$ to $+0.595$ across the four blocks ($99.375\%$ multiplicity-adjusted CIs strictly positive; $16$D subspace covers $42.14\%$ of response energy);
  \item \textbf{ResNet-18 \texttt{layer4.0}:} cosine advantages $+0.397$ to $+0.422$ ($14.09\%$ energy coverage);
  \item \textbf{ViT-B/16 Block~6:} cosine advantages $+0.874$ to $+0.884$ ($49.26\%$ energy coverage).
\end{itemize}
Even when each side sees only its own disjoint channels and the development-set texture-contrast mean is removed, block and complement recover strongly aligned instance-specific coordinates within the pre-fixed $16$D subspace. To probe the lower limit of factor strength, we also evaluate a weak warm-versus-cool localized background tint applied to $1{,}024$ selected background pixels in Pets ResNet-18 using a frozen linear probe per block: on $442$ newly eligible held-out images, the four disjoint blocks achieve $248, 269, 250,$ and $253$ correct predictions ($56.1\%\text{--}60.9\%$, joint one-sided $p = 0.0042$, whereas the smaller $64$-image exploratory batch had $p = 0.546$; \cref{fig:s2a}). Weak localized background tints are thus detectable at scale across disjoint blocks, though with far smaller effect size than structured natural textures.

\paragraph{Cross-role relational shared encoding in frozen language models.}
We apply the language operationalization of \cref{def:shared_encoding} to frozen Qwen2.5-1.5B (Block~14) and Llama-3.2-1B (Block~8) in a multi-entity attribute-binding task \citep{shu2026shared}. A relational binding direction is estimated under one query role on development names and evaluated under the opposite query role across four pre-fixed, non-overlapping hidden-dimension blocks on $256$ zero-overlap held-out name pairs (\cref{fig:fig2}C). Specifically, for each held-out name pair and held-out query role, we compute the \emph{within-name-pair, cross-role ranking AUC} comparing the two same-attribute states ($a=b$) against the two distinct-attribute states ($a\neq b$)---allowing pair-specific score baselines---and average across roles and name pairs. Every one of the eight blocks across both models achieves within-name-pair cross-role ranking $\mathrm{AUC} = 1.000$, with continuous cross-role cosine alignment ranging from $0.792$ to $0.816$ in Qwen2.5 and from $0.827$ to $0.844$ in Llama-3.2. Thus, across both vision and language architectures, controlled input factors remain recognizable across mutually disjoint blocks of hidden coordinates. Having established how factors are \emph{encoded}, we turn to how downstream readouts \emph{selectively use} them.

\subsection{Fixed-Task Channel Synthesis and the Two-Scale Bridge}
\label{sec:exp_channels}

If multiple channel blocks retain background structure, how does a fixed linear classifier suppress background while preserving foreground? We test the Type~I specialization of \cref{thm:synthesis} directly on Oxford-IIIT Pets ResNet-18 (512 final channels), fixing each source image's clean top-1 class and partitioning channels strictly by the sign of their class-head weights into positive and negative groups. Each source image receives four exclusive donor perturbations inside either the foreground or background mask (after summing donor energies per image).

\begin{figure}[t]
  \centering
  \includegraphics[width=\textwidth]{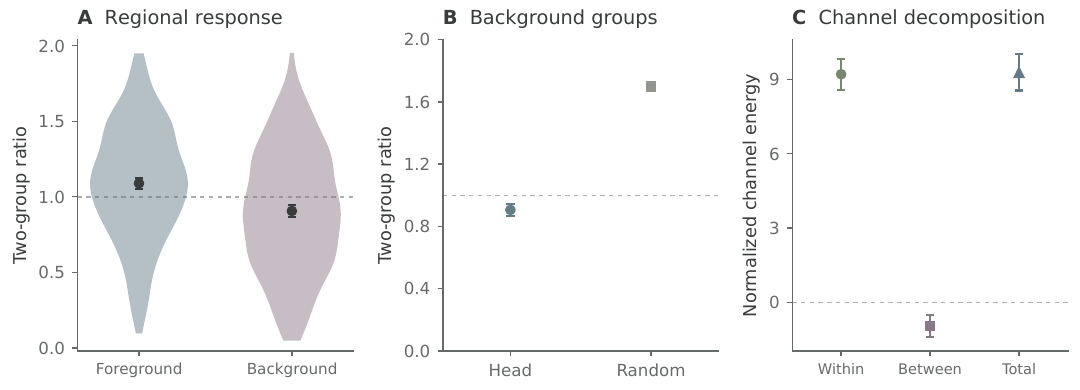}
  \caption{\textbf{Fixed-task channel synthesis across two grouping scales in Pets ResNet-18 ($444$ held-out images).} \textbf{A,} Distributions and image means ($95\%$ bootstrap CIs) of the two-group interference ratio $R^{(2)}$ under the fixed clean top-1 class-head sign partition: foreground edits reinforce ($1.089\,[1.054, 1.125]$) whereas background edits cancel ($0.906\,[0.868, 0.944]$). \textbf{B,} Holding the exact same $512$ signed background contributions fixed, the head-weight sign partition yields substantially stronger cancellation than size-matched random partitions (paired difference $-0.792\,[-0.821, -0.761]$). \textbf{C,} Exact $512$-channel decomposition on the same $444$ images ($R^{(512)} = 1 + 9.200 - 0.937 = 9.263$): $266/444$ images have positive within-sign and negative between-sign cross-terms, and $248/444$ individual images simultaneously satisfy $R^{(2)} < 1$ and $R^{(512)} > 1$ (net reinforcement across the $512$-channel decomposition).}
  \label{fig:fig3}
\end{figure}

\paragraph{Independent large-sample confirmation of between-group background cancellation.}
Across two independent confirmation cohorts, the two-group interference ratio $R^{(2)}$ cleanly separates background cancellation from foreground reinforcement (\cref{fig:fig3}A; cohort provenance including the earlier $16$, $48$, and $64$ exploratory batches with $R_{\mathrm{bg}}^{(2)} = 0.817, 0.839, 0.834$ is detailed in \cref{fig:s3a}):
\begin{itemize}
  \item \textbf{First fresh cohort ($128$ images):} background $R_{\mathrm{bg}}^{(2)} = 0.858\,[0.787, 0.929]$ ($81/128 < 1$), foreground $R_{\mathrm{fg}}^{(2)} = 1.122\,[1.058, 1.189]$;
  \item \textbf{Expanded fresh cohort ($444$ images):} background $R_{\mathrm{bg}}^{(2)} = 0.906\,[0.868, 0.944]$ ($266/444 < 1$), foreground $R_{\mathrm{fg}}^{(2)} = 1.089\,[1.054, 1.125]$.
\end{itemize}
Switching the fixed readout from clean top-1 logit to the fixed top-1 minus top-2 margin preserves background cancellation ($R_{\mathrm{bg}}^{(2)} = 0.879$ on the $444$ images), and native directional derivatives closely track real finite-dose logit changes (Spearman correlation $0.871$ at full RGB dose $0.125$ and $0.936$ at half dose $0.0625$; \cref{fig:s3a}).

\paragraph{Attribution to weight-sign grouping and same-image two-scale coexistence.}
To verify that background cancellation reflects the actual head-weight sign partition rather than arbitrary two-group averaging, we hold the exact same $512$ signed channel contributions fixed on every image and compare the true head-weight partition against $1{,}024$ size-matched random two-group partitions (\cref{fig:fig3}B). On the $444$ images, the true sign partition lowers background $R^{(2)}$ relative to random partitions by $-0.792\,[-0.821, -0.761]$ ($-0.816$ on the $128$-image cohort), isolating the suppression precisely to the learned weight-sign membership.

At the same time, evaluating all $512$ individual channels on the exact same $444$ background perturbations yields a fine-scale interference ratio far above one: $R^{(512)} = 1 + 9.200 - 0.937 = 9.263$ (\cref{fig:fig3}C), where positive within-sign synergy ($+9.200$) outweighs negative between-sign cancellation ($-0.937$). By the exact two-scale bridge in \cref{eq:two_scale}, these two measurements are algebraically consistent image by image. Specifically, among the $444$ confirmation images, $266/444$ images exhibit positive within-group cross-terms together with negative between-group cross-terms (equivalently $R^{(2)} < 1$), and \textbf{$248/444$ individual images simultaneously satisfy both $R^{(2)} < 1$ (between-group cancellation) and $R^{(512)} > 1$ (net reinforcement across the $512$-channel decomposition, $426/444$)}. Within-group cooperation and between-group cancellation therefore coexist natively on the same input. We next ask how this signed synthesis extends when the downstream readout traverses multiple nonlinear blocks and architectural pathways.

%% file: sections/04b_paths_conditions.tex
\subsection{Fixed-Task Multi-Pathway Geometry and Coarse-to-Fine Grouping Scale}
\label{sec:exp_paths}

\begin{figure}[t]
  \centering
  \includegraphics[width=\textwidth]{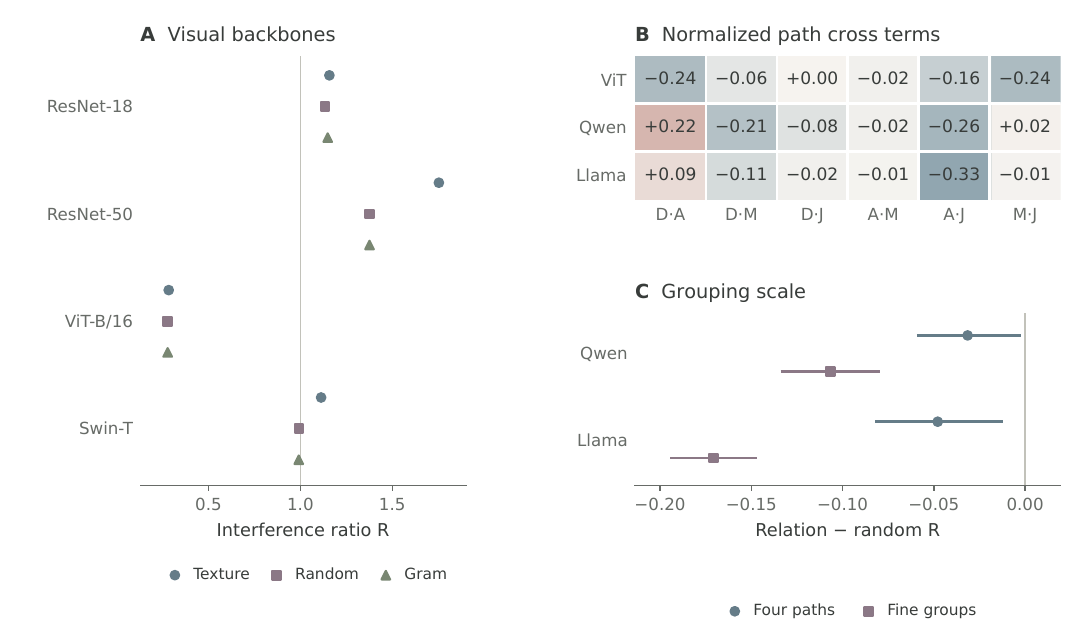}
  \caption{\textbf{Multi-pathway interference geometry and coarse-to-fine grouping scale.} \textbf{A,} Mean directional interference ratio $R$ across four frozen visual backbones on the same $1{,}024$ CUB images for measured texture edits, matched-norm random directions, and exact path-gradient Gram integrals (\cref{eq:gram_integral}); for the measured texture direction, ResNet-18/50 and Swin-T ($1.113$, whereas Swin-T's random baseline lies just below $1$ at $0.992$) lie above $1$ while ViT-B/16 exhibits strong cancellation ($0.284$). \textbf{B,} Mean normalized pairwise cross-terms across ViT-B/16 ($1{,}024$ images), Qwen2.5-1.5B, and Llama-3.2-1B ($256$ confirmation name pairs each): both $A\cdot J$ and $D\cdot M$ are negative in all three Transformers. \textbf{C,} Coarse four-path versus fine functional-MLP decomposition at identical total directional derivative on $256$ additional zero-overlap name pairs: isolating development-selected MLP units deepens relation-minus-random $\Delta R$ by $+0.0753$ in Qwen2.5 and $+0.1227$ in Llama-3.2 ($95\%$ paired name-level CIs).}
  \label{fig:fig4}
\end{figure}

Moving from a single linear head (Type~I) to intermediate nonlinear cuts (Type~II/III), we decompose the downstream dual covector into four complete tangent-space computational pathways: residual/convolutional paths $(SS, SF, FS, FF)$ in ResNets and bypass, attention-only, MLP-only, and mixed paths $(D, A, M, J)$ in Transformers (\cref{app:thm1}). In the CUB visual pathway experiments (\cref{fig:fig4}A), we first apply the $5\%$ background-only pixel texture edits to estimate the channel-mean response difference at cut $T$, project that difference onto the cut's frozen $16$-dimensional development subspace, normalize the resulting channel vector to unit $L_2$ norm, and inject it into the clean image's hidden cut (broadcast uniformly across spatial locations in ResNets and Swin-T, and added as the same $768$-dimensional vector across all $197$ tokens in ViT-B/16); matched random directions are drawn in the exact same channel space. These experiments therefore measure how downstream pathways read a hidden-layer channel direction estimated from background texture edits.

\paragraph{Path-gradient Gram baselines and architectural diversity across four visual backbones.}
On the same $1{,}024$ held-out CUB images (\cref{fig:fig4}A), the path-gradient Gram integral (\cref{eq:gram_integral}) predicts the isotropic Gaussian random-direction interference baseline of every tested visual cut, while exposing sharp architectural divergence:
\begin{itemize}
  \item \textbf{ResNet-18 (\texttt{layer3}):} texture $R = 1.158$ versus frozen random-direction bank $R = 1.133$ (analytic Gram integral $1.1488$; paired image-wise integral-minus-bank difference $+0.0156\,[0.0119, 0.0193]$, reflecting the finite random-direction bank while both remain strongly constructive); gradient energy is heavily concentrated in two aligned convolutional paths ($FF$ $90.6\%$, $SF$ $9.3\%$).
  \item \textbf{ResNet-50:} texture $R = 1.754$ versus random $R = 1.377$ (texture minus random $+0.376\,[0.348, 0.405]$).
  \item \textbf{Swin-T:} texture $R = 1.113$ versus random $R = 0.992\,[0.986, 0.998]$ (texture minus random $+0.121\,[0.104, 0.138]$). This establishes an essential boundary: \emph{Transformers do not universally cancel texture perturbations}.
  \item \textbf{ViT-B/16 (Block~6):} texture $R = 0.284\,[0.272, 0.295]$ versus random $R = 0.277\,[0.274, 0.280]$ (Gram integral $0.2779$); all four paths carry balanced energy with strong opposition between the mixed path $J$ and the remaining paths (confirmed via native forward clamping, yielding a normalized mixed-versus-other cross-term rounding to $-0.396$---95\% CI $[-0.408, -0.384]$---at full dose, half dose, and first order).
\end{itemize}
Within the same frozen ViT-B/16, two further controls verify the taxonomy in \cref{tab:cut_taxonomy}. Editing non-\texttt{[CLS]} patches at the terminal layer under the \texttt{[CLS]}-only head produces exact zero margin gradient and exact zero finite logit change across all $1{,}024$ images (structural Type~0). At Block~6, regrouping the exact same total derivative from attention/MLP modules into early-three versus late-three depth blocks shifts texture $R$ from $0.284$ to $0.409$ (random $0.277 \to 0.331$, Gram $0.332$; \cref{fig:s4_regroup}), confirming that $\Rint$ depends on the chosen grouping boundary even at fixed total sensitivity. In addition, while both measured texture and matched random directions exhibit strong net cancellation in ViT-B/16, any excess cancellation for texture relative to random is unresolved: texture minus random $R$ is $+0.006\,[-0.004, +0.017]$ on the $1{,}024$ images ($4{,}094$ of $4{,}096$ image--readout combinations span both $R<1$ and $R>1$ among random directions; \cref{fig:s4_fine}).

\paragraph{Recurring negative pathway cross-terms across vision and language Transformers.}
Inspecting all six normalized pairwise cross-terms ($\sum_{i<j} 2t_i t_j / \Eincoh = R - 1$) across ViT-B/16 ($1{,}024$ CUB images), Qwen2.5-1.5B ($256$ confirmation name pairs), and Llama-3.2-1B ($256$ confirmation name pairs) reveals two consistent cross-modal negative interactions (\cref{fig:fig4}B):
\begin{itemize}
  \item \textbf{Attention-only $\times$ mixed writes ($A \times J$)} is strongly negative in all three models: $-0.157\,[-0.170, -0.145]$ in ViT-B/16, $-0.262\,[-0.284, -0.240]$ in Qwen2.5, and $-0.332\,[-0.350, -0.315]$ in Llama-3.2.
  \item \textbf{Bypass $\times$ MLP-only writes ($D \times M$)} is likewise negative across all three models: $-0.061\,[-0.067, -0.054]$ in ViT-B/16, $-0.213\,[-0.233, -0.194]$ in Qwen2.5, and $-0.108\,[-0.126, -0.090]$ in Llama-3.2.
\end{itemize}
In contrast, $A \times M$ is weakly negative in ViT-B/16 ($-0.020$) and Qwen2.5 ($-0.023$) and crosses zero in Llama-3.2 ($-0.006\,[-0.018, +0.005]$; full six-pair table in \cref{tab:s4_cross}).

\paragraph{Coarse pathway aggregation dilutes observable relation cancellation.}
In the language pathway tests, each name pair's own relational interaction is projected onto the frozen development subspace and injected at the attention-module output at the query token (\cref{tab:protocols}). Under the coarse four-path decomposition $(D, A, M, J)$, both language models exhibit net cancellation overall ($R < 1$), yet the extra cancellation of the relation direction relative to matched-norm random directions ($\Delta R := R_{\mathrm{rel}} - R_{\mathrm{rand}}$) is modest or unresolved: in the first $256$-pair cohort, $\Delta R = -0.0192\,[-0.0545, +0.0162]$ in Qwen2.5 and $-0.0700\,[-0.1041, -0.0360]$ in Llama-3.2; in the additional $256$-pair cohort, $\Delta R = -0.0315\,[-0.0592, -0.0025]$ in Qwen2.5 and $-0.0479\,[-0.0822, -0.0123]$ in Llama-3.2.

Guided by the grouping-scale principle (\cref{eq:two_scale}), we tested whether merging opposing MLP neurons into the single coarse $M$ and $J$ pathways diluted $\Delta R$ (designed after observing the coarse four-path result and confirmed independently on the $256$ additional zero-overlap name pairs). Keeping the exact same inputs, edits, and total directional derivative untouched on every trial, we split the last-layer MLP (Qwen Layer~28, $1{,}120/8{,}960$ units; Llama Layer~16, $1{,}024/8{,}192$ units) into development-selected functional units, remaining same-layer units, and layer-bypassing paths (\cref{fig:fig4}C). Separating functional units sharply deepens relation-minus-random cancellation in both models:
\begin{itemize}
  \item \textbf{Qwen2.5-1.5B (new $256$ pairs):} $\Delta R$ deepens from $-0.0315$ to $-0.1068\,[-0.1339, -0.0794]$, giving a paired coarse-minus-fine contrast $B = +0.0753\,[0.0564, 0.0948]$ (closely matching $+0.0751$ on the replay cohort). The largest single cross-term contributing to $B$ is the intra-MLP interaction between selected units and remaining same-layer units ($+0.0836\,[0.0668, 0.1018]$). Relative to tight $\pm 1\%$ strength-matched random controls, the functional group's overall coarse-minus-fine contrast $B$ retains a paired advantage of $+0.0699\,[0.0563, 0.0851]$.
  \item \textbf{Llama-3.2-1B (new $256$ pairs):} $\Delta R$ deepens from $-0.0479$ to $-0.1706\,[-0.1943, -0.1471]$, with $B = +0.1227\,[0.0876, 0.1595]$ ($+0.1124$ on the replay cohort). Again, the dominant component is the intra-$M$ selected-versus-remaining cross-term ($+0.1039\,[0.0786, 0.1304]$, with the corresponding $J$-path cross-term adding $+0.0376\,[0.0256, 0.0499]$), and the functional group's overall $B$ exceeds tight strength-matched controls by $+0.1372\,[0.1061, 0.1702]$.
\end{itemize}
Collapsing oppositely acting units into a single coarse pathway therefore masks substantial relation-specific cancellation inside the MLP layer. Having dissected fixed-task readouts across channels, paths, and scales, we now invert the experimental constraint: hold the factor fixed and vary the readout condition.

\subsection{Varying Readout Conditions Under a Fixed Factor}
\label{sec:exp_conditions}

If selectivity arises from conditional signed readout rather than hard-coded upstream erasure, then fixing the exact same hidden perturbation and changing only the readout condition $q$ should reshape the factor's gain or even reverse its sign.

\begin{figure}[t]
  \centering
  \includegraphics[width=\textwidth]{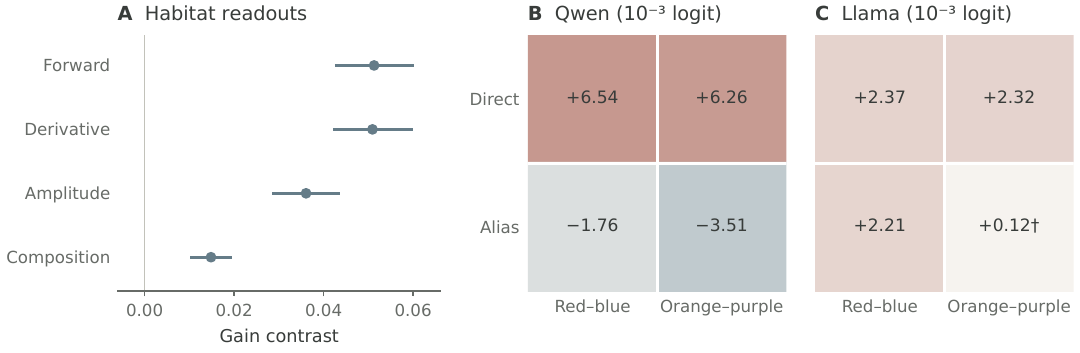}
  \caption{\textbf{Conditional readout shifts under a fixed factor edit.} \textbf{A,} Visual readout switch on $1{,}024$ CUB images (ResNet-18 \texttt{layer3}): re-pairing the exact same eight bird classes from within-habitat to cross-habitat contrasts produces a positive background-minus-random gain contrast ($+0.05127$ finite forward; $+0.05091$ first-order derivative) that decomposes symmetrically into both amplitude ($+0.03608$) and signed-composition ($+0.01484$) contributions ($95\%$ image-level CIs). \textbf{B--C,} Language $2\times 2$ factorial design ($52$ shared dual-tokenizer name pairs, paired injection $0.5ab\,v$ identical across Direct and Alias for each $(a,b,r)$ tuple, mean selection contrast $\Delta G$ in $10^{-3}$ logit, shared zero-centered scale): under fixed \emph{Red--blue} answer words, switching Direct to Alias flips the mean finite contrast in frozen Qwen2.5 ($+6.54 \to -1.76$) while frozen Llama-3.2 exhibits a strong presentation $\times$ vocabulary interaction ($+2.050$; dagger $\dagger$ marks the sole cell whose $95\%$ CI includes zero: $+0.12\,[-0.39, +0.64]$).}
  \label{fig:fig5}
\end{figure}

\paragraph{Visual readout re-pairing and amplitude--composition decomposition.}
On the $1{,}024$ held-out CUB images, using the exact same hidden-layer injection protocol as in \cref{sec:exp_paths}, we fix the clean image, ResNet-18 \texttt{layer3} cut, subspace-projected unit-normalized background channel vector (broadcast across spatial locations), edit dose, and the \emph{exact same eight bird species}, changing only how those eight classes are paired into class-margin contrasts: within-habitat versus cross-habitat pairs (each class appearing exactly once in both sets). Subtracting the corresponding cross-minus-within difference of matched-norm random directions isolates the readout-conditioned background shift (\cref{fig:fig5}A). In real finite forward passes, re-pairing the eight classes shifts the absolute background margin gain by a difference-in-differences (DiD) of $+0.05127\,[0.04250, 0.06016]$ logit per unit channel-$L_2$ dose (half dose: $+0.05097\,[0.04215, 0.05989]$). Applying the exact symmetric product decomposition of $|\tau| = \sqrt{\Eincoh}\,\sqrt{\Rint}$ (\cref{eq:factorization}, \cref{app:thm3}) to the first-order total DiD ($+0.05091$) cleanly resolves \emph{both} components at large sample size: the path-amplitude term contributes $+0.03608$ and the signed-composition term contributes $+0.01484$, with amplitude significantly larger than composition (paired difference $+0.02124\,[0.01238, 0.03001]$). Switching the visual class contrast thus modulates both the magnitude and the signed phase synthesis of the identical background perturbation.

\paragraph{Prompt-conditioned sign reversal and vocabulary interaction in frozen LLMs.}
In language models, we freeze the pretrained weights and a single cross-name relation axis $v$, injecting $0.5ab\,v$ at the query-token attention-module output where $a, b \in \{-1, +1\}$ encode the two entities' attributes and $r \in \{-1, +1\}$ specifies the query role. For any matched $(a, b, r)$ tuple, Direct and Alias prompts receive the exact same injection. Writing $m_{\mathrm{edit}}$ and $m_{\mathrm{clean}}$ for the signed answer-pair logit difference (blue minus red, or purple minus orange), each name pair's finite selection contrast is defined as $\Delta G := \operatorname{mean}_{a,b,r}\big[(a - b)r(m_{\mathrm{edit}} - m_{\mathrm{clean}})\big]$. We execute a $2\times 2$ factorial design crossing prompt presentation (\emph{Direct} versus \emph{Alias}) and candidate answer pair (\emph{Red--blue} versus \emph{Orange--purple}) on the $52$ zero-overlap name pairs that remain single tokens in both tokenizers (\cref{fig:fig5}B--C; mean $\Delta G$ across name pairs in $10^{-3}$ logit):
\begin{itemize}
  \item \textbf{Qwen2.5-1.5B:} Under the fixed \emph{Red--blue} answer pair, switching prompt presentation from Direct to Alias reverses the sign of the mean finite selection contrast from $+6.540\,[5.066, 8.031]$ to $-1.756\,[-2.586, -0.878]$ (and similarly from $+6.263$ to $-3.513$ under \emph{Orange--purple}). Both main effects and the interaction are cleanly resolved: presentation $-9.035\,[-10.873, -7.250]$, vocabulary $+1.017\,[0.242, 1.770]$, and interaction $+1.480\,[0.106, 2.872]$.
  \item \textbf{Llama-3.2-1B:} Under \emph{Red--blue}, both presentations remain positive ($+2.368\,[1.311, 3.432]$ in Direct and $+2.214\,[1.476, 2.915]$ in Alias), whereas under \emph{Orange--purple} the mean contrast drops from $+2.321$ to $+0.117\,[-0.394, +0.641]$ (crossing zero). Across the $2\times 2$ design, the presentation main effect ($-1.179\,[-2.235, -0.151]$), vocabulary main effect ($+1.072\,[0.646, 1.495]$), and presentation $\times$ vocabulary interaction ($+2.050\,[1.192, 2.914]$) are statistically resolved.
\end{itemize}
Crucially, two-choice candidate accuracy between the paired answer words confirms that the sign reversal in Qwen2.5 does not reflect task collapse under Alias prompts: Direct two-choice accuracy is $77.2\%\text{--}78.8\%$ in Qwen2.5 and $90.4\%\text{--}91.1\%$ in Llama-3.2, while on the core \emph{Red--blue} Alias condition two-choice accuracy remains high at $81.7\%$ in Qwen2.5 and $79.8\%$ in Llama-3.2 (the lower $67.1\%$ and $63.2\%$ accuracies occur only on Alias \emph{Orange--purple}; \cref{fig:s5_llm}). Because the prompt switch jointly alters wording, sequence length, and query position under a finite $0.5ab\,v$ injection, we report the measured finite reversal directly rather than attributing it exclusively to a single-point local Jacobian. Together, both models confirm conditional readout on frozen weights---manifesting as an explicit finite sign flip in Qwen2.5 and vocabulary-dependent gain modulation in Llama-3.2.

%% file: sections/04c_gating_pruning.tex
\subsection{Prospective Gating and Held-Out Tests of Conditional Readout}
\label{sec:exp_gating}

The analyses in \cref{sec:exp_channels,sec:exp_paths,sec:exp_conditions} characterize signed readout inside fixed and switched tasks. We now test whether this mechanism yields accurate, counterintuitive \emph{out-of-sample} predictions when we actively intervene on frozen unit groups during real forward passes. In all primary gating tests, candidate unit groups are selected solely on the development split, frozen before opening the held-out evaluation split, and then evaluated via real forward passes applying $\lambda_0 = 0.1$ ($10\%$) attenuation exclusively to their edit-minus-clean activation increments (preserving the clean operating point; companion physical output scaling, six-dose curves, and drift checks appear in \cref{supp:s6}). Specifically, ViT-B/16 and language attention-head groups are selected directly via development-set local gating slopes $\kappa_G$ for the given edits, whereas last-layer MLP functional groups in Qwen2.5 and Llama-3.2 are selected via first-order local scores evaluated at the clean working point; in both cases, all reported evaluation metrics come from actual finite $10\%$ attenuation forward passes, separating same-cohort \emph{replay} (holding weights, layer, and prior test inputs fixed while changing only unit membership) from independent held-out \emph{confirmation} on fresh samples. Positive $H_G$ (\cref{eq:gating_H}) indicates that weakening the group \emph{amplifies} the edit's decision effect (true antagonism); negative $H_G$ indicates promotion. Because the experiments span original prospective tests, same-weight follow-up member replacements, and sensitivity controls, we explicitly mark the provenance and matching protocol of each comparison.

\begin{figure}[t]
  \centering
  \includegraphics[width=\textwidth]{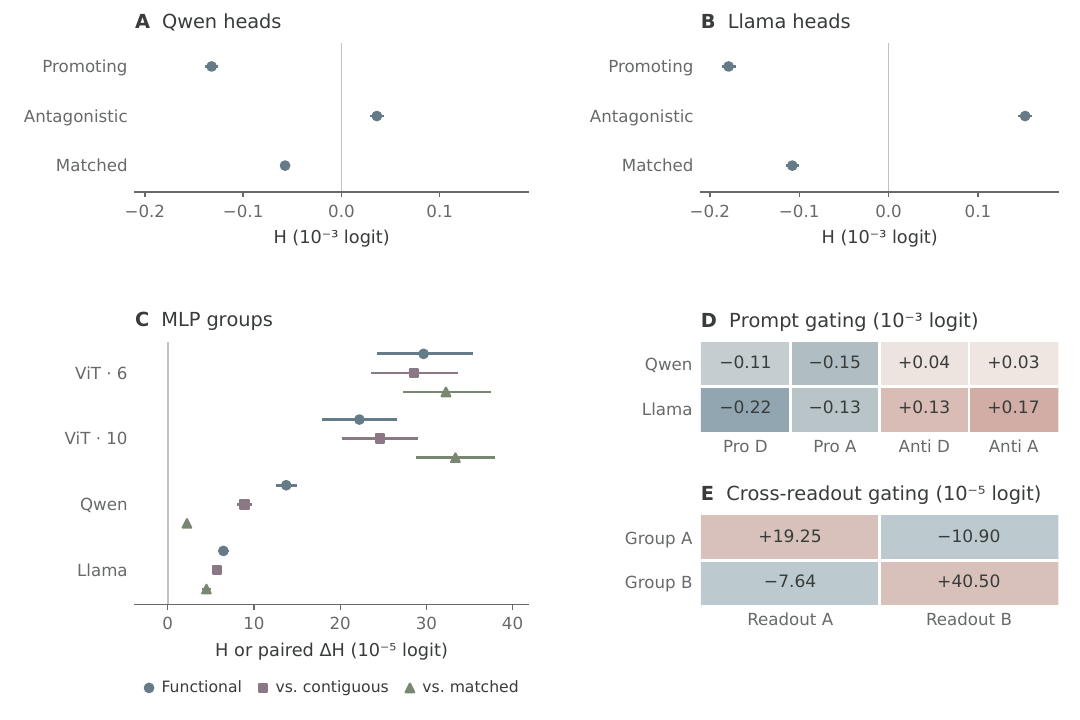}
  \caption{\textbf{Prospective gating and same-forward-pass cross-readout sign reversal.} \textbf{A--B,} Bidirectional attention-head gating in frozen Qwen2.5-1.5B and Llama-3.2-1B ($256$ held-out name pairs, $10\%$ edit-increment attenuation, $10^{-3}$ logit): weakening promoting heads reduces $|\Delta m|$ ($H < 0$), whereas weakening antagonistic heads amplifies $|\Delta m|$ ($H > 0$) and exceeds strength-matched random heads. \textbf{C,} Functional MLP groups ($1/8$-layer budget, $10^{-5}$ logit) and their paired advantages over best contiguous blocks and strength-matched controls in original ImageNet ViT-B/16 ($264$ reserve images, ViT matching protocol; $1{,}024$-image Block~6 replay yields $+30.41$ vs.\ contiguous in \cref{supp:s6}) and Qwen2.5/Llama-3.2 ($256$ additional shared name pairs, tight $\pm 1\%$ matching: post-hoc sensitivity in Qwen, pre-frozen in Llama). \textbf{D,} Bidirectional head gating stratified by Direct (D) and Alias (A) prompts ($10^{-3}$ logit). \textbf{E,} Same-forward-pass cross-readout matrix of held-out means in ImageNet ViT-B/16 Block~10 ($264$ reserve images, $10^{-5}$ logit): both margins are read from the same gated output for each group; attenuating Group~A increases the mean effect on Readout~A ($+19.25$) and decreases it on Readout~B ($-10.90$), while Group~B reverses the pattern ($-7.64$ vs.\ $+40.50$), yielding a four-cell paired group-by-readout interaction of $+78.30\,[64.73, 91.92]$ (fill tint encodes sign and printed entries give magnitude).}
  \label{fig:fig6}
\end{figure}

\paragraph{Bidirectional attention-head gating in frozen language models.}
Using $16$ development name pairs to lock two-head promoting, antagonistic, and strength-matched random groups, we evaluate real forward gating on $256$ zero-overlap confirmation name pairs (\cref{fig:fig6}A--B; units $10^{-3}$ logit). Both frozen models satisfy all five joint criteria---exceeding the unmatched $128$-group random strength bank for both selected groups, decreasing $|\Delta m|$ under promoting attenuation, increasing $|\Delta m|$ under antagonistic attenuation, and exceeding strength-matched random controls:
\begin{itemize}
  \item \textbf{Qwen2.5-1.5B:} promoting $H = -0.132\,[-0.139, -0.125]$, antagonistic $H = +0.036\,[+0.029, +0.043]$, and antagonistic minus strength-matched random $+0.093\,[+0.084, +0.103]$.
  \item \textbf{Llama-3.2-1B:} promoting $H = -0.179\,[-0.187, -0.171]$, antagonistic $H = +0.153\,[+0.145, +0.161]$, and antagonistic minus strength-matched random $+0.261\,[+0.247, +0.275]$.
\end{itemize}
Separating Direct and Alias prompts preserves both signs in every cell (\cref{fig:fig6}D). Moreover, replacing synthetic hidden vector edits with real lexical relation word swaps preserves the bidirectional sign structure at much larger scale (Qwen2.5: $-43.49 / +17.61 \times 10^{-3}$ logit; Llama-3.2: $-58.66 / +70.42 \times 10^{-3}$ logit; \cref{fig:s6_llm_dose}).

\paragraph{Functional MLP grouping resolves contiguous-index dilution across vision and language.}
In our initial prospective ViT-B/16 test, candidate groups were constrained to $48$ contiguous channel-index sub-blocks ($384/3{,}072$ MLP units per block across Blocks~6--11). Whereas an initial $32$-image pilot selection yielded a contiguous Block~6 candidate, re-running contiguous development selection on $512$ CUB images found $0/24$ strength-eligible positive contiguous candidates (\cref{fig:s6_drift_dev}), and the original Block~6 contiguous antagonist remained unresolved when evaluated on the $1{,}024$ confirmation images ($H = +0.46\,[-0.34, +1.24] \times 10^{-5}$ logit, even though the promoting group passed at $-6.80\,[-8.13, -5.55]$, antagonistic-minus-random passed at $+6.13\,[+4.48, +7.81]$, and companion physical scaling passed all three criteria at $-9.88 / +1.70 / +4.67$).

To isolate whether contiguous indexing itself diluted antagonism, we performed a controlled member-replacement test across original ImageNet ViT-B/16, Qwen2.5-1.5B, and Llama-3.2-1B under a uniform $1/8$-layer MLP budget and $10\%$ increment attenuation (\cref{fig:fig6}C; units $10^{-5}$ logit):
\begin{itemize}
  \item \textbf{Original ImageNet ViT-B/16 (same-weight replay on $1{,}024$ images + held-out confirmation on $264$ reserve images):} Freezing the exact ImageNet-1K checkpoint, layer, $384$-unit budget, four bird class margins, and $1{,}024$ confirmation images from the unresolved Block~6 contiguous test (with functional membership fitted on the separate $512$-image development split), replacing \emph{only} contiguous membership with $384$ scattered functional units (evenly spread across all eight index segments: $49, 44, 50, 45, 55, 48, 49, 44$) raises held-out antagonism from $+0.46$ to $+30.86\,[28.09, 33.68]$ (paired advantage over best contiguous: $+30.41\,[27.77, 33.06]$). On the separate $264$-image reserve independent of both the $512$-image development split and the $1{,}024$-image replay split (also shared with a CUB-finetuned ViT-B/32 comparison in \cref{supp:s6}), the primary Block~10 functional group (segment counts $41, 56, 58, 46, 41, 53, 46, 43$) achieves $H = +22.23\,[17.84, 26.57]$, decisively beating best contiguous ($-2.39\,[-4.63, -0.30]$; paired $+24.62\,[20.21, 29.04]$) and ViT strength-matched random controls ($-11.13$; paired $+33.36\,[28.75, 38.01]$), while Block~6 achieves $+29.68\,[24.25, 35.44]$ ($+28.57\,[23.58, 33.70]$ over best contiguous; $+32.27\,[27.34, 37.46]$ over strength-matched).
  \item \textbf{Frozen Qwen2.5 (Layer~28, $1{,}120/8{,}960$) and Llama-3.2 (Layer~16, $1{,}024/8{,}192$) on $256$ additional shared name pairs:} Functional MLP groups achieve $H = +13.74\,[12.54, 15.02]$ (Qwen2.5) and $+6.46\,[5.83, 7.10]$ (Llama-3.2), outperforming best contiguous blocks by $+8.88\,[8.00, 9.80]$ and $+5.71\,[5.12, 6.32]$, respectively. Because coarse $64$-bin strength stratification still left a $16.5\%$ mean development-strength gap in Qwen2.5 (narrowing the paired advantage from $+8.47$ once controlled), we evaluated $32$ tight $\pm 1\%$ strength-matched random controls (actual development strength $99.07\%\text{--}100.98\%$ of the functional group; conducted as a post-hoc sensitivity analysis in Qwen2.5 and locked prior to holdout evaluation in Llama-3.2). Functional selection retains clear positive advantages over tight strength-matched controls in both models: $+2.22\,[1.93, 2.54]$ in Qwen2.5 ($+2.72$ on replay) and $+4.49\,[3.94, 5.05]$ in Llama-3.2 ($+4.92$ on replay), alongside positive group-direction $\times$ edit-direction interactions ($+0.66\,[0.38, 0.97]$ in Qwen2.5; $+2.77\,[2.15, 3.43]$ in Llama-3.2).
\end{itemize}

\paragraph{Same-unit, same-forward-pass cross-readout sign flip.}
Finally, we test the sharpest causal consequence of conditional signed readout in original ImageNet ViT-B/16 (Block~10, $264$ reserve images; \cref{fig:fig6}E, units $10^{-5}$ logit). We lock two single-task functional groups of $384$ MLP units on the development split: Group~A selected for class margin~A (\emph{American coot} minus \emph{brambling}) and Group~B selected for class margin~B (\emph{black swan} minus \emph{ruffed grouse}). For each held-out image, texture edit, and gated group, both class margins are read from the exact same output logit vector (averaging first across the four texture edits per image and then across the $264$ held-out images):
\begin{itemize}
  \item Attenuating frozen \textbf{Group~A} increases the \emph{mean} absolute background-edit effect on Readout~A ($H = +19.25\,[8.66, 29.72]$) while decreasing it on Readout~B ($H = -10.90\,[-15.86, -6.24]$; individually, $97/264$ images exhibit simultaneous positive Readout~A and negative Readout~B shifts).
  \item Attenuating frozen \textbf{Group~B} shows the opposite pattern, decreasing the mean effect on Readout~A ($H = -7.64\,[-11.99, -3.34]$) while increasing it on Readout~B ($H = +40.50\,[28.96, 51.37]$; $115/264$ images individually exhibit simultaneous negative Readout~A and positive Readout~B shifts).
\end{itemize}
Combining all four cells across Groups~A and~B yields a paired group-by-readout interaction of $+78.30\,[64.73, 91.92] \times 10^{-5}$ logit (replicated at even larger scale, $+352.59\,[319.71, 386.80]$, in CUB-finetuned ViT-B/32; \cref{fig:s6_functional}). Because both class margins are extracted from the exact same gated forward pass for a fixed image, background edit, and frozen unit group, this experiment directly establishes the readout dependence of unit roles (complementing the encoding evidence in \cref{sec:exp_encoding}): \emph{whether a group of units promotes or antagonizes a factor is not an intrinsic label of the units---it is defined only jointly with the target readout.}

\subsection{Quantitative Prediction of Channel Recovery in Structured Pruning}
\label{sec:exp_pruning}

We now invert the intervention from attenuating antagonistic units to restoring a single removed channel under a fixed linear successor, testing \cref{thm:fwl} (the FWL three-factor identity) across $10$ cuts spanning ResNet-50, ViT-S/16, and Swin-T on $4{,}096$ held-out ImageNet images. At each cut, we freeze the retained channel subset $S$ and optimal least-squares weight folding on development activations, restore each omitted candidate channel $j \notin S$ individually on the held-out images, and compare the measured linear-successor MSE drop against two frozen development predictors: conditional residual variance $v_j$ alone (encoding irreplaceability) versus the full FWL score incorporating fixed downstream readout coupling (\cref{eq:fwl_identity}).

\begin{figure}[t]
  \centering
  \includegraphics[width=\textwidth]{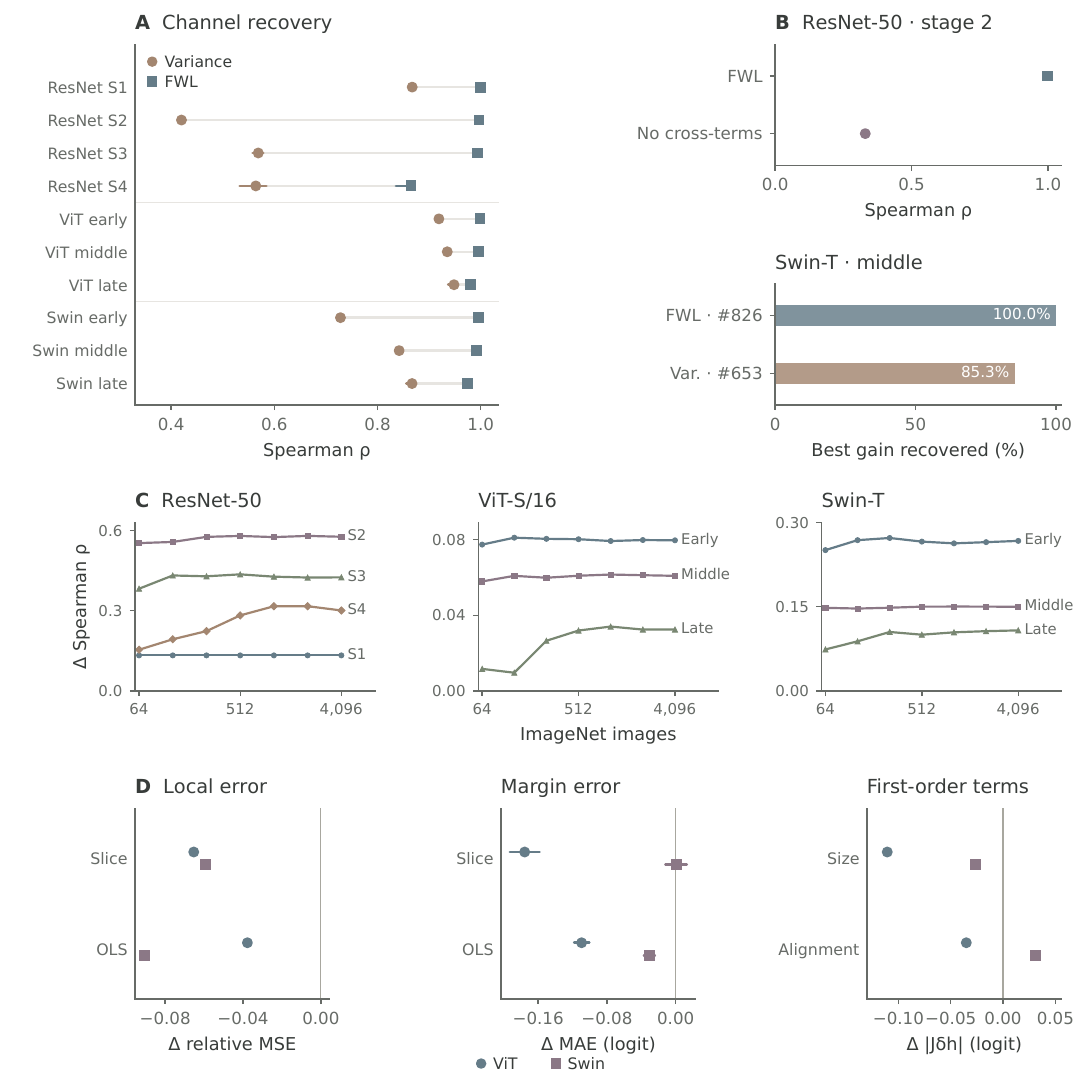}
  \caption{\textbf{Quantitative prediction of single-channel recovery in structured pruning ($4{,}096$ ImageNet images).} \textbf{A,} Held-out candidate recovery rank correlations across all $10$ cuts of ResNet-50, ViT-S/16, and Swin-T: adding fixed downstream readout coupling (FWL, squares) consistently outperforms conditional residual variance alone (circles), reaching $0.865\text{--}1.000$ overall and $0.975\text{--}0.999$ across the six ViT/Swin cuts ($97.5\%/98.75\%$ family-adjusted CIs). \textbf{B,} Signed cross-term ablation in ResNet-50 Stage~2 ($0.997 \to 0.330$, top) and top-candidate switch in Swin-T middle MLP from variance-selected $\#653$ ($85.3\%$) to FWL-selected $\#826$ ($100.0\%$, bottom). \textbf{C,} Stability of the FWL-minus-variance Spearman gain across nested ImageNet subsets ($64$ to $4{,}096$ images). \textbf{D,} Middle-cut pre-to-post-GELU shifts in relative local MSE, class-margin MAE, and symmetric first-order perturbation-size versus gradient-alignment decomposition under direct slicing (explaining why Swin-T's $-0.0266$ size drop and $+0.0310$ alignment increase cancel under direct slicing, whereas under OLS folding post-GELU selection lowers margin MAE relative to pre-GELU selection in both models).}
  \label{fig:fig7}
\end{figure}

\paragraph{Cross-architecture ranking confirmation across 10 cuts.}
Incorporating fixed downstream readout coupling improves held-out candidate recovery ranking at every single one of the $10$ cuts (\cref{fig:fig7}A,C; full table in \cref{tab:s7_ten_cuts}), achieving Spearman rank correlations of $0.865\text{--}1.000$ across all $10$ cuts and $0.975\text{--}0.999$ across the six Vision Transformer cuts:
\begin{itemize}
  \item \textbf{ResNet-50 (four bottleneck cuts \texttt{layer1.0}--\texttt{layer4.0.conv2}):} variance-only correlations $0.867, 0.420, 0.569, 0.564$ jump to FWL correlations $1.000, 0.997, 0.994, 0.865$. Critically, ablating only the signed cross-terms in Stage~2 collapses the correlation from $0.997$ down to $0.330$ (\cref{fig:fig7}B, top), confirming directly that signed readout synthesis---not unsigned energy alone---governs candidate recovery value.
  \item \textbf{ViT-S/16 (early, middle, late post-GELU \texttt{fc2} inputs; $614$ candidates each):} correlations rise from $0.919 \to 0.999$ ($+0.080$), $0.935 \to 0.996$ ($+0.061$), and $0.948 \to 0.981$ ($+0.032$).
  \item \textbf{Swin-T (early, middle, late post-GELU \texttt{fc2} inputs; $307 / 614 / 614$ candidates):} correlations rise from $0.728 \to 0.995$ ($+0.267$), $0.842 \to 0.992$ ($+0.150$), and $0.867 \to 0.975$ ($+0.108$). In the middle cut, FWL reverses the top candidate selection from channel $\#653$ (chosen by variance alone, recovering $85.3\%$ of the optimal gain) to channel $\#826$, which attains $100.0\%$ of the held-out optimum (\cref{fig:fig7}B, bottom).
\end{itemize}
Nested sample-size trajectories from $64$ to $4{,}096$ images confirm that the FWL ranking gains are stable across every depth and model (\cref{fig:fig7}C).

\paragraph{Local MSE recovery versus class-margin error decoupling.}
Finally, we examine how local cut recovery propagates to the final clean top-1 minus top-2 class margin at the middle MLP block under $40\%$ channel removal (\cref{fig:fig7}D). Under direct channel slicing (no OLS folding), moving channel-selection statistics from pre-GELU to post-GELU reduces normalized local \texttt{fc2} output MSE in \emph{both} architectures (ViT-S/16: $0.1724 \to 0.1071$; Swin-T: $0.4376 \to 0.3784$). Yet final class-margin MAE drops sharply in ViT-S/16 (post minus pre: $-0.1756\,[-0.1929, -0.1583]$ logit) while remaining statistically flat in Swin-T ($+0.0009\,[-0.0120, +0.0130]$ logit).

An image-wise symmetric first-order decomposition of $|J_{\mathrm{down}}\delta h|$ explains this decoupling directly within our readout framework (\cref{fig:fig7}D, right). In ViT-S/16, moving selection post-GELU reduces both perturbation magnitude ($-0.1106$ logit) and gradient alignment ($-0.0351$ logit). In Swin-T, smaller local perturbations lower predicted margin error by $-0.0266$ logit, but stronger alignment of the residual perturbation with the downstream margin gradient increases it by $+0.0310$ logit---nearly cancelling the local MSE gain. When both selectors are paired with OLS weight folding, moving from pre-GELU to post-GELU selection significantly lowers real class-margin MAE in both models: by $-0.0306\,[-0.0370, -0.0241]$ logit in Swin-T and by $-0.1094\,[-0.1181, -0.1004]$ logit in ViT-S/16. The core lesson of \cref{sec:exp_pruning} thus mirrors the entire manuscript: \emph{even knowing a channel's conditional residual variance, predicting how much local function restoring it recovers still requires knowing how the fixed downstream readout combines its signed contributions.}

%% file: sections/05_discussion.tex
\section{Discussion, Scope, and Mechanistic Boundaries}
\label{sec:discussion}

A mechanistic explanation is only as useful as the clarity of its boundaries. Our large-sample experiments not only confirm shared encoding and conditional signed readout across vision and language architectures, but also pinpoint four regimes where naive generalizations fail.

\subsection{Grouping Scale and Architectural Non-Universality}
\label{sec:disc_scale}

Two distinct aggregation effects emerged across our experiments. In frozen language models (\cref{sec:exp_paths}), collapsing opposing last-layer MLP units inside coarse structural pathways ($M$ and $J$) masked relation-specific cancellation; separating functional units at identical total derivative deepened $\Delta R$ by $+0.0753$ in Qwen2.5 and $+0.1227$ in Llama-3.2. In original ImageNet ViT-B/16 (\cref{sec:exp_gating}), grouping MLP neurons into contiguous coordinate-index blocks diluted held-out antagonism to $+0.46 \times 10^{-5}$ logit; keeping weights, layer, budget, and images identical while replacing contiguous indices with scattered functional units unlocked $+30.86 \times 10^{-5}$ logit. Together, these findings show that signed antagonism is organized along functional combinations distributed across coordinate indices rather than inside contiguous blocks or coarse layer bins.

Equally important, interference polarity is not an intrinsic architectural slogan. Although ViT-B/16 exhibits strong pathway cancellation ($R = 0.284$), Swin-T exhibits net constructive texture interference ($R = 1.113$, random baseline $0.992$), and in ViT-B/16 both measured texture and random directions cancel strongly while any excess cancellation for texture over random remains unresolved ($+0.006\,[-0.004, +0.017]$ on $1{,}024$ images). Whether a cut reinforces or cancels a factor reflects the joint geometry of trained weights, cut placement, task readout, and grouping scale, rather than a binary distinction between convolutional networks and Transformers.

\subsection{Relative Cancellation Versus Absolute Sensitivity Under Distribution Shift}
\label{sec:disc_corruption}

Under distribution shift, the dimensionless interference ratio $\Rint$ and absolute output sensitivity can separate sharply. We examined Oxford-IIIT Pets ResNet-18 across seven operating points---the clean baseline plus Gaussian noise and Gaussian blur at severities $1, 3,$ and $5$---holding the clean top-1 class, head-weight sign partition, and donor edits fixed across $444$ confirmation images (full dose curves and replication on the $128$-image cohort in \cref{supp:s8}). As corruption severity rises, classification accuracy collapses from $87.6\%$ (clean) to $34.7\%$ (highest noise) and $25.9\%$ (highest blur), while both background and foreground two-group ratios fall sharply below one: the equal-weight highest-severity background shift is $-0.300\,[-0.347, -0.254]$ (noise $-0.313$, blur $-0.287$; replicated at $-0.296$ on the $128$-image cohort), and foreground $R^{(2)}$ drops from $1.089$ to $0.777$ (noise) and $0.897$ (blur).

The clearest illustration of why $\Rint$ must be read alongside absolute response energy occurs under highest-severity blur: background $R^{(2)}$ drops from $0.906$ to $0.619$ (stronger relative between-group cancellation), yet the net coherent background response energy \emph{rises} from $0.00351$ to $0.00559$ (\cref{fig:s8_all}). Applying the exact two-group identity $R^{(2)} = 1 + bc$ (\cref{app:thm3}) to the equal-weight highest-severity background shift decomposes the $-0.300$ drop into approximately $83\%$ from stronger angular opposition between positive- and negative-weight group response vectors ($-0.2495$; mean cosine drops from $-0.140$ to $-0.452$) and $17\%$ from closer norm balance ($-0.0503$; balance rises from $0.676$ to $0.829$), closely reproduced on the $128$-image batch ($-0.2527$ and $-0.0432$; \cref{fig:s8_geom}). Larger relative cancellation therefore does not automatically imply smaller residual sensitivity or higher robustness under corruption.

\subsection{Single-Cut Local Linearization Versus Whole-Network Pruning}
\label{sec:disc_pruning}

Theorem~\ref{thm:fwl} is exact for a single Type~I cut with a fixed linear successor, positive semidefinite output metric, and optimal least-squares weight folding, where it elevates held-out candidate recovery ranking across all $10$ tested cuts ($4{,}096$ ImageNet images; \cref{sec:exp_pruning}). Beyond this single-cut regime, two boundaries remain explicit. First, when testing whether Swin-T's nine shifted spatial windows improve middle-block OLS compensation, real window stratification outperforms equal-capacity random stratification ($0.017068$ versus $0.017425$ MSE; difference $+0.000357\,[0.000347, 0.000367]$), yet global position-shared OLS achieves the lowest held-out error ($0.016910$, difference $-0.000158$ vs.\ real windows; \cref{supp:s7}), reflecting bias--variance tradeoffs under frozen calibration samples and co-variation between windows and spatial coordinates. Second, single-cut direct slicing does not by itself explain historical whole-network multi-layer pruning differences across six architecture families (\cref{tab:s7b_whole}, $3{,}200$ historical ImageNet images, $257$ overlapping with our local cohort), where $40\%$ unlabeled pruning gains over $L_2$ shift from $+24.56$~pp in ViT-S/16 to $-10.69$~pp in Swin-T and task-conditioned rankings depend on the comparator baseline ($+4.66$~pp over Fisher versus $-7.56$~pp below Wanda). Extending fixed-readout FWL theory to coupled multi-layer nonlinear compression remains an open problem.

\subsection{Source-Coordinate Inversion as a Discovery Instrument}
\label{sec:disc_inversion}

Our large-sample boundary tests in \cref{supp:s9} confirm both halves of \cref{ax:inversion_boundary} and \cref{thm:commutation}. On the discovery side, source-grounded inversion with shared-axis residualization (Eq.~15 of \citet{shu2026source}) provides a genuine, reproducible signal for locating candidate region-sensitive channel subsets: across $1{,}020$ eligible CUB confirmation images, de-shared residual groups outperform strength-matched random controls across all four cascade endpoints---$7\times 7$ spatial RMS ($+0.02812$), GAP response ($+0.01096$), unsigned channel weight sum ($+0.00362$), and matched-minus-mismatched absolute margin-restoration magnitude (signed sum within each group followed by the absolute value of the restored margin change, minus strength-matched random controls: $+0.00199\,[0.00026, 0.00373]$; de-shared versus full-map difference $+0.00122\,[-0.00102, +0.00345]$ remains unresolved; \cref{fig:s9_main}). Evaluating signed group predictions calibrated on the $32$-image development split (frozen projection slope $0.03187$) yields a small positive mean per-image group-level signed Pearson correlation of $+0.0485\,[0.0114, 0.0859]$ across the $1{,}020$ held-out images. Likewise, changing LAC weighting from $\nabla_z \odot z$ to $\nabla_z$ reproducibly raises foreground energy share by $+9.43$~pp ($56.62\% \to 66.06\%$) across $444$ Pets images (\cref{fig:s9_main}).

On the boundary side, however, source-coordinate inversion cannot replace native dual sensitivity. Across the $444$ Pets images, covariance-weighted native gradients $\Psi = \Sigma_X \nabla_x m_q$ rank signed candidate pixel edits with a Spearman correlation of $0.9936$, whereas whole-field Wiener inversion achieves only $0.0139\,[-0.0272, +0.0556]$ (mean normalized commutation defect $\|\Dcomm\|_2 / \|\Psi\|_2 = 1.0001\,[0.9992, 1.0011]$; \cref{fig:s9_main}). Image-level rank correlations between LAC and native two-group $R^{(2)}$ are weak ($0.114$ foreground, $0.088$ background crossing zero). Furthermore, cross-operator first-shared-axis projections confirm that different linear inverters exhibit incompatible signed interference structure: $\text{LAC} \to \text{LAC}$ yields $R_b = 0.500\,[0.435, 0.578]$ (passing sign-flip, permutation, and wrong-class nulls), whereas $\text{Wiener} \to \text{Wiener}$ is strongly constructive ($33.93$ in Pets, $25.58$ in CUB), and projecting LAC weights onto the same-image Wiener first axis yields $R_b = 1.282\,[1.207, 1.361]$ despite strong unsigned axis cosine ($0.585$ versus $0.102$ mismatched; \cref{fig:s9_axis}). Source-grounded inversion is thus best used as a hypothesis-generating microscope whose structural leads are subsequently tested natively.

\subsection{Limitations and Open Directions}
\label{sec:disc_limitations}

Four limitations bound our conclusions and point to concrete next steps. First, all experiments study parameter-frozen trained inference; how shared encoding and conditional signed cancellation emerge during optimization dynamics, pretraining data scaling, or fine-tuning remains open. Second, our controlled factors focus on spatial background edits in vision and structured entity-binding relations in language; extending native disjoint-block and functional-gating tests to richer semantic compositions and generative multi-token rollouts is a natural continuation. Third, functional MLP grouping was executed in ViT-B/16, Qwen2.5, and Llama-3.2; applying the same development-selected functional gating to intermediate nonlinear convolutional layers remains to be tested directly. Fourth, as noted in \cref{sec:disc_pruning}, generalizing single-cut FWL recovery to multi-layer joint compression under nonlinear interactions remains an open theoretical and engineering challenge.

%% file: sections/06_conclusion.tex
\section{Conclusion}
\label{sec:conclusion}

We addressed a foundational question about inference in trained deep neural networks: how can internal representations broadly retain input information across disjoint units while producing outputs that are sharply selective for the current task? Across vision classifiers and frozen language models, our primal--dual framework and large-sample native experiments resolve this question through two coupled mechanisms. Deep networks need not erase contextual or relational factors upstream to behave selectively: multiple disjoint representational units can jointly retain recognizable structure of the same factor (\emph{shared encoding}), while downstream channels and nonlinear pathways combine those units' contributions with readout-dependent magnitudes and signs (\emph{conditional signed readout}).

Framing fixed-task decomposition and varying-task switching as two projections of one local synthesis law ($\tau_q = \sum_\alpha t_\alpha$, $|\tau_q| = \sqrt{\Eincoh}\,\sqrt{\Rint}$) unifies static class-head background cancellation in CNNs, multi-path Gram geometry and grouping-scale dilution in Transformers, and prompt-conditioned modulation or sign reversal in frozen language models. Most critically, the framework survives risky out-of-sample tests under real interventions: attenuating development-selected antagonistic units amplifies a factor's decision impact on unseen inputs, attenuating the same frozen units during a single ViT-B/16 forward pass increases the mean magnitude of the background-edit effect on one class margin while decreasing it on another (with the opposite pattern for a second frozen group; four-cell paired interaction $+78.30 \times 10^{-5}$ logit), and projecting the same readout coupling onto fixed linear successors predicts single-channel pruning recovery across $10$ cuts and $4{,}096$ ImageNet images. Together with explicit boundaries on grouping scale, distribution shift, multi-layer pruning, and source-coordinate inversion, these results provide a grounded, predictive account of how deep networks encode and selectively use information.

\paragraph{Code, data, and supplementary materials.}
All primary tables and all $86$ non-duplicate supplementary graphic plates (spanning every confirmation cohort, early/middle/late pruning surgery sheet, and historical exploratory control) are embedded directly in Sections~\ref{app:notation}--\ref{supp:s9} as \cref{fig:s1a}--\cref{fig:s9_extra} (omitting only a standalone full-width duplicate of \cref{fig:fig7}C). Frozen confirmation summaries, analysis code, and plotting scripts are publicly available at \url{https://github.com/k-Jayus/How-Do-Deep-Networks-Encode-and-Selectively-Use-Information-}.

%% file: supplementary/S01_notation.tex
\section{Unified Conflict-Free Notation}
\label{app:notation}

\Cref{tab:notation} consolidates every mathematical symbol used across the main text (\cref{sec:framework,sec:experiments,sec:discussion}) and the supplementary material, explicitly separating primal representation objects from dual sensitivity objects in accordance with \cref{ax:primal_dual}.

\begin{table}[htbp]
\centering
\caption{\textbf{Conflict-free notation table.} Primal objects transform covariantly with coordinate rescalings $H_{T,k} \mapsto d_k H_{T,k}$, whereas dual objects transform contravariantly ($d_k^{-1}$).}
\label{tab:notation}
\scriptsize
\begin{tabularx}{\textwidth}{@{}l l >{\raggedright\arraybackslash}X@{}}
\toprule
\textbf{Symbol} & \textbf{Space / Type} & \textbf{Definition} \\
\midrule
$x \in \Xcal$ & Primal input & Source input ($L_0 \times C_0$: pixels in vision, token embeddings in language); empirical second-moment / covariance matrix $\Sigma_X$. \\
$T,\, H_T(x) \in \Hcal_T$ & Primal cut state & Internal cut $T$ and hidden activation vector/tensor $H_T(x) = f_{\mathrm{up}}(x)$. \\
$\delta H_T^{(f)} \in \Hcal_T$ & Primal perturbation & First-order pushforward $J_{\mathrm{up},T}(x)\delta x$ or direct hidden injection induced by controlled factor $f$. \\
$\{\Pi_k\}_{k=1}^K,\, \Gcal$ & Orthogonal projections & Complete mutually orthogonal coordinate/block partition ($\Pi_j\Pi_k = \delta_{jk}\Pi_k$, $\sum_k \Pi_k = I_{\Hcal_T}$); coarse grouping $\Gcal=\{g_1,\dots,g_M\}$. \\
$J_{\mathrm{up}}(x),\, J_{F_{\mathrm{down}}}(x;q)$ & Linear operators & Upstream and downstream output-map Jacobians evaluated at working point $x$; $^*$ denotes adjoint transpose. \\
$m_q(x) \in \Rbb,\, u_{\mathrm{out}}(q)$ & Scalar margin & Task- or query-conditioned scalar margin $m_q(x) = u_{\mathrm{out}}(q)^\top F_{\mathrm{down}}(H_T(x);x,q)$ selected by contrast vector $u_{\mathrm{out}}(q)$. \\
$\ell_q(x) \in \Hcal_T^*$ & Dual readout covector & Downstream sensitivity $\ell_q(x) := \nabla_{H_T} m_q(x) = J_{F_{\mathrm{down}}}(x;q)^* u_{\mathrm{out}}(q)$, decomposed over paths $\ell_q = \sum_{\pi\in\Pcal} \ell_q^{(\pi)}$. \\
$t_{k,\pi}(x,q;f) \in \Rbb$ & Scalar pairing & Signed contribution $\inner{\ell_q^{(\pi)}(x)}{\Pi_k \delta H_T^{(f)}}$ for component $\alpha=(k,\pi)\in\Omega(T)$. \\
$\tau_q(x,f),\, \Eincoh$ & Net effect / energy & Net first-order sensitivity $\tau_q = \sum_\alpha t_\alpha$ and readout-weighted incoherent energy $\Eincoh = \sum_\alpha t_\alpha^2$. \\
$\Rint^{\Omega(T)}(f;x,q)$ & Dimensionless ratio & Phase-coherence interference ratio $\tau_q^2 / \Eincoh = 1 + \sum_{\alpha<\beta} 2 t_\alpha t_\beta / \Eincoh$ ($\Eincoh > 0$). \\
$G_{ij} = a_i^\top a_j$ & $P\times P$ PSD matrix & Path-gradient Gram matrix built from path readout gradients $a_i = J_i^* u_q$. \\
$H_G(\lambda)$ & Gating endpoint & Held-out paired change $|\Delta m_G(\lambda)| - |\Delta m(0)|$ under edit-increment attenuation $\lambda$; local dev slope $\kappa_G = \partial |\Delta m_G|/\partial\lambda|_{\lambda=0}$. \\
$A,\, W,\, \Hout$ & Pruning cut objects & Centered calibration activations $A\in\Rbb^{N\times C}$, fixed linear successor $Y = A W^\top$, PSD output metric $\Hout \succeq 0$. \\
$r_{j|S},\, v_j,\, d_{j|S}$ & FWL innovation terms & Conditional residual $r_{j|S} = (I-P_S)a_j$, conditional residual variance $v_j = \frac{1}{N}\|r_{j|S}\|_2^2$, conditional innovation vector $d_{j|S} = A^\top r_{j|S}/\|r_{j|S}\|_2^2$ with $(d_{j|S})_j = 1$. \\
$\nabla_x m_q,\, \Psi(x)$ & Dual / Primal input & Native input gradient $\nabla_x m_q \in \Xcal^*$ and covariance-weighted primal counterpart $\Psi(x) := \Sigma_X \nabla_x m_q(x) \in \Xcal$. \\
$\Aop_x,\, \Wop_{v\leftarrow u},\, \Dcomm$ & Source-inversion objects & Closed-form source-grounded estimator $\Aop_x$, exact linear MMSE estimator $\Wop_{v\leftarrow u}$, commutation defect vector $\Dcomm(x;q)$, and scalar normalized defect $\|\Dcomm\|_2/\|\Psi\|_2$. \\
$u_q = a_q b + r_q$ & Unit source maps & Unit-normalized single-channel inversion map decomposed along first right singular axis $b$ and orthogonal residual $r_q \perp b$. \\
\bottomrule
\end{tabularx}
\end{table}

%% file: supplementary/S02_thm1_proofs.tex
\section{Proof of Theorem~\ref{thm:synthesis}(i), Pathway Inclusion--Exclusion, and Invariance}
\label{app:thm1}

\subsection{Complete Tangent-Space Pathway Decompositions}
\label{app:thm1_paths}

At a cut $T$, let downstream modules be linearized around the clean working point $x$. Three canonical decompositions are used in \cref{sec:experiments}:
\begin{enumerate}
  \item \textbf{Multi-block Transformer four-path expansion ($D, A, M, J$):} For a single linearized residual block $(I + M)(I + A)v = v + Av + Mv + MAv$, the four terms separate bypass, attention-only, MLP-only, and mixed paths. Across multiple downstream blocks, let $\ell_q^{(0)}$ denote the full downstream sensitivity, $\ell_q^{\mathcal{A}}$ the sensitivity with all designated attention writes zero-clamped in tangent space, $\ell_q^{\mathcal{M}}$ with all designated MLP writes clamped, and $\ell_q^{\mathcal{A}\mathcal{M}}$ with both clamped. Inclusion--exclusion yields an exact four-way partition:
  \begin{align}
    \ell_q^{(D)} &= \ell_q^{\mathcal{A}\mathcal{M}}, &
    \ell_q^{(A)} &= \ell_q^{\mathcal{M}} - \ell_q^{\mathcal{A}\mathcal{M}}, \nonumber\\
    \ell_q^{(M)} &= \ell_q^{\mathcal{A}} - \ell_q^{\mathcal{A}\mathcal{M}}, &
    \ell_q^{(J)} &= \ell_q^{(0)} - \ell_q^{\mathcal{A}} - \ell_q^{\mathcal{M}} + \ell_q^{\mathcal{A}\mathcal{M}}.
    \label{eq:ie_transformer}
  \end{align}
  Summing the four expressions identically recovers $\ell_q^{(D)} + \ell_q^{(A)} + \ell_q^{(M)} + \ell_q^{(J)} \equiv \ell_q^{(0)}$. Note that across multiple blocks, $\ell_q^{(J)}$ collects \emph{all} paths traversing at least one designated attention write and at least one designated MLP write in either temporal order (attention-then-MLP and earlier-MLP-then-later-attention). When splitting a single layer's MLP (as in \cref{sec:exp_paths}) into selected functional units $S$, remaining same-layer units $R$, and bypass, $\ell_q^{(M)}$ and $\ell_q^{(J)}$ partition further while leaving $\ell_q^{(0)}$ and $\tau_q$ identically unchanged.
  \item \textbf{ResNet two-block four-path expansion ($SS, SF, FS, FF$):} Across two consecutive residual blocks $(S_1 + F_1)(S_0 + F_0)$ (where $S_0$ includes the $1\times 1$ downsampling projection in \texttt{layer4.0}), expanding the tangent composition yields $\ell_q = \ell_q^{(SS)} + \ell_q^{(SF)} + \ell_q^{(FS)} + \ell_q^{(FF)}$; for a single residual block (\texttt{layer4.0} $\to$ \texttt{layer4.1}), $\ell_q = \ell_q^{(\mathrm{skip})} + \ell_q^{(F_1)}$.
  \item \textbf{Intra-attention routing versus value split:} Linearizing $P(x)V(x)$ splits $\delta(PV) = (\delta P)V + P(\delta V)$, yielding $\ell_q = \ell_q^{\mathrm{route}} + \ell_q^{\mathrm{val}}$.
\end{enumerate}

\subsection{Complete Vertex Cuts, Input-Edit Specialization, and Eight-Condition Language Protocol}

\paragraph{Complete vertex cut versus partial branch.}
When $\delta H_T^{(f)} = J_{\mathrm{up},T}(x)\delta x$, each component contribution rewrites as $t_{k,\pi} = \inner{\gamma_{k,\pi}(x;q)}{\delta x}$ with input-space covector $\gamma_{k,\pi}(x;q) := J_{\mathrm{up},T}(x)^*\Pi_k^*\ell_q^{(\pi)}(x) \in \Xcal^*$. If $T$ forms a complete vertex cut intercepting every computational path from $x$ to $m_q$, the chain rule guarantees $\sum_{k,\pi}\gamma_{k,\pi}(x;q) \equiv \nabla_x m_q(x)$. If $T$ intercepts only a subset of paths (such as a single residual sub-branch), then $\nabla_x m_q = \nabla_x^{(T)} m_q + \nabla_x^{(\mathrm{bypass}\,T)} m_q$, while the direct cut-injection identity $\sum_{k,\pi} t_{k,\pi} = \inner{\ell_q}{\delta H_T}$ holds unconditionally.

\paragraph{Eight-condition language protocol.}
In the eight-condition entity-binding protocol of \citet{shu2026shared}, conditions are indexed by $q = (a,b,r) \in \{-1,+1\}^3$ with relational injection $\delta H_T(q) = ab\,v$ and selection summary weight $(ar - br)$. Since $a^2 = b^2 = 1$, the combined weight obeys $\chi_G(q) = (ar - br)ab \equiv br - ar$, so the mean signed selection sensitivity decomposes identically across components as $\beta_G = \sum_{k,\pi} \Ebb_q[(br - ar)\,t_{k,\pi}(x_q, q;\, v)]$.

\subsection{Proof of Exact Invariance Properties}

\begin{proof}[Proof of Theorem~\ref{thm:synthesis}(i)]
Let $M = \bigoplus_{k=1}^K M_k$ be any invertible linear transformation commuting with the orthogonal partition $\{\Pi_k\}$ ($M\Pi_k = \Pi_k M$ for all $k$). Under the coordinate change $\tilde{H}_T = M H_T$, primal perturbations transform covariantly as $\delta\tilde{H}_T = M\delta H_T$, whereas every path covector transforms contravariantly as $\tilde{\ell}_q^{(\pi)} = (M^{-1})^* \ell_q^{(\pi)}$ (equivalently $M^{-\top}\ell_q^{(\pi)}$ in Euclidean coordinates). Therefore, for every $\alpha = (k,\pi) \in \Omega(T)$,
\begin{equation}
  \tilde{t}_{k,\pi}
  = \inner{(M^{-1})^* \ell_q^{(\pi)}}{\Pi_k M \delta H_T}
  = \inner{\ell_q^{(\pi)}}{M^{-1} M \Pi_k \delta H_T}
  = \inner{\ell_q^{(\pi)}}{\Pi_k \delta H_T}
  = t_{k,\pi}.
\end{equation}
Because every individual term $t_\alpha$ is invariant, $\tau_q = \sum_\alpha t_\alpha$, $\Eincoh = \sum_\alpha t_\alpha^2$, and $\Rint^{\Omega(T)} = \tau_q^2 / \Eincoh$ are strictly invariant under arbitrary block-invertible rescalings and sign flips. Moreover, because the definition of $t_\alpha$ pairs the tangent perturbation directly with the dual covector without dividing by baseline activation norms $\|H_{T,k}\|_2$, it avoids zero-norm division singularities while faithfully reflecting any real downstream sensitivity (including Jacobian attenuation through LayerNorm or RMSNorm).
\end{proof}

\begin{example}[Unequal-scale $2\times 2$ exact cancellation]
Let $x = (b, f)^\top \in \Rbb^2$ represent background $b$ and foreground $f$, encoded into two unequally scaled hidden units $h_1 = 10b + f$ and $h_2 = -b + f$. Under the linear readout $m_q(x) = f = \frac{1}{11} h_1 + \frac{10}{11} h_2$, the component input covectors are $\gamma_1 = \frac{1}{11}(10, 1)^\top$ and $\gamma_2 = \frac{1}{11}(-10, 10)^\top$. Their sum $\gamma_1 + \gamma_2 = (0, 1)^\top$ cancels background exactly ($\tau_b = 0$, $\Eincoh(b) = \frac{200}{121} > 0$, $\Rint(b) = 0$) while preserving foreground constructively ($\tau_f = 1$, $\Eincoh(f) = \frac{101}{121}$, $\Rint(f) = \frac{121}{101} > 1$).
\end{example}

%% file: supplementary/S03_thm3_commutation.tex
\section{Proof of Theorem~\ref{thm:commutation} and Ideal Linear Commutative Diagram}
\label{app:thm2}

\begin{proof}[Proof of Theorem~\ref{thm:commutation}]
Let $H_u = J_v H_v$ be a sample-independent linear map between two representations $H_v$ and $H_u$ with positive semidefinite second-moment matrices $\Sigma_v$ and $\Sigma_u = J_v \Sigma_v J_v^*$. Define $B := \Sigma_v^{1/2} J_v^*$. Then $\Sigma_v J_v^* = \Sigma_v^{1/2} B$ and $\Sigma_u = B^* B$. Because $\ker(B^* B) = \ker(B)$, the Moore--Penrose pseudoinverse obeys the universal identity $B (B^* B)^+ (B^* B) \equiv B$ without any full-rank assumption on $J_v$ or $\Sigma_v$. Left-multiplying by $\Sigma_v^{1/2}$ and applying both sides to any dual covector $\ell_u$ gives
\begin{equation}
  \Wop_{v\leftarrow u}(\Sigma_u \ell_u)
  = \Sigma_v J_v^* (J_v \Sigma_v J_v^*)^+ \Sigma_u \ell_u
  = \Sigma_v^{1/2} B (B^* B)^+ (B^* B) \ell_u
  \equiv \Sigma_v^{1/2} B \ell_u
  = \Sigma_v (J_v^* \ell_u)
  = \Sigma_v \ell_v.
\end{equation}
\end{proof}

\paragraph{Ideal linear commutative diagram.}
Iterating \cref{eq:wiener_commute} along a sample-independent linear chain $x = H_0 \to H_1 \to \dots \to H_T$ yields a two-row commutative diagram linking dual backpropagation (top row) and primal MMSE Wiener pullback (bottom row) via the layer covariance metrics $\Sigma_k$:
\begin{equation}
\begin{array}{ccccccc}
  \text{Dual space:} & \ell_T & \xrightarrow{\;J_T^*\;} & \ell_{T-1} & \xrightarrow{\;J_{T-1}^*\;} \cdots \xrightarrow{\;J_1^*\;} & \ell_0 = \nabla_x m_q \\[4pt]
  & \Big\downarrow\scriptstyle{\Sigma_T} & & \Big\downarrow\scriptstyle{\Sigma_{T-1}} & & \Big\downarrow\scriptstyle{\Sigma_0 = \Sigma_X} \\[4pt]
  \text{Primal space:} & U_T & \xrightarrow{\;\Wop\;} & U_{T-1} & \xrightarrow{\;\Wop\;} \cdots \xrightarrow{\;\Wop\;} & U_0 \equiv \Sigma_X \nabla_x m_q = \Psi(x)
\end{array}
\label{eq:comm_diagram}
\end{equation}

\paragraph{Three corollaries.}
\begin{enumerate}
  \item \emph{Basis invariance of the transported composite pullback:} Under any invertible coordinate transform $H_T \mapsto M H_T$, $\Sigma_T \mapsto M \Sigma_T M^*$ and $\ell_q \mapsto (M^{-1})^* \ell_q$, so $\Sigma_T \ell_q \mapsto M \Sigma_T \ell_q$ transforms covariantly as a primal vector; when the same linear estimator $\Aop_x$ is transported covariantly into the new coordinates ($\tilde{\Aop}_x = \Aop_x M^{-1}$), the composite pullback $\tilde{\Aop}_x(\tilde{\Sigma}_T \tilde{\ell}_q) = \Aop_x(\Sigma_T \ell_q)$ is strictly invariant. (By contrast, refitting a ridge estimator with an unchanged isotropic scalar penalty in rescaled coordinates defines a different operator.)
  \item \emph{First-order global covariance weighting:} The primal vector $\Psi(x) = \Sigma_X \nabla_x m_q(x)$ weights the native input gradient by the global empirical input covariance $\Sigma_X$ (rather than performing an exact nonlinear manifold projection at $x$).
  \item \emph{Non-equivalence of upstream commutation defect and downstream nonlinearity:} The empirical defect $\Dcomm(x;q) = \Aop_x(\Sigma_T^{\mathrm{global}}\ell_q(x)) - \Sigma_X J_{\mathrm{up}}(x)^*\ell_q(x)$ measures upstream deviation from sample-independent linearity and estimation exactness between $0$ and $T$; it is conceptually distinct from downstream Type~II/III nonlinearity beyond $T$, and cannot be attributed to a single error source without dedicated ablations (\cref{supp:s9}).
\end{enumerate}

%% file: supplementary/S04_gram_decomps.tex
\section{Proof of Theorem~\ref{thm:synthesis}(ii), Two-Scale Bridge, Gram Integral, and Exact Decompositions}
\label{app:thm3}

\subsection{Rademacher Expectation Baseline and Sign-Sampling Rules}

\begin{proof}[Proof of Theorem~\ref{thm:synthesis}(ii)]
Let $s_\alpha \in \{-1,+1\}$ be independent Rademacher random variables ($\Ebb[s_\alpha] = 0$, $\Ebb[s_\alpha s_\beta] = \delta_{\alpha\beta}$). Whenever $\Eincoh = \sum_\alpha t_\alpha^2 > 0$, note that randomizing signs leaves the denominator $\sum_\alpha (s_\alpha t_\alpha)^2 = \sum_\alpha t_\alpha^2 = \Eincoh$ deterministic. Taking expectation of the numerator yields Hutchinson's identity \citep{hutchinson1989stochastic}:
\begin{equation}
  \Ebb_s\!\left[\Rint^{\Omega(T)}(s \odot t)\right]
  = \frac{\Ebb_s\!\left[\big(\sum_\alpha s_\alpha t_\alpha\big)^2\right]}{\Eincoh}
  = \frac{\sum_\alpha t_\alpha^2 + \sum_{\alpha\neq\beta} \Ebb[s_\alpha s_\beta]\, t_\alpha t_\beta}{\Eincoh}
  = 1.
\end{equation}
For any single draw of signs $s$, the realized ratio may lie above or below $1$; the exact baseline $1$ is an expectation over independent sign flips.
\end{proof}
To test pure unit interference, only $s_k$ is randomized; to test pure path interference, only $s_\pi$ is randomized; to test joint unit--path interference, independent $s_{k,\pi}$ are drawn. Output contrast sign inversion ($A - B \leftrightarrow B - A$) negates every $t_\alpha$ identically and therefore leaves $\Rint$ strictly invariant (serving as a symmetry check distinct from genuine task switching $q \to q'$).

\subsection{Target Energy Ratio (\texorpdfstring{$\TER$}{TER}) Decomposition and Five-Mode Table}

For disjoint foreground ($M_{\mathrm{fg}}$) and background ($M_{\mathrm{bg}}$) supports under a metric $\Sigma$ with coherent and incoherent regional energies, the coherent foreground energy share $\TER_{\mathrm{coh}} := E_{\mathrm{coh}}^\Sigma(M_{\mathrm{fg}}) / (E_{\mathrm{coh}}^\Sigma(M_{\mathrm{fg}}) + E_{\mathrm{coh}}^\Sigma(M_{\mathrm{bg}}))$ satisfies
\begin{equation}
  \TER_{\mathrm{coh}} \;=\; \left[ 1 + \frac{\Rint^\Sigma(M_{\mathrm{bg}})}{\Rint^\Sigma(M_{\mathrm{fg}})} \cdot \frac{\Eincoh^\Sigma(M_{\mathrm{bg}})}{\Eincoh^\Sigma(M_{\mathrm{fg}})} \right]^{-1}.
  \label{eq:ter_decomp}
\end{equation}
Consequently, an increase in $\TER_{\mathrm{coh}}$ alone cannot prove background cancellation: for two vectors $v_1 = (1, 1, 0)$ and $v_2 = (1, 0, 1)$ (first coordinate foreground, last two background), $\Rint(M_{\mathrm{fg}}) = 2.0 > 1$ while $\Rint(M_{\mathrm{bg}}) = 1.0$, raising $\TER_{\mathrm{coh}}$ purely through foreground reinforcement. \Cref{tab:ter_modes} enumerates all five diagnostic regimes.

\begin{table}[htbp]
\centering
\caption{\textbf{Five interference regimes for regional foreground energy concentration ($\TER_{\mathrm{coh}}$).} Measuring $\Rint$ separately on foreground and background disambiguates cancellation from reinforcement.}
\label{tab:ter_modes}
\scriptsize
\begin{tabular}{@{}l c c c l@{}}
\toprule
\textbf{Mode} & $\Rint(M_{\mathrm{bg}})$ & $\Rint(M_{\mathrm{fg}})$ & $\TER_{\mathrm{coh}}$ & \textbf{Physical mechanism} \\
\midrule
Mode~1 & $< 1$ & $\approx 1$ & Increases & Pure background cancellation \\
Mode~2 & $\approx 1$ & $> 1$ & Increases & Pure foreground reinforcement (e.g.\ $v_1=(1,1,0)$, $v_2=(1,0,1)$) \\
Mode~3a & $< 1$ & $> 1$ & Strongly increases & Joint background cancellation + foreground reinforcement (Pets native \cref{sec:exp_channels}) \\
Mode~3b & $< \Rint(M_{\mathrm{fg}}) < 1$ & $< 1$ & Increases & Both regions attenuated, background more strongly (LAC maps, \cref{supp:s9}) \\
Mode~4 & $\approx 1$ & $\approx 1$ & Unchanged & Incoherent / random-phase superposition \\
\bottomrule
\end{tabular}
\end{table}

\subsection{Proof of the Two-Scale Bridge Identity}

\begin{proof}[Proof of \cref{eq:two_scale}]
Let $\Gcal = \{g_1,\dots,g_M\}$ partition $\Omega(T)$ into disjoint groups with sums $U_m = \sum_{\alpha\in g_m} t_\alpha$. Expanding the fine-scale numerator and each group square gives
\begin{align}
  \sum_{m=1}^M U_m^2 &= \sum_{m=1}^M \left(\sum_{\alpha \in g_m} t_\alpha^2 + \sum_{\alpha \neq \beta \in g_m} t_\alpha t_\beta\right) = \Eincoh\,(1 + R_{\mathrm{within}}), \\
  \left(\sum_{m=1}^M U_m\right)^2 &= \sum_{m=1}^M U_m^2 + \sum_{m \neq m'} U_m U_{m'} = \Eincoh\,(1 + R_{\mathrm{within}} + R_{\mathrm{between}}).
\end{align}
Dividing the second identity by $\Eincoh$ yields $\Rint^{\mathrm{fine}} = 1 + R_{\mathrm{within}} + R_{\mathrm{between}}$. Dividing the second identity by the first (whenever $1 + R_{\mathrm{within}} > 0$) yields
\begin{equation}
  \Rint^{\mathrm{group}}
  = \frac{\big(\sum_m U_m\big)^2}{\sum_m U_m^2}
  = \frac{1 + R_{\mathrm{within}} + R_{\mathrm{between}}}{1 + R_{\mathrm{within}}}
  \equiv \frac{\Rint^{\mathrm{fine}}}{1 + R_{\mathrm{within}}}
  = 1 + \frac{R_{\mathrm{between}}}{1 + R_{\mathrm{within}}}.
\end{equation}
When partitioning weights by sign, exact zero-weight channels are retained in a separate zero group $\hat{\Gamma}_0$ so $\hat{\Gamma} = \hat{\Gamma}_+ + \hat{\Gamma}_- + \hat{\Gamma}_0$ remains exact.
\end{proof}

\subsection{Exact Gaussian Quadrature Formula for Path-Gradient Gram Matrices}

Let $a_1,\dots,a_P \in \Rbb^D$ denote the $P$ path readout gradients at cut $T$, with Gram matrix $G_{ij} = a_i^\top a_j$ and eigendecomposition $G = U \diag(\lambda_1,\dots,\lambda_P) U^\top$. For an isotropic Gaussian (equivalently, uniformly spherical) random direction $v_{\mathrm{rand}}$ (note that merely specifying a covariance proportional to $I_D$ does not uniquely determine the expectation of a ratio), $\Rint(v_{\mathrm{rand}})$ is zero-homogeneous in $v_{\mathrm{rand}}$, so we may evaluate the expectation under a standard Gaussian vector $z \sim \mathcal{N}(0, I_D)$. Then the path contribution vector $g := (a_1^\top z,\dots,a_P^\top z)^\top$ follows $\mathcal{N}(0, G)$, and $y := U^\top g \sim \mathcal{N}(0, \diag(\lambda))$ has independent coordinates with $\sum_i g_i^2 = \sum_j y_j^2$ and $\sum_i g_i = \sum_j c_j y_j$, where $c_j := \mathbf{1}^\top U_{:j}$. Using the Laplace identity $1/w = \int_0^\infty e^{-tw}\,dt$ ($w > 0$) and Fubini's theorem:
\begin{align}
  \Ebb\!\left[\Rint(v_{\mathrm{rand}})\right]
  &= \Ebb\!\left[\frac{\big(\sum_j c_j y_j\big)^2}{\sum_k y_k^2}\right]
   = \int_0^\infty \Ebb\!\left[\Big(\sum_j c_j y_j\Big)^2 \exp\!\Big(-t \sum_k y_k^2\Big)\right] dt \nonumber\\
  &= \sum_{j=1}^P c_j^2 \int_0^\infty \Ebb\!\left[y_j^2 e^{-t y_j^2}\right] \prod_{k \neq j} \Ebb\!\left[e^{-t y_k^2}\right] dt.
\end{align}
Evaluating the elementary Gaussian integrals $\Ebb[e^{-t y_k^2}] = (1 + 2t\lambda_k)^{-1/2}$ and $\Ebb[y_j^2 e^{-t y_j^2}] = \lambda_j (1 + 2t\lambda_j)^{-3/2}$ yields \cref{eq:gram_integral}:
\begin{equation}
  \Ebb\!\left[\Rint(v_{\mathrm{rand}})\right]
  = \sum_{j=1}^P (\mathbf{1}^\top U_{:j})^2 \int_0^\infty \frac{\lambda_j}{1 + 2t\lambda_j}\prod_{k=1}^P (1 + 2t\lambda_k)^{-1/2}\,dt.
\end{equation}
Evaluated via $256$- or $384$-point Gauss--Legendre quadrature after substituting $t = u/(1-u)$, this population integral closely tracks the finite Monte Carlo random-direction banks across all four visual backbones (on ResNet-18, both predict constructive interference: analytic integral $1.1488$ versus frozen random-direction bank $1.1332$, paired image-wise difference $+0.0156\,[0.0119, 0.0193]$; \cref{sec:exp_paths}, \cref{supp:s4}).

\subsection{Exact Symmetric Product Decompositions}

For any product $Z = X Y$ changing from state $0$ to state $1$, the exact symmetric Shapley/midpoint decomposition splits $\Delta Z := Z_1 - Z_0$ into two additive terms with zero remainder:
\begin{equation}
  \Delta Z \;\equiv\; \underbrace{\frac{Y_0 + Y_1}{2}\,(X_1 - X_0)}_{\text{change due to } X} \;+\; \underbrace{\frac{X_0 + X_1}{2}\,(Y_1 - Y_0)}_{\text{change due to } Y}.
  \label{eq:sym_decomp}
\end{equation}
We apply \cref{eq:sym_decomp} in three places across the study:
\begin{enumerate}
  \item \textbf{Amplitude versus composition (\cref{sec:exp_conditions}):} With $|\tau| = \sqrt{\Eincoh}\,\sqrt{\Rint}$, setting $X = \sqrt{\Eincoh}$ and $Y = \sqrt{\Rint}$ splits the first-order CUB readout-switching DiD ($+0.05091$) exactly into amplitude ($+0.03608$) and composition ($+0.01484$).
  \item \textbf{Two-group corruption geometry (\cref{sec:disc_corruption}, \cref{supp:s8}):} Let $P, N \in \Rbb^4$ be the positive- and negative-weight group response vectors across the four donor edits on a source image. The two-group interference ratio obeys the exact identity
  \begin{equation}
    R^{(2)} = \frac{\|P + N\|_2^2}{\|P\|_2^2 + \|N\|_2^2} \equiv 1 + b\,c, \qquad
    b := \frac{2\|P\|_2\|N\|_2}{\|P\|_2^2 + \|N\|_2^2} \in [0,1], \qquad
    c := \frac{P^\top N}{\|P\|_2 \|N\|_2} \in [-1,1],
    \label{eq:r_eq_1_plus_bc}
  \end{equation}
  where $b$ measures norm balance and $c$ measures group cosine. Applying \cref{eq:sym_decomp} to $b c$ image by image splits the corruption shift $\Delta R^{(2)} = -0.2998$ ($444$ images) exactly into angle change ($-0.2495$) and norm-balance change ($-0.0503$).
  \item \textbf{First-order pruning margin decomposition (\cref{sec:exp_pruning}, \cref{supp:s7}):} Writing $|J_{\mathrm{down}}\delta h| = \|\delta h\|_2 \cdot \|J_{\mathrm{down}}\|_2 \cdot |\cos(\delta h, J_{\mathrm{down}}^*)|$ and applying \cref{eq:sym_decomp} to perturbation norm versus gradient alignment isolates why pre-to-post-GELU slicing cancels in Swin-T ($-0.0266$ size vs.\ $+0.0310$ alignment) while reinforcing in ViT-S/16 ($-0.1106$ and $-0.0351$).
\end{enumerate}

%% file: supplementary/S05_fwl_toy_audit.tex
\section{Proof of Theorem~\ref{thm:fwl}, Unified 2D Model, Gating Overshoot, and Pruning Audit}
\label{app:thm4}

\subsection{Proof of the FWL Three-Factor Channel-Recovery Identity}

\begin{proof}[Proof of Theorem~\ref{thm:fwl}]
Let $A \in \Rbb^{N \times C}$ be centered calibration activations at a Type~I cut with fixed linear successor $Y = A W^\top \in \Rbb^{N \times d_{\mathrm{out}}}$ and output metric $\Hout \succeq 0$. For any retained subset $S \subset \{1,\dots,C\}$ with complement $R = S^c$, optimal least-squares folding uses $B_S = A_S^+ A_R$ and $W'_{\mathrm{keep}} = W_S + W_R B_S^\top$. Because $A_S B_S = P_S A_R$ where $P_S := A_S A_S^+$ is the orthogonal projector onto $\operatorname{col}(A_S)$, the folded prediction equals
\begin{equation}
  \widehat{Y}_S = A_S (W_S + W_R B_S^\top)^\top = A_S W_S^\top + P_S A_R W_R^\top = P_S (A_S W_S^\top + A_R W_R^\top) \equiv P_S Y,
\end{equation}
yielding reconstruction loss $L(S) = \frac{1}{N}\|(I - P_S)Y\|_{F,\Hout}^2$. Now restore a single omitted channel $j \notin S$ with conditional residual $r_{j|S} = (I - P_S) a_j$. If $v_j := \frac{1}{N}\|r_{j|S}\|_2^2 = 0$, then $a_j \in \operatorname{col}(A_S)$, $P_{S\cup\{j\}} = P_S$, and $L(S) - L(S\cup\{j\}) = 0$. When $v_j > 0$, Gram--Schmidt orthogonalization gives the rank-one projector increment
\begin{equation}
  P_{S\cup\{j\}} - P_S \;=\; \frac{r_{j|S} r_{j|S}^\top}{\|r_{j|S}\|_2^2}.
\end{equation}
By orthogonality of $P_{S\cup\{j\}} - P_S$ and $I - P_{S\cup\{j\}}$, the Frisch--Waugh--Lovell (FWL) theorem \citep{frisch1933partial,lovell1963seasonal} yields
\begin{equation}
  L(S) - L(S\cup\{j\})
  = \frac{1}{N}\norm{(P_{S\cup\{j\}} - P_S) Y}_{F,\Hout}^2
  = \frac{1}{N}\norm{r_{j|S} b_{j|S}^\top}_{F,\Hout}^2
  = v_j \norm{b_{j|S}}_{\Hout}^2,
\end{equation}
where $b_{j|S} := Y^\top r_{j|S} / \|r_{j|S}\|_2^2 = W A^\top r_{j|S} / \|r_{j|S}\|_2^2 = W d_{j|S} = \sum_k z_k$ with $z_k := W \Pi_k d_{j|S}$. Whenever $\sum_k \|z_k\|_{\Hout}^2 > 0$, factoring out the incoherent output energy $\sum_k \|z_k\|_{\Hout}^2$ gives the exact three-factor identity \cref{eq:fwl_identity}:
\begin{equation}
  L(S) - L(S\cup\{j\}) \;\equiv\; v_j \cdot \left(\sum_k \norm{z_k}_{\Hout}^2\right) \cdot \Rint^{\Hout}(d_{j|S}),
\end{equation}
where $\Rint^{\Hout}(d_{j|S}) := \|\sum_k z_k\|_{\Hout}^2 / \sum_k \|z_k\|_{\Hout}^2$. Centering absorbs constant intercepts into biases or batch-normalization shifts; strict zero-intercept cuts replace centered covariances by uncentered second moments. Numerically, fast FWL scores match explicit least-squares refitting to $\le 9.43\times 10^{-18}$ on early/middle Vision Transformer cuts and $\le 1.005\times 10^{-14}$ across all six cuts on $4{,}096$ ImageNet images (\cref{sec:exp_protocol}, \cref{tab:s1_audit}).
\end{proof}

\subsection{Unified 2D Analytical Toy Model}

Let $s, t$ be independent zero-mean scalar factors with $\sigma_s^2 \gg \sigma_t^2$, encoded into two shared units $h_+ = s + t$ and $h_- = s - t$. Consider a task-conditioned linear readout $Y_c = \eta_c s + t = \frac{\eta_c + 1}{2} h_+ + \frac{\eta_c - 1}{2} h_-$ under Euclidean metric $\Hout = 1$:
\begin{enumerate}
  \item \textbf{Shared encoding dominance:} The first principal axis $s$ accounts for $\EVR_1 = \sigma_s^2 / (\sigma_s^2 + \sigma_t^2) \to 1$, and both $h_+$ and $h_-$ individually carry $s$.
  \item \textbf{Positive factor utility via partial cancellation:} At $\eta_c = 0$, $s$ cancels completely ($\Rint(s) = 0$) while $t$ reinforces ($\Rint(t) = 2$). At $\eta_c = 0.5$ ($Y = 0.75 h_+ - 0.25 h_-$), $s$ exerts a clearly positive net effect ($\tau_s = +0.5$) even though its interference ratio is strictly below one ($\Rint(s) = 0.5^2 / (0.75^2 + 0.25^2) = 0.4 < 1$), demonstrating that models can utilize a positive factor through partial cancellation.
  \item \textbf{Readout-conditioned sign reversal and pruning-order flip:} Reversing $\sign(\eta_c)$ flips both the decision sensitivity $\partial Y_c / \partial s = \eta_c$ and the first-step FWL pruning recovery order:
  \begin{equation}
    L(\emptyset) - L(\{h_\pm\}) \;=\; \frac{(\eta_c \sigma_s^2 \pm \sigma_t^2)^2}{\sigma_s^2 + \sigma_t^2}.
  \end{equation}
  Thus $h_+$ is the superior unit to retain when $\eta_c > 0$, whereas $h_-$ becomes superior when $\eta_c < 0$, despite identical upstream activation covariance.
\end{enumerate}

\subsection{Linearized Promoting-Group Sign-Preservation and Magnitude-Reduction Bounds}

In a linearized model where the total edit effect is $\tau(0) = \sum_\alpha t_\alpha \neq 0$ and attenuating a unit group $G$ with total contribution $t_G = \sum_{\alpha\in G} t_\alpha$ shifts the net effect to $\tau_G(\lambda) = \tau(0) - \lambda t_G$:
\begin{itemize}
  \item If $G$ is a \emph{promoting} group ($\sign(t_G) = \sign(\tau(0))$), then $\tau_G(\lambda)$ retains its original sign without crossing zero if and only if
  \begin{equation}
    0 \;<\; \lambda \;<\; \frac{\abs{\tau(0)}}{\abs{t_G}},
    \label{eq:no_sign_flip}
  \end{equation}
  whereas the weaker magnitude-reduction condition $|\tau_G(\lambda)| < |\tau(0)|$ (which permits sign inversion across zero without overshooting past $-\tau(0)$, e.g.\ $\tau(0)=1, t_G=3, \lambda=0.5 \Rightarrow \tau_G=-0.5$) holds if and only if
  \begin{equation}
    0 \;<\; \lambda \;<\; \frac{2\abs{\tau(0)}}{\abs{t_G}}.
    \label{eq:overshoot_bound}
  \end{equation}
  For doses exceeding $2|\tau(0)|/|t_G|$ (possible when $|t_G| > 2|\tau(0)|$), attenuating a promoting group can flip the margin sign and increase $|\tau_G(\lambda)|$, motivating small primary doses ($\lambda_0 = 0.1$) in our gating protocol (\cref{sec:exp_gating}).
  \item If $G$ is an \emph{antagonistic} group ($\sign(t_G) = -\sign(\tau(0))$), linearized attenuation strictly increases $|\tau_G(\lambda)| = |\tau(0)| + \lambda |t_G|$ for all $\lambda > 0$; in real nonlinear networks, however, finite-dose interactions and sample-to-sample heterogeneity can break linearity, making whether development-selected groups predict the sign and magnitude of finite gating at $\lambda_0 = 0.1$ on held-out inputs a genuinely falsifiable empirical test.
\end{itemize}

\subsection{Pruning Interface Audit for ViT-S/16 and Swin-T}

Auditing the released structured-pruning codebase of \citet{shu2026pruning} confirmed that both ViT-S/16 and Swin-T prune output rows of \texttt{mlp.fc1} jointly with input columns of \texttt{mlp.fc2} (so both cuts belong to Type~I with linear successor $W = W_{\mathrm{fc2}}$). However, the released unlabeled selector collected activation statistics \emph{before} GELU (\texttt{fc1} output) and did not apply ridge/OLS weight folding, whereas \cref{thm:fwl} requires observing activations \emph{after} GELU (the direct input to \texttt{fc2}) paired with OLS folding. Our controlled six-arm protocol in \cref{sec:exp_pruning} and \cref{supp:s7} isolates pre-GELU versus post-GELU selection and direct slicing versus OLS folding cleanly across early, middle, and late blocks.

%% file: supplementary/S06_to_S10_experiments.tex
\section{Cohorts, Provenance, Reuse Matrix, and Precision Audits}
\label{supp:s1}

This section documents checkpoint provenance, exact cohort separation and reuse disclosures, statistical resampling conventions, and numerical closure checks backing \cref{sec:exp_protocol}.

\begin{figure}[p]
  \centering
  \includegraphics[width=0.95\textwidth]{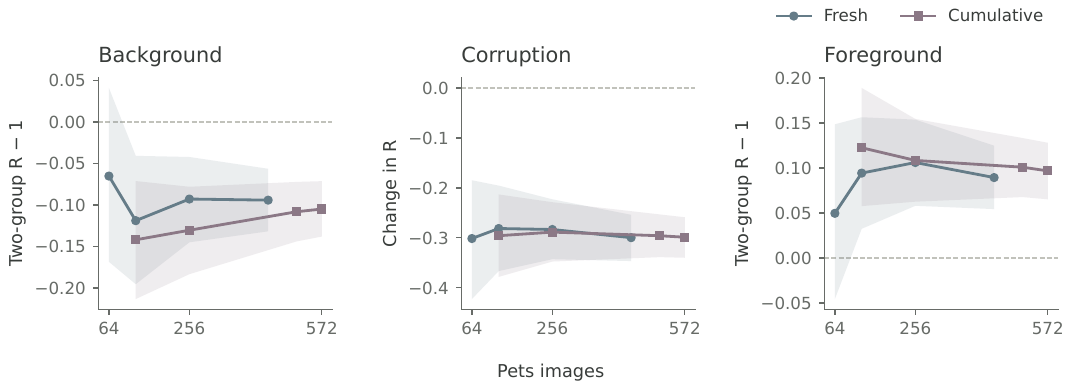}\\[0.35em]
  \includegraphics[width=0.95\textwidth]{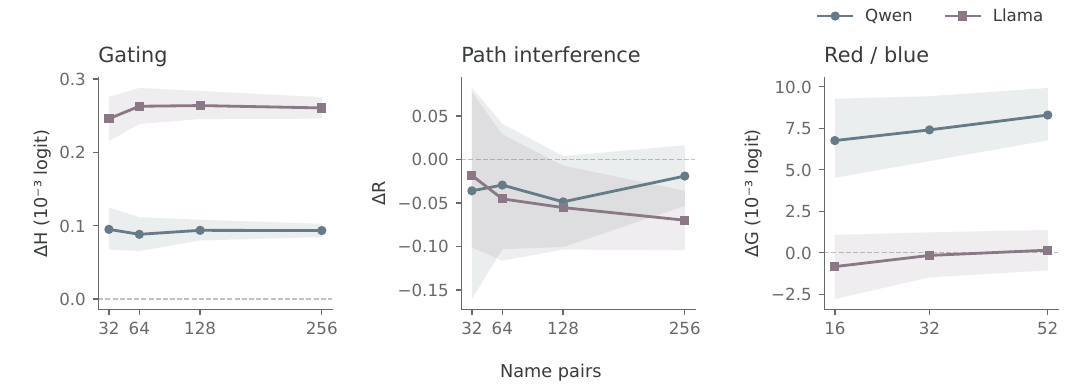}\\[0.35em]
  \includegraphics[width=0.95\textwidth]{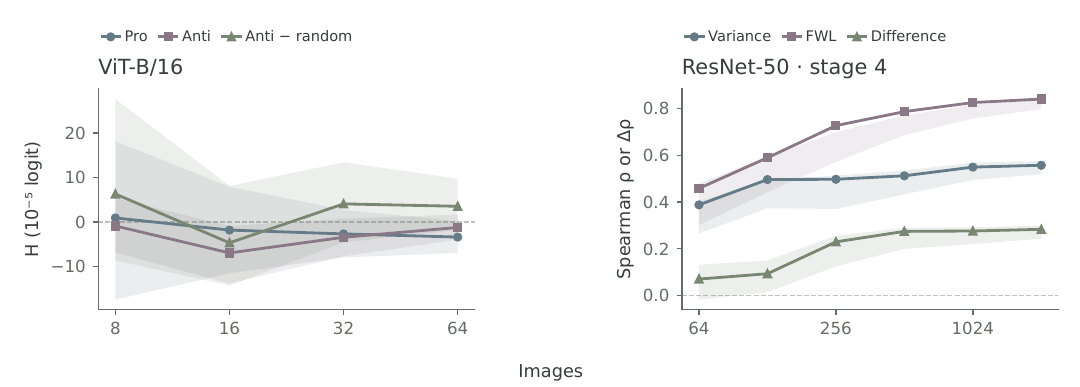}
  \caption{\textbf{Sample-size stability in Pets (top), language confirmation cohorts (middle), and historical pilots (bottom).} Top: mean background/foreground two-group interference ratios and highest-corruption shifts across nested subsets of the $444$ fresh Pets images and cumulative $572$ images ($95\%$ pointwise CIs). Middle: antagonistic-minus-matched head gating, four-path relation-minus-random $\Delta R$, and red/blue Direct-minus-Alias contrasts across nested name-pair prefixes in Qwen2.5 and Llama-3.2. Bottom: historical pilot sample-count sensitivity from the original $64$-image CUB ViT-B/16 contiguous-gate pilot (left) and $2{,}048$-image ResNet-50 Stage~4 pruning pilot (right).}
  \label{fig:s1a}
\end{figure}

\paragraph{Checkpoint provenance and cohort reuse.}
Vision experiments use task-trained Oxford-IIIT Pets ResNet-18 (\texttt{resnet18\_pets.pth}), CUB-200-2011 ResNet-18 (\texttt{resnet18\_cub200-1.pth}) and ViT-B/32 (\texttt{vit\_b\_32\_cub200-1.pth}), and pretrained ImageNet-1K checkpoints specified by experiment: ViT-B/16 (Torch Hub SHA-256 \texttt{c867db91d3e12c6cbadabb610d73c24a546bf82d8c03a9fea34f43a712ddb0e9}), \texttt{torchvision} ResNet-50 (\texttt{resnet50\_imagenet-1.pth}) for the CUB pathway-geometry analyses (\cref{fig:fig4}), and \texttt{timm} ResNet-50 (\texttt{resnet50\_a1\_0}), ViT-S/16, and Swin-T for the ImageNet pruning analyses (\cref{fig:fig7}; the same Swin-T checkpoint is used in \cref{fig:fig4}). Language experiments use frozen pretrained Qwen2.5-1.5B and Llama-3.2-1B base models with original tokenizers and deterministic CUDA configurations; paired bootstrap resampling uses $10{,}000$ draws by default, with explicit exceptions of $20{,}000$ draws for language attention-head gating and CUB regional signed Pearson correlation, $4{,}000$ draws for ImageNet pruning, and $1{,}000$ draws for Pets dose-level finite-difference calibration Spearman correlations. All independent statistical units are complete source images or complete entity name pairs; shared evaluation cohorts (CUB $1{,}024$ images across visual backbones; $256$ shared confirmation name pairs and $256$ additional shared MLP-grouping pairs across Qwen2.5 and Llama-3.2; $264$ reserve CUB images shared across ViT-B/16 and ViT-B/32 functional comparisons; and $4{,}096$ ImageNet images across all $10$ pruning cuts, with $257$ images overlapping the $3{,}200$-image historical whole-network benchmark) are never summed as independent replicates. \Cref{fig:s1a} shows sample-size stability across nested subsets in Pets, language models, and historical pilot studies; \cref{tab:s1_audit} summarizes core numerical closure audits, and \cref{tab:master_summary} provides the consolidated confirmation ledger across all study endpoints.

\begin{table}[htbp]
\centering
\caption{\textbf{Representative numerical closure, gradient-vs-finite agreement, and physical-surgery precision audits across final confirmation cohorts.} Sample size ($n$) is listed explicitly for every row.}
\label{tab:s1_audit}
\scriptsize
\setlength{\tabcolsep}{4pt}
\begin{tabularx}{\textwidth}{@{}l l r >{\raggedright\arraybackslash}X@{}}
\toprule
\textbf{Audit check} & \textbf{Model / Cut (cohort size)} & \textbf{Measured value} & \textbf{Unit / Scope} \\
\midrule
Tangent-space path sum closure & CUB ResNet-18 \texttt{layer3} / \texttt{layer4.0} ($n=1{,}024$) & $5.26\times 10^{-7}$ / $4.20\times 10^{-8}$ & max abs error \\
Tangent-space path sum closure & Qwen2.5 / Llama-3.2 4 paths ($n=256$ pairs) & $\le 2.78 \times 10^{-17}$ & max abs error \\
Coordinate-scaling covariance check & Pets ResNet-18 Wiener transport ($n=444$) & $2.32 \times 10^{-6}$ & max relative $L_2$ \\
Two-scale exact bridge error & Pets ResNet-18 ($n=128$, $896$ points) & $4.44 \times 10^{-16}$ & max abs error \\
Gram quadrature ($256$ vs.\ $384$ nodes) & R50 / Swin-T / ViT-B/16 ($n=1{,}024$) & $\le 1.16 \times 10^{-14}$ & max abs $\Rint$ \\
Finite-diff vs.\ derivative sign match & CUB ResNet-18 early / late ($n=1{,}024$) & $98.27\% / 99.44\%$ & fraction (main) \\
Finite-diff vs.\ derivative sign match & ViT-B/16 Block~6 ($n=1{,}024$, main / half) & $99.988\% / 99.994\%$ & fraction \\
Fast FWL vs.\ explicit OLS refit & ViT-S/16 \& Swin-T mid MLP ($n=4{,}096$) & $9.11 / 8.81 \times 10^{-18}$ & max abs MSE ($\le 1.01\times 10^{-14}$ all 6) \\
Physical surgery margin check & 36 settings $\times 4{,}096$ images & $147{,}456 / 147{,}456$ & passed margin tol. \\
Late Swin direct slicing fp64 replay & $732$ float32 MSE exceptions ($n=4{,}096$) & $6.22 \times 10^{-15}$ & max abs MSE (fp64) \\
Full-network vs.\ batched MLP gate & Qwen / Llama ($21{,}504$ cells each) & exact match & condition-level $H$ \\
\bottomrule
\end{tabularx}
\end{table}

\begin{table}[htbp]
\centering
\caption{\textbf{Consolidated confirmation ledger and scope boundaries.} All primary estimates use the final large-sample confirmation cohorts; exploratory pilots are archived separately across Supplementary Sections~\ref{supp:s1}--\ref{supp:s9}.}
\label{tab:master_summary}
\scriptsize
\begin{tabularx}{\textwidth}{@{}p{2.3cm} >{\raggedright\arraybackslash}p{6.6cm} >{\raggedright\arraybackslash}X@{}}
\toprule
\textbf{Question (\S)} & \textbf{Primary confirmation results (95\% CI unless noted)} & \textbf{Scope and explicit boundaries} \\
\midrule
Shared encoding across disjoint blocks (\ref{sec:exp_encoding}) & CUB (1,024): centered block matching $+0.555\text{--}+0.595$ (R18 L3, 42.1\% energy), $+0.397\text{--}+0.422$ (L4.0, 14.1\%), $+0.874\text{--}+0.884$ (ViT B6, 49.3\%; 99.375\% CIs). Pets localized tint (442): joint $p=0.0042$ (56.1\%--60.9\%). LLMs (256): all 4 blocks within-pair cross-role AUC~$=1.000$ (cos $0.792\text{--}0.844$). & Per-texture-pair cross-image GAP mean directions cover on average only $1.52\%$ energy; vision proves paired complement matching within a pre-fixed 16D subspace (not intrinsic dimensionality or single-block identity decoding). \\
Fixed-task channel synthesis (\ref{sec:exp_channels}) & Pets (444): BG $R^{(2)} = 0.906\,[0.868, 0.944]$ vs.\ FG $1.089\,[1.054, 1.125]$ (128 batch: $0.858$ vs.\ $1.122$); sign vs.\ random partition $-0.792\,[-0.821, -0.761]$; 512-ch BG $1+9.200-0.937=9.263$; 248/444 images simultaneously satisfy $R^{(2)}<1$ and $R^{(512)}>1$ (266/444 have positive within and negative between terms). & Conditional on fixed clean top-1 class, 4 donor edits, and head sign grouping; alternative top1$-$top2 margin gives BG $R^{(2)}=0.879$. \\
Multi-path geometry \& scale (\ref{sec:exp_paths}) & CUB (1,024) tex/rand $R$: R18 $1.158/1.133$ (Gram $1.1488$, diff $+0.0156$), R50 $1.754/1.377$, Swin-T $1.113/0.992$, ViT $0.284/0.277$ (Gram $0.2779$; clamped $J$ rounds to $-0.396$). Across ViT/Qwen/Llama, both $A\times J$ ($-0.157/-0.262/-0.332$) and $D\times M$ ($-0.061/-0.213/-0.108$) are negative. Splitting MLP deepens $\Delta R$ by $+0.0753$ (Qwen) and $+0.1227$ (Llama). & Swin-T $>1$ refutes universal Transformer cancellation; ViT texture minus random is $+0.006\,[-0.004, +0.017]$; ViT terminal non-\texttt{[CLS]} exact $0$ (Type~0). \\
Fixed factor, varied readout (\ref{sec:exp_conditions}) & CUB (1,024): 8-species re-pairing DiD $+0.05127\,[0.04250, 0.06016]$; first-order splits into amplitude $+0.03608$ + composition $+0.01484$ (diff $+0.02124$). LLMs (52 pairs, $10^{-3}$ logit): Qwen RB mean $\Delta G$ flips $+6.540 \to -1.756$; Llama RB $+2.368 \to +2.214$ with interaction $+2.050\,[1.192, 2.914]$. & Finite $0.5ab\,v$ injection in LLMs; prompt switch covaries wording, length, and query position; Alias RB two-choice accuracy stays high (81.7\% Qwen, 79.8\% Llama). \\
Prospective gating \& readout flip (\ref{sec:exp_gating}) & LLM heads (256, $10^{-3}$): Qwen pro $-0.132$ / anti $+0.036$ ($+0.093$ vs.\ rand); Llama $-0.179$ / $+0.153$ ($+0.261$). MLP functional groups ($10^{-5}$): ViT-B/16 B6 replay $+30.86$ ($+30.41$ vs.\ contiguous); B10 holdout $+22.23$ ($+24.62$ vs.\ cont., $+33.36$ vs.\ matched); Qwen/Llama beat tight $\pm 1\%$ matched by $+2.22$ / $+4.49$. ViT B10 same-forward means: Group~A $+19.25/-10.90$ ($97/264$ imgs), Group~B $-7.64/+40.50$ ($115/264$), 4-cell interaction $+78.30$. & Qwen $\pm 1\%$ matching is post-hoc sensitivity; Llama $\pm 1\%$ locked pre-holdout; ViT 264 reserve shared with B/32 comparison. \\
Pruning channel recovery (\ref{sec:exp_pruning}) & ImageNet (4,096, 10 cuts): FWL Spearman $0.865\text{--}1.000$ overall ($0.975\text{--}0.999$ across 6 ViT/Swin cuts); R50 stage~2 jumps $0.420 \to 0.997$ (no cross-terms $0.330$); Swin mid switches top unit $\#653$ (85.3\%) $\to \#826$ (100\%). & Holds for single-cut fixed linear successor + OLS; under direct slicing Swin post$-$pre MAE is flat ($+0.0009$), whereas under OLS both post$-$pre drop (Swin $-0.0306$, ViT $-0.1094$); whole-network \cref{tab:s7b_whole} open. \\
Corruption \& inversion bounds (\ref{sec:discussion}, \ref{supp:s8}--\ref{supp:s9}) & Pets blur5 (444): BG $R^{(2)}$ drops $0.906 \to 0.619$ while coherent BG energy rises $0.00351 \to 0.00559$ (accuracy $87.6\% \to 25.9\%$). Pets Wiener edit ranking $0.0139$ vs.\ native $\Psi$ $0.9936$ (mean norm.\ defect $\|\Dcomm\|_2/\|\Psi\|_2 = 1.0001\,[0.9992, 1.0011]$); CUB de-shared absolute margin advantage $+0.00199\,[0.00026, 0.00373]$, mean group signed $r = +0.0485$. & Relative cancellation and absolute sensitivity decouple under blur; source inversion discovers candidate regions but cannot replace native readout. \\
\bottomrule
\end{tabularx}
\end{table}

\FloatBarrier
\section{Extended Shared-Encoding Analyses}
\label{supp:s2}

\begin{figure}[p]
  \centering
  \includegraphics[width=0.84\textwidth]{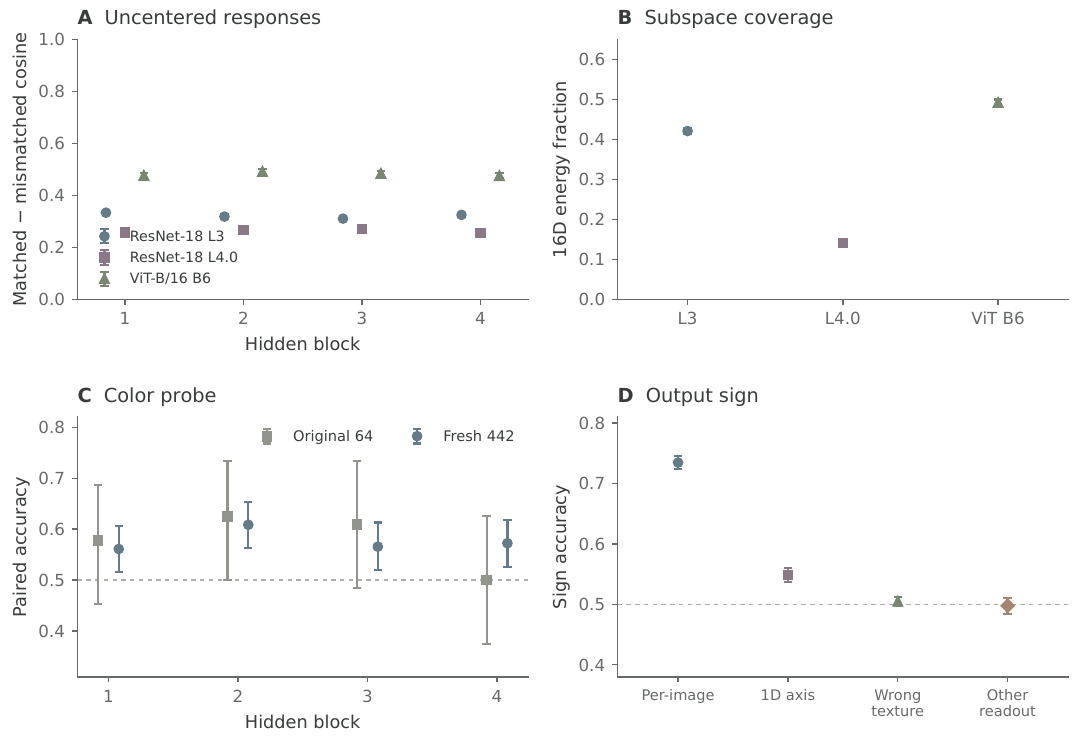}\\[0.35em]
  \includegraphics[width=0.84\textwidth]{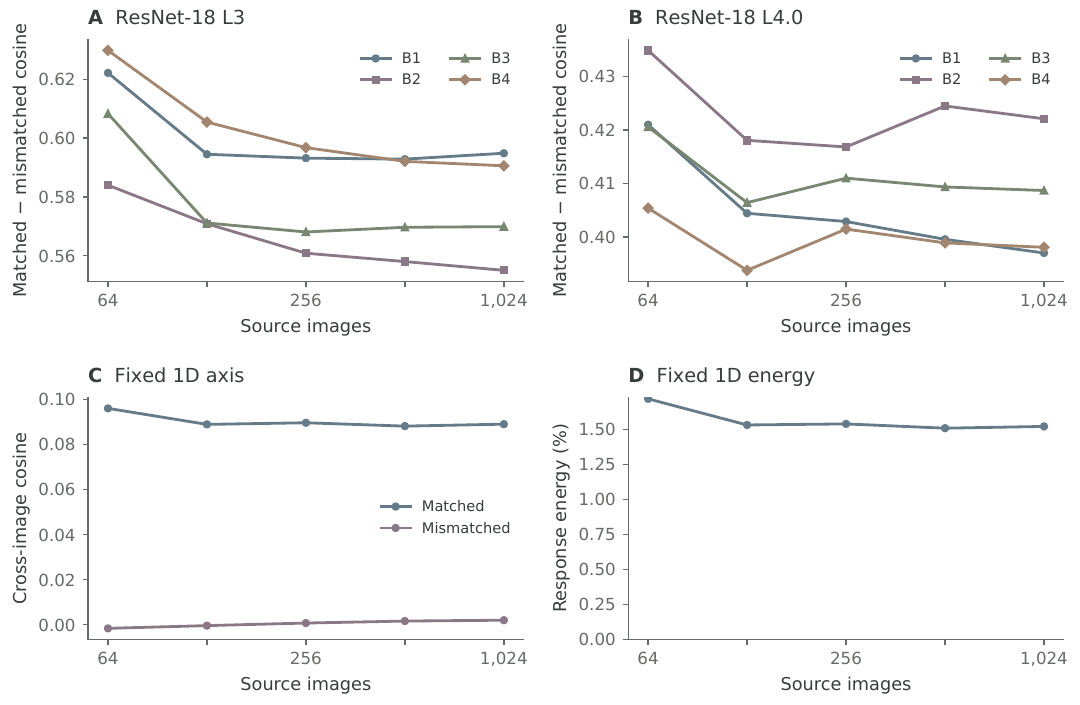}
  \caption{\textbf{Visual shared-encoding controls (top) and sample-size stability (bottom).} Top: uncentered block matching ($99.375\%$ CIs), cut-specific $16$D subspace energy fractions, weak localized warm/cool Pets tint probe ($n=64$ vs.\ $n=442$), and CUB output-sign forecast comparison ($n=1{,}024$). Bottom: nested CUB prefixes for centered block matching at ResNet-18 \texttt{layer3}/\texttt{layer4.0} and terminal GAP 1D axis alignment and energy share.}
  \label{fig:s2a}
\end{figure}

\begin{figure}[htbp]
  \centering
  \includegraphics[width=0.95\textwidth]{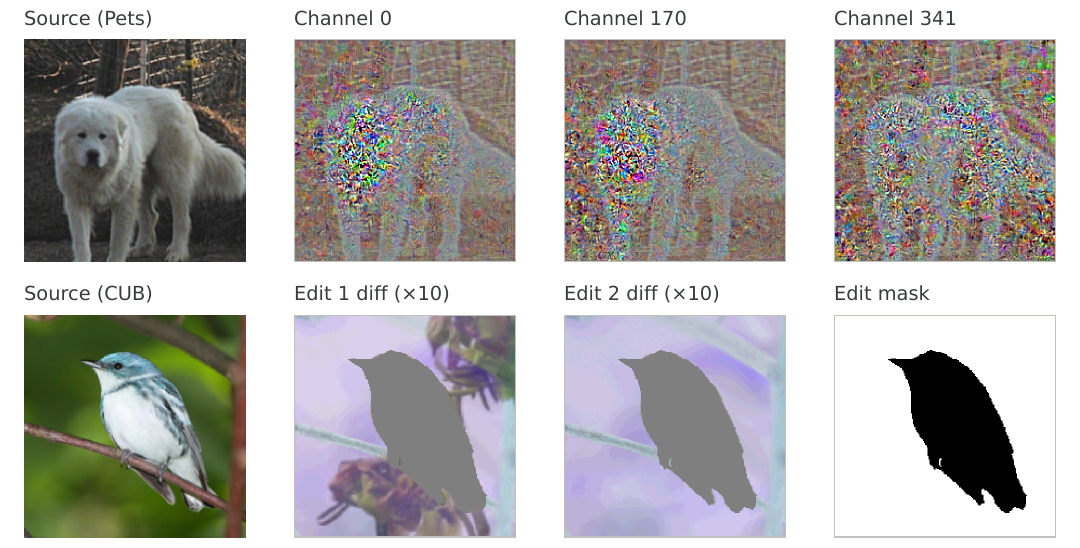}\\[0.35em]
  \includegraphics[width=0.95\textwidth]{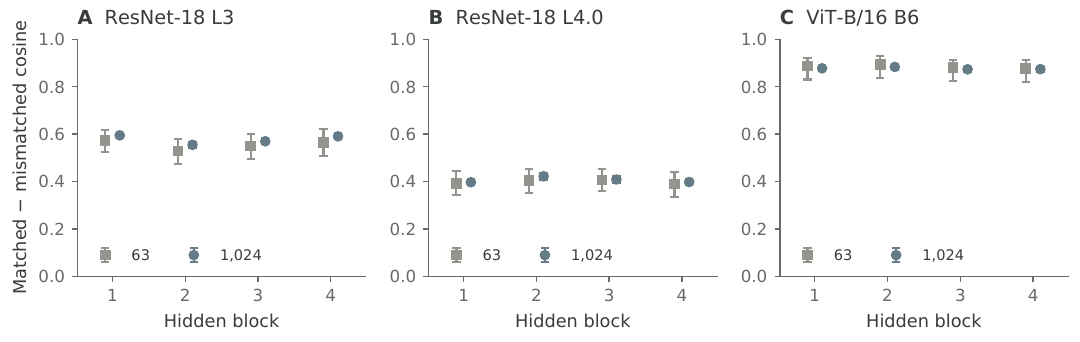}
  \caption{\textbf{Visual source fields (top) and independent cohort replication (bottom).} Top: frozen Pets source \texttt{great\_pyrenees\_9} with LAC fields for channels $0, 170,$ and $341$ (centered at mid-grey and scaled to $99$th percentile magnitude for display), followed by \texttt{CUB-09631}, $10\times$ amplified signed pixel differences (centered at mid-grey) for two distinct $5\%$ background-texture edits confirming strictly zero foreground change, and the editable background mask. Bottom: centered disjoint-block matching compared across the independent $63$-image exploratory and $1{,}024$-image confirmation CUB cohorts ($99.375\%$ CIs).}
  \label{fig:s2b}
\end{figure}

\Cref{fig:s2a,fig:s2b} expand \cref{sec:exp_encoding} across four complementary diagnostics: (i) uncentered versus mean-centered visual matching and cut-specific $16$D subspace energy coverages ($42.14\%$ ResNet-18 \texttt{layer3}, $14.09\%$ \texttt{layer4.0}, $49.26\%$ ViT-B/16 Block~6); (ii) comparison of the per-texture-pair cross-image mean directions at terminal GAP (averaging $1.52\%$ of held-out response energy and $54.86\%$ output sign agreement, versus $73.47\%$ for the per-image native directional derivative, which drops to $50.50\%$ under a mismatched texture pair and $49.77\%$ under a reassigned readout); (iii) statistical-power progression of the weak Pets localized warm/cool tint probe from the original $64$ images ($37, 40, 39, 32$; joint $p=0.546$) to $442$ newly eligible images ($248, 269, 250, 253$; $56.1\%\text{--}60.9\%$, joint $p=0.0042$); and (iv) replication across independent $63$-image exploratory and $1{,}024$-image confirmation CUB cohorts alongside source-image visual examples (\cref{fig:s2b}).

\FloatBarrier
\section{Extended Two-Scale Channel Synthesis on Pets}
\label{supp:s3}

\begin{figure}[p]
  \centering
  \includegraphics[width=0.86\textwidth]{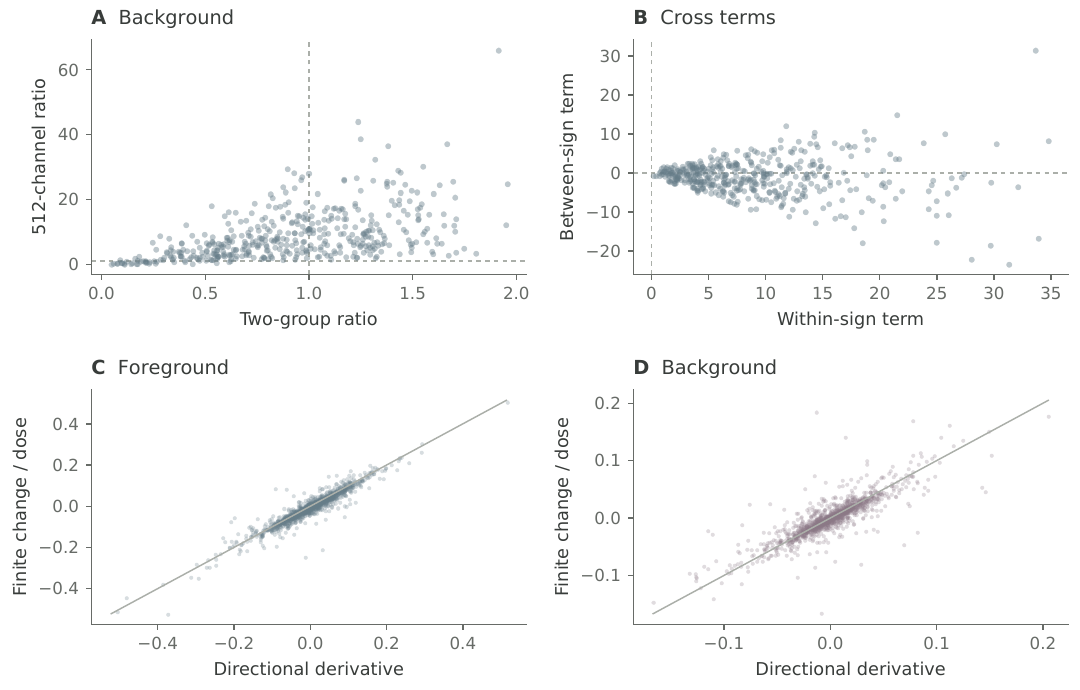}\\[0.35em]
  \includegraphics[width=0.86\textwidth]{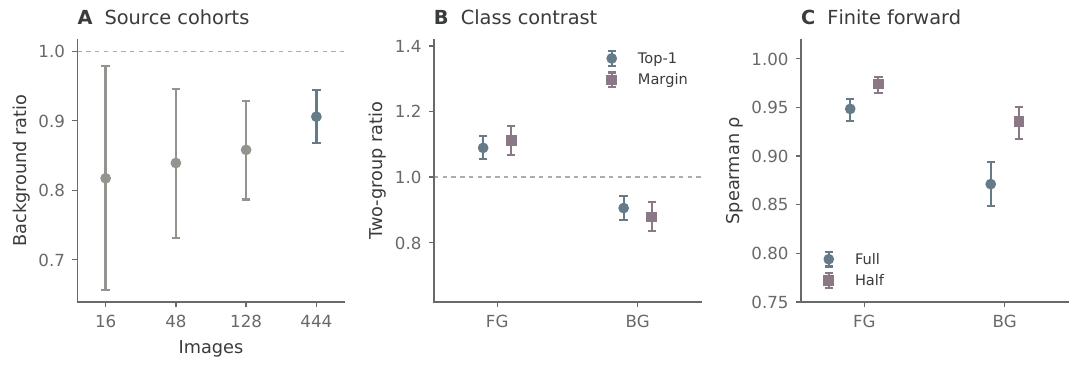}\\[0.35em]
  \includegraphics[width=0.86\textwidth]{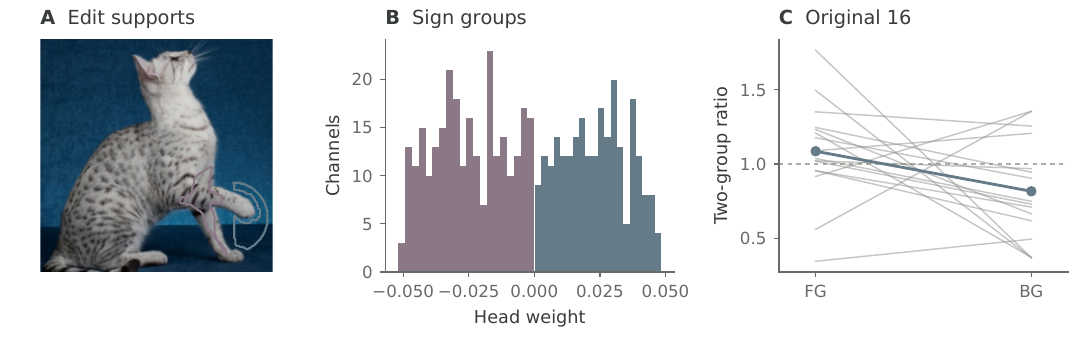}
  \caption{\textbf{Image-level channel decomposition and finite-dose validation (top), cohort/readout controls (middle), and edit supports with the original $16$-image discovery set (bottom).} Top: per-image two-group versus $512$-channel interference ratios and signed cross-terms ($n=444$; $248/444$ satisfy $R^{(2)}<1$ and $R^{(512)}>1$ simultaneously), plus donor-level directional derivative versus full-dose ($0.125$) finite logit changes. Middle: cohort provenance ($16, 48, 128, 444$), clean top-1 versus top-1 minus top-2 margin readouts, and full/half-dose derivative correlations. Bottom: first confirmation source (\texttt{Egyptian\_Mau\_9}) with foreground/background donor supports, head-weight histogram ($243$ positive, $269$ negative), and paired ratios across the original $16$ sources.}
  \label{fig:s3a}
\end{figure}

\Cref{fig:s3a} provides the complete cohort lineage, alternative-readout controls, edit supports, and image-level validation supporting \cref{sec:exp_channels}. Across cohorts, historical background $R_{\mathrm{bg}}^{(2)}$ values are $0.817$ (original $16$ sources), $0.839$ (remaining $48$ sources from the original $64$-image batch), and $0.834$ (combined original $64$; never double-counted as three separate replications), followed by independent confirmation batches of $0.858\,[0.787, 0.929]$ ($128$ new sources) and $0.906\,[0.868, 0.944]$ ($444$ new sources). Changing the readout to the fixed top-1 minus top-2 margin preserves cancellation ($R_{\mathrm{bg}}^{(2)} = 0.879$), and derivative-to-finite Spearman rank correlations reach $0.871$ (RGB dose $0.125$) and $0.936$ (half dose $0.0625$).

\FloatBarrier
\section{Extended Multi-Pathway Geometry, Controls, and Falsification Archive}
\label{supp:s4}

\begin{figure}[htbp]
  \centering
  \includegraphics[page=3,width=0.95\textwidth]{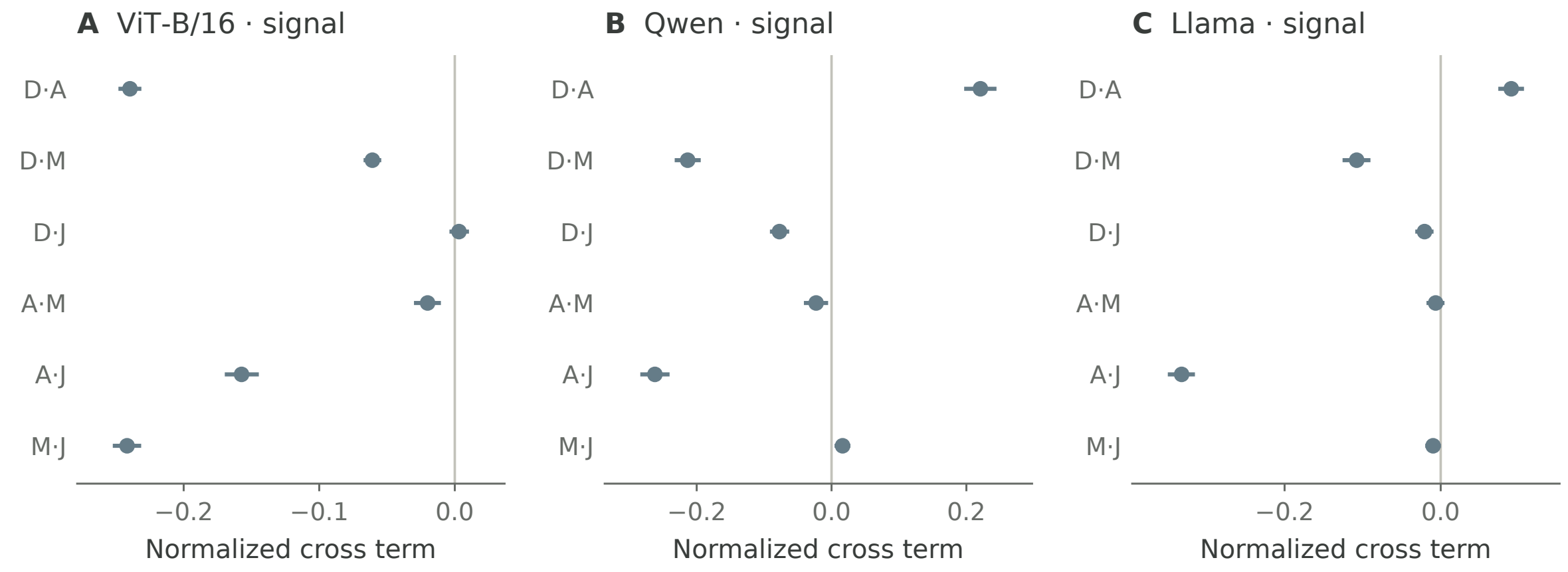}
  \caption{\textbf{Six-pair signal pathway cross-terms across vision and language Transformers.} Measured texture (ViT-B/16, $1{,}024$ images) and relation (Qwen2.5-1.5B, Llama-3.2-1B, $256$ pairs) normalized cross-terms with $95\%$ CIs, visually accompanying \cref{tab:s4_cross}.}
  \label{fig:s4_cross_sig}
\end{figure}

\begin{figure}[p]
  \centering
  \includegraphics[page=1,width=0.94\textwidth]{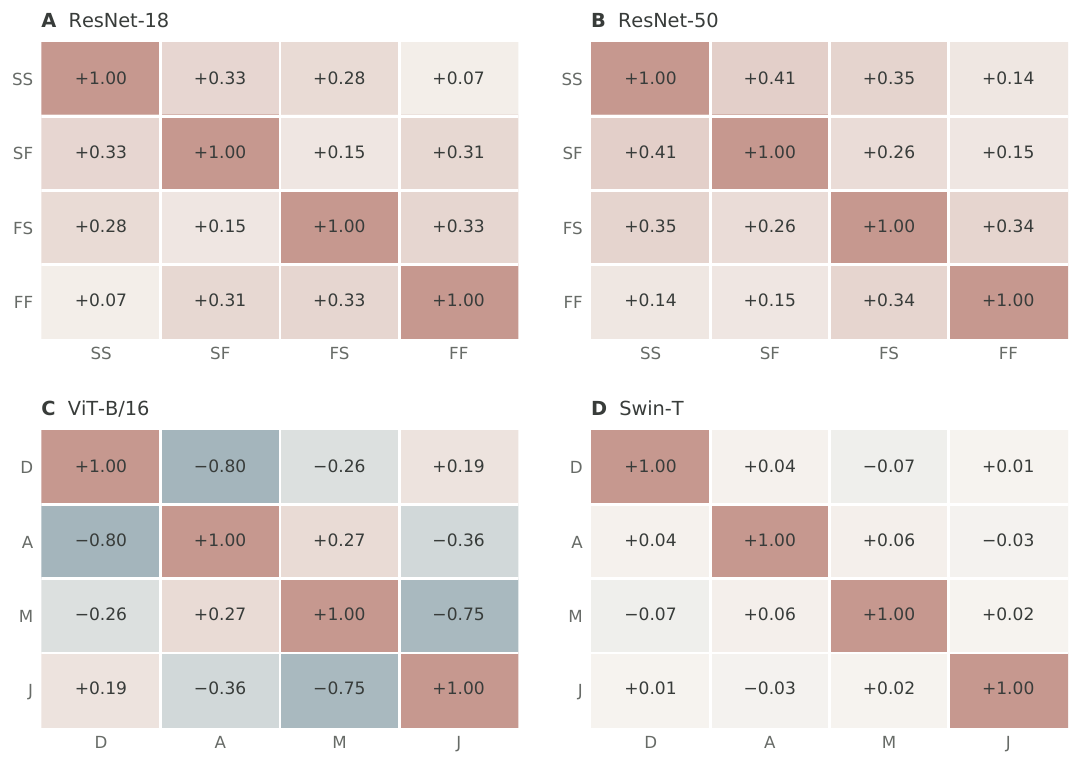}\\[0.35em]
  \includegraphics[page=2,width=0.94\textwidth]{S4_path_geometry.pdf}
  \caption{\textbf{Complete path-gradient Gram geometry across four visual backbones ($1{,}024$ CUB images).} Pairwise path-gradient cosines, gradient energy fractions, exact Gaussian-integral expectations, and trace-ratio approximations across ResNet-18, ResNet-50 (paths $SS, SF, FS, FF$), Swin-T, and ViT-B/16 (paths $D, A, M, J$).}
  \label{fig:s4_geom}
\end{figure}

\begin{figure}[p]
  \centering
  \includegraphics[page=4,width=0.90\textwidth]{S4_path_geometry.pdf}\\[0.35em]
  \includegraphics[page=10,width=0.90\textwidth]{S4_path_geometry.pdf}\\[0.35em]
  \includegraphics[width=0.90\textwidth]{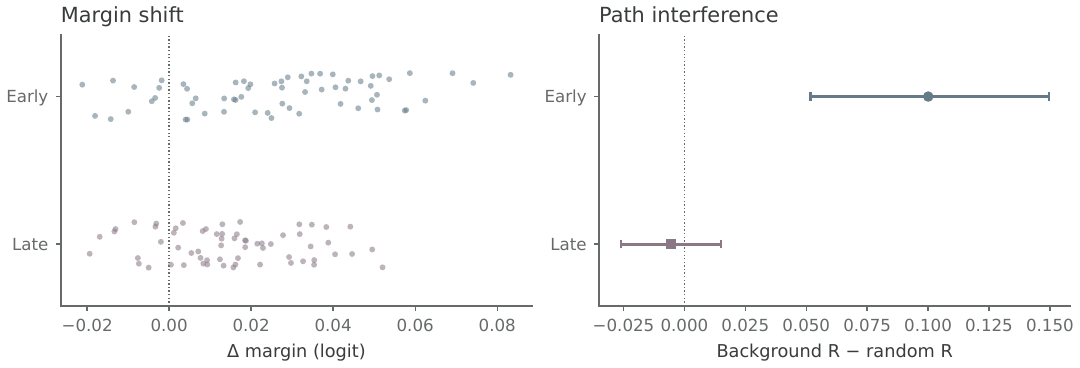}
  \caption{\textbf{Random-direction six-pair cross-terms (top), initial-versus-final replication (middle), and exploratory ResNet-18 pilot (bottom).} Top: matched random-direction normalized cross-terms ($95\%$ CIs). Middle: initial-versus-final cohort replication of signed signal cross-terms. Bottom: original $63$-image ResNet-18 injection pilot (P1, early four-path and late two-path cuts).}
  \label{fig:s4_cross}
\end{figure}

\begin{figure}[p]
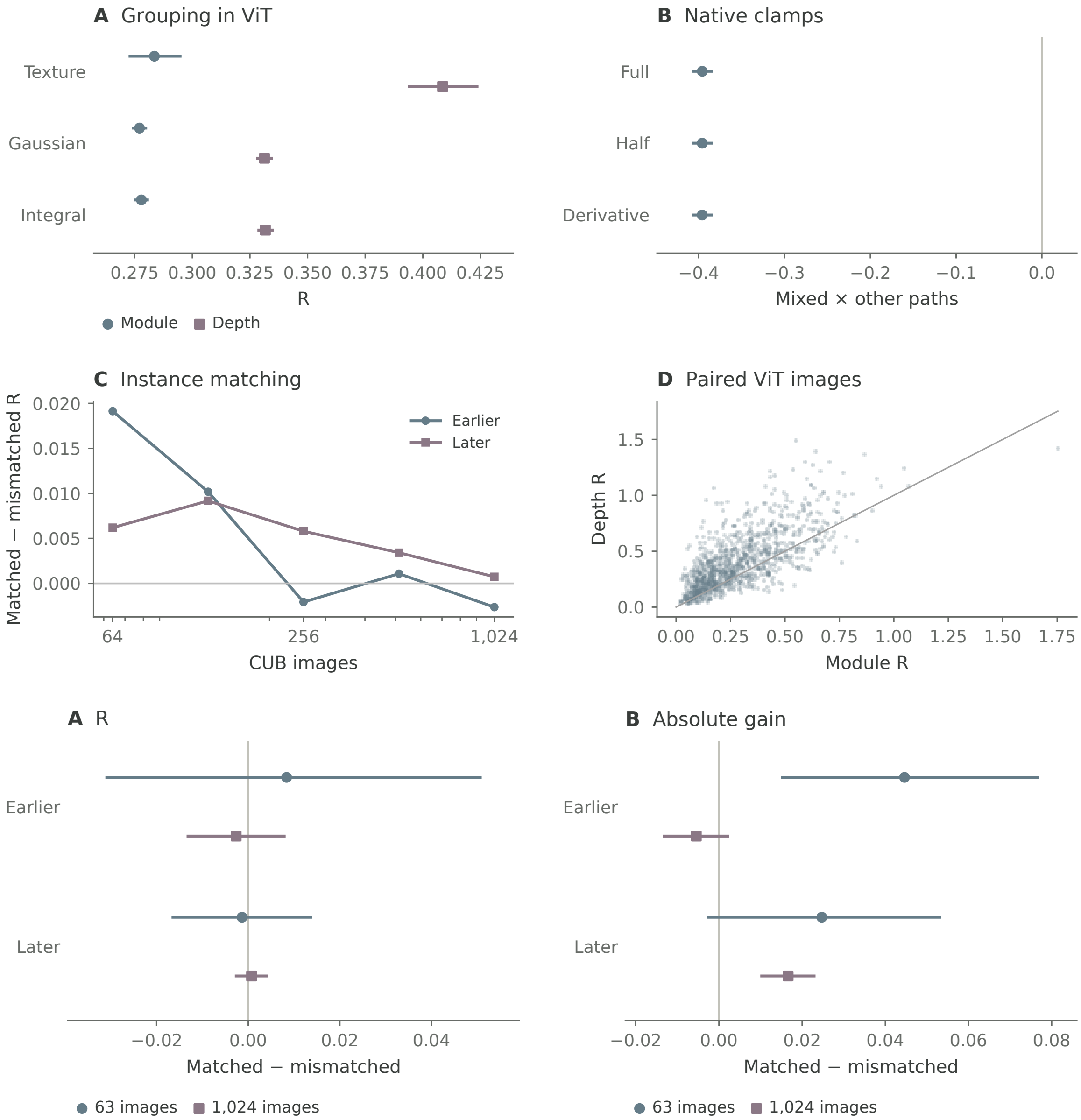

  \centering
  \includegraphics[page=5,width=0.94\textwidth]{S4_path_geometry.pdf}\\[0.35em]
  \includegraphics[page=6,width=0.94\textwidth]{S4_path_geometry.pdf}
  \caption{\textbf{ViT regrouping, forward clamping, and same-versus-wrong image controls.} Top: module-versus-depth regrouping in ViT-B/16 Block~6, native clamped mixed-path cross-term (rounding to $-0.396$), ResNet-18 instance matching, and paired ViT source comparisons. Bottom: comparison across two independent CUB cohorts on same-image minus wrong-image effects.}
  \label{fig:s4_regroup}
\end{figure}

\begin{figure}[p]
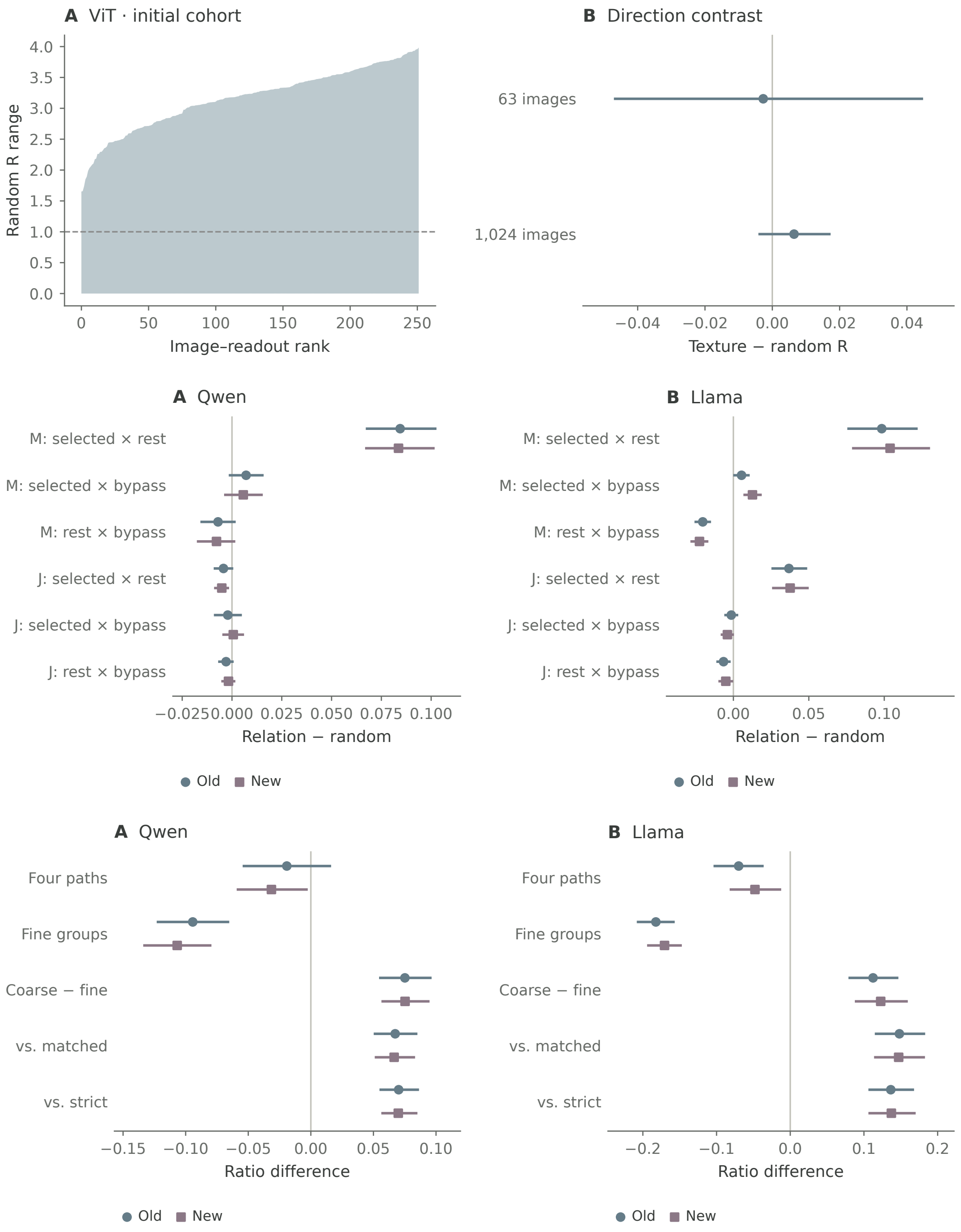

  \centering
  \includegraphics[page=7,width=0.86\textwidth]{S4_path_geometry.pdf}\\[0.35em]
  \includegraphics[page=8,width=0.86\textwidth]{S4_path_geometry.pdf}\\[0.35em]
  \includegraphics[page=9,width=0.86\textwidth]{S4_path_geometry.pdf}
  \caption{\textbf{ViT random-direction range (top) and coarse-to-fine MLP pathway splitting in Qwen2.5 and Llama-3.2 (middle and bottom).} Top: range of $256$ random directions per image--readout combination ($4{,}094/4{,}096$ span both sides of $R=1$ in the final cohort) and initial/final texture-minus-random tests. Middle: breakdown of $M$-path and $J$-path cross-terms separating selected functional MLP units, remaining same-layer units, and bypass paths. Bottom: complete old and new $256$-pair coarse-versus-fine $\Delta R$ contrasts and paired advantages over stratified and tight $\pm 1\%$ strength-matched controls.}
  \label{fig:s4_fine}
\end{figure}

\begin{figure}[p]
  \centering
  \includegraphics[page=1,width=0.77\textwidth]{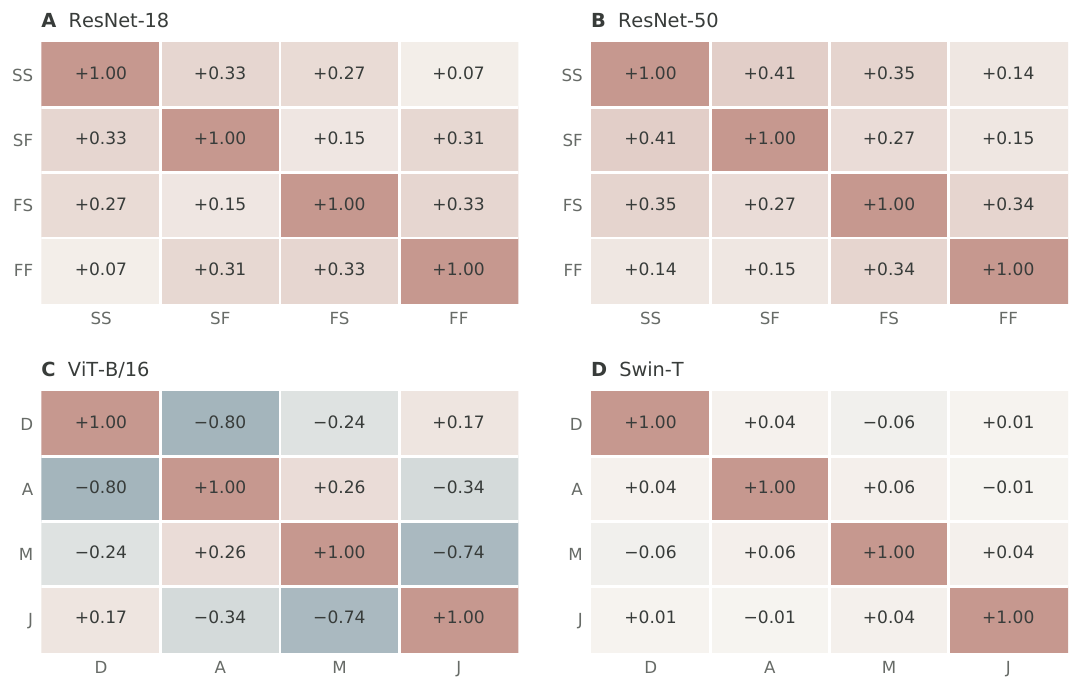}\\[0.35em]
  \includegraphics[page=2,width=0.77\textwidth]{S4_initial_geometry.pdf}\\[0.35em]
  \includegraphics[page=3,width=0.77\textwidth]{S4_initial_geometry.pdf}
  \caption{\textbf{Initial-cohort pathway geometry and nested visual/language confirmation stability.} Top and middle: initial-cohort Gram cosines, measured directional interference ratios, and gradient energy shares across visual backbones. Bottom: nested visual confirmation curves and analytic, finite, and frozen-norm language $\Delta R$ comparisons.}
  \label{fig:s4_init}
\end{figure}

This section archives the full six-pair normalized cross-term table across all three Transformers (\cref{tab:s4_cross}) and all fourteen multi-pathway diagnostic plates (\cref{fig:s4_cross_sig,fig:s4_geom,fig:s4_cross,fig:s4_regroup,fig:s4_fine,fig:s4_init}): full four-backbone path-gradient Gram geometry, signal and random six-pair cross-terms (plus initial-vs-final replication), ViT-B/16 module-versus-depth regrouping and clamped forward validation, two transparently archived unsupported/unresolved exploratory tests---the polarity-alignment blind test in ViT-B/16 (archived internally as H-path; texture minus random $+0.006\,[-0.004, +0.017]$ on $1{,}024$ CUB images, with $4{,}094/4{,}096$ combinations spanning both sides of $R=1$) and same-image versus wrong-image texture injection in ResNet-18 (where $R$ differences cross zero at both cuts, $-0.0026$ early and $+0.0007$ late, while late absolute gain shows a same-image advantage of $+0.0166$, $97.5\%$ CI $[0.0090, 0.0242]$)---the original $63$-image ResNet-18 injection pilot, the complete Qwen2.5 and Llama-3.2 coarse-versus-fine MLP pathway breakdown, and initial-cohort Gram summaries.

\begin{table}[htbp]
\centering
\caption{\textbf{Full six-pair normalized pathway cross-terms across vision and language Transformers.} Each entry is independently rounded to three decimal places from full-precision summaries (exact unrounded cross-terms sum with $+1$ to $R$). ViT-B/16 uses $1{,}024$ CUB images; Qwen2.5-1.5B and Llama-3.2-1B use the first $256$-pair confirmation cohort.}
\label{tab:s4_cross}
\scriptsize
\begin{tabular}{@{}l r r r r r r r@{}}
\toprule
\textbf{Model \& direction} & $D\times A$ & $D\times M$ & $D\times J$ & $A\times M$ & $A\times J$ & $M\times J$ & $R$ \\
\midrule
ViT-B/16 texture ($1{,}024$) & $-0.240$ & $-0.061$ & $+0.003$ & $-0.020$ & $-0.157$ & $-0.242$ & $0.284$ \\
ViT-B/16 random & $-0.304$ & $-0.082$ & $+0.029$ & $+0.058$ & $-0.121$ & $-0.303$ & $0.277$ \\
Qwen2.5 relation ($256$) & $+0.221$ & $-0.213$ & $-0.077$ & $-0.023$ & $-0.262$ & $+0.016$ & $0.662$ \\
Qwen2.5 random & $+0.231$ & $-0.170$ & $-0.084$ & $-0.022$ & $-0.295$ & $+0.020$ & $0.681$ \\
Llama-3.2 relation ($256$) & $+0.090$ & $-0.108$ & $-0.021$ & $-0.006$ & $-0.332$ & $-0.010$ & $0.614$ \\
Llama-3.2 random & $+0.044$ & $-0.064$ & $-0.004$ & $+0.051$ & $-0.303$ & $-0.040$ & $0.684$ \\
\bottomrule
\end{tabular}
\end{table}

\FloatBarrier
\section{Extended Conditional Readout Diagnostics}
\label{supp:s5}

\begin{figure}[htbp]
  \centering
  \includegraphics[page=1,width=0.94\textwidth]{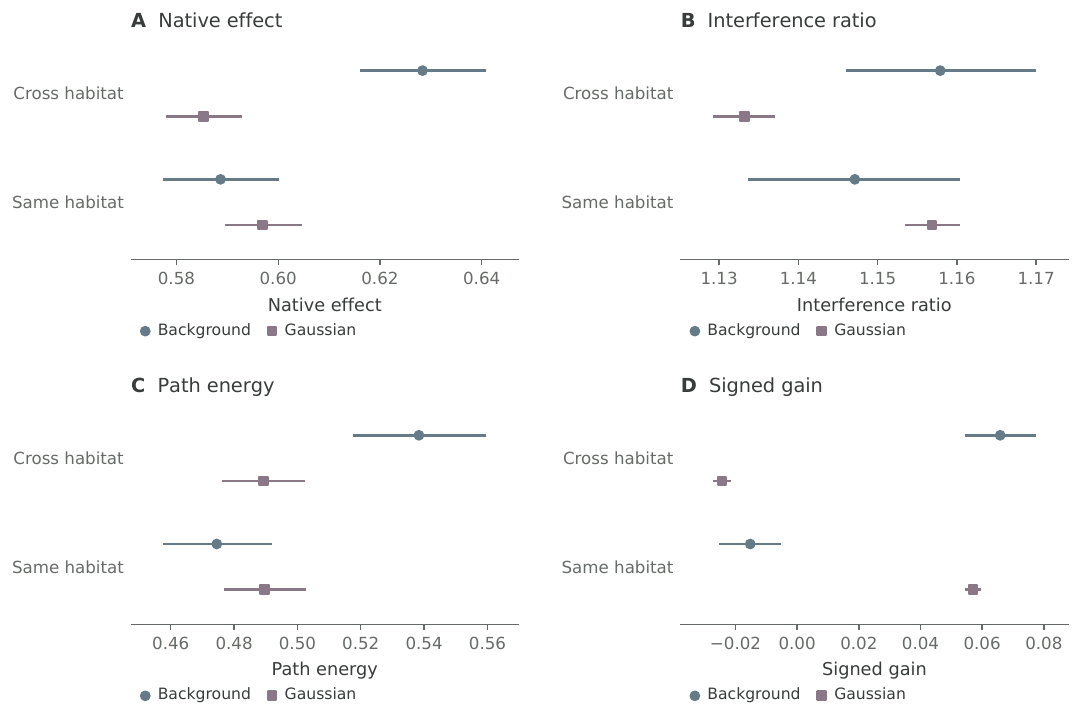}
  \caption{\textbf{CUB eight-species readout re-pairing cell diagnostics ($1{,}024$ images).} Separate cross-habitat and within-habitat cells for background and matched Gaussian edits across absolute gain, interference ratio $R$, incoherent path energy $E$, and signed gain ($95\%$ image-level CIs).}
  \label{fig:s5_cub}
\end{figure}

\begin{figure}[p]
  \centering
  \includegraphics[page=2,width=0.77\textwidth]{S5_readout_conditions.pdf}\\[0.35em]
  \includegraphics[page=1,width=0.77\textwidth]{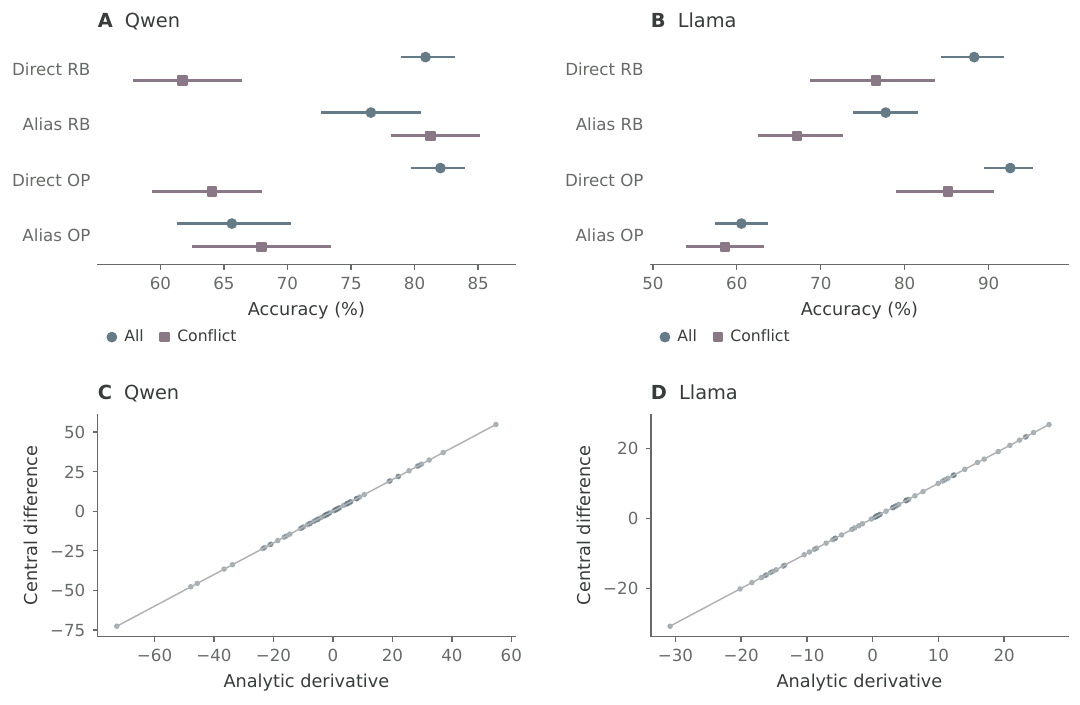}\\[0.35em]
  \includegraphics[page=2,width=0.77\textwidth]{S5_initial_controls.pdf}
  \caption{\textbf{CUB dose/sample-size decomposition (top) and initial-cohort controls in language (middle) and vision (bottom).} Top: full- and half-dose finite effects across $63$- and $1{,}024$-image CUB cohorts and nested trajectories of the symmetric amplitude--composition decomposition. Middle: per-cell accuracy in the initial $32$-pair language cohort and analytic versus central-difference derivative verification ($10^{-3}$ logit per unit dose). Bottom: absolute finite gains across all eight fixed CUB class contrasts in the initial $63$-image cohort.}
  \label{fig:s5_init}
\end{figure}

\begin{figure}[p]
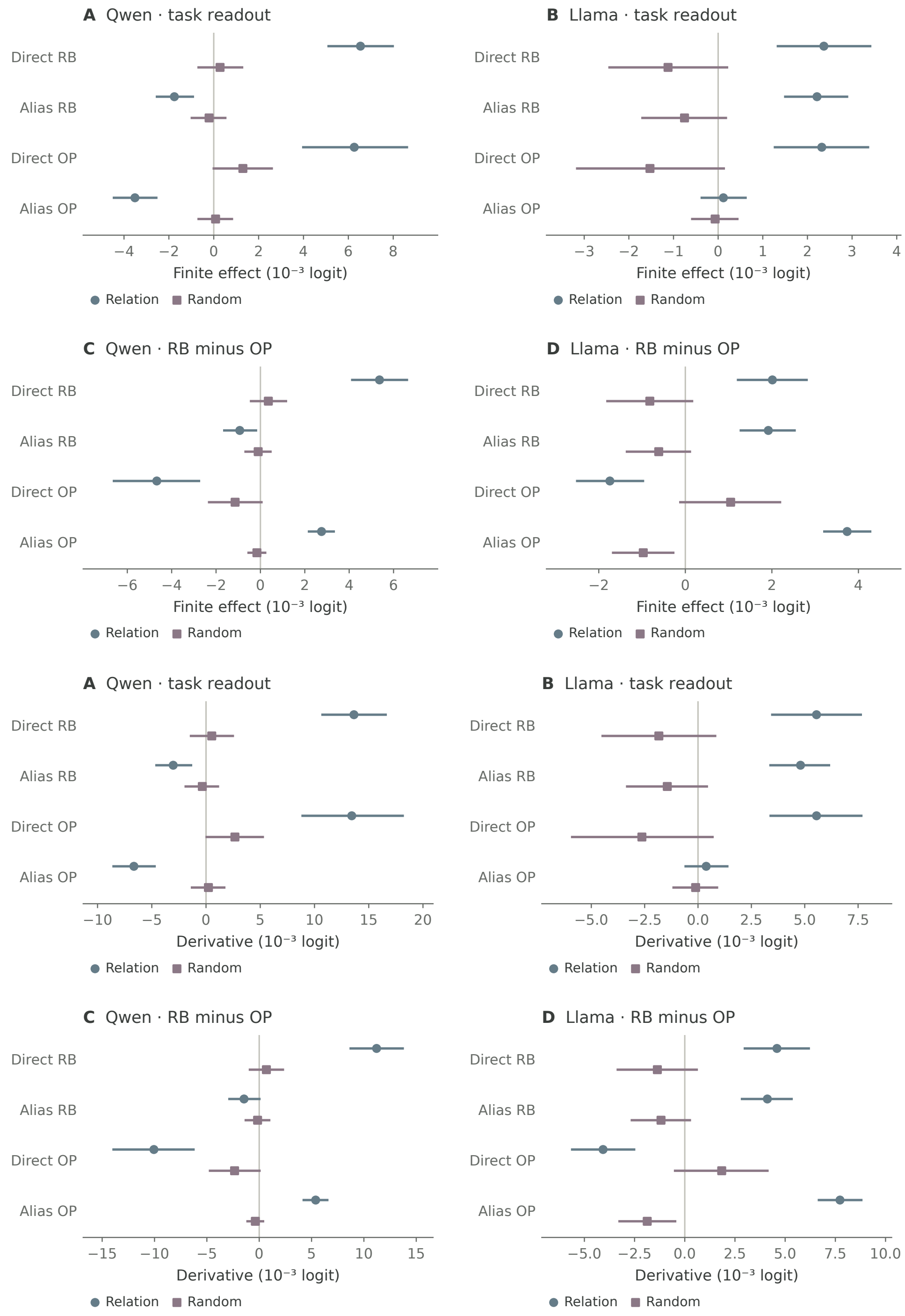

  \centering
  \includegraphics[page=3,width=0.78\textwidth]{S5_readout_conditions.pdf}\\[0.35em]
  \includegraphics[page=4,width=0.78\textwidth]{S5_readout_conditions.pdf}
  \caption{\textbf{Language finite versus first-order derivative readout comparisons ($52$ shared pairs).} Declared readout (upper rows) and difference between two readouts of the same forward pass (lower rows) for relation and matched random directions across Qwen2.5-1.5B (top plate) and Llama-3.2-1B (bottom plate).}
  \label{fig:s5_deriv}
\end{figure}

\begin{figure}[p]
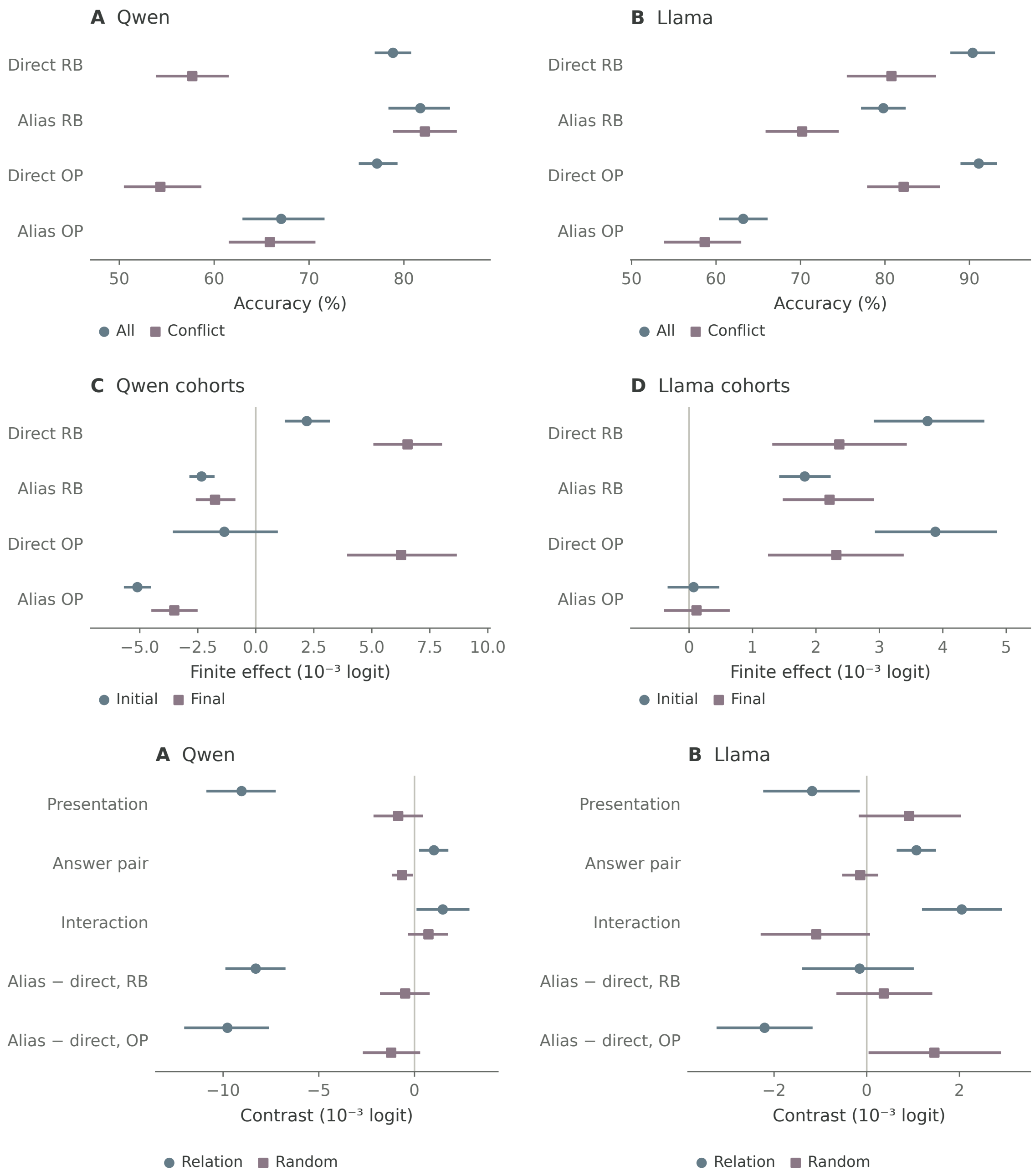

  \centering
  \includegraphics[page=5,width=0.92\textwidth]{S5_readout_conditions.pdf}\\[0.35em]
  \includegraphics[page=6,width=0.92\textwidth]{S5_readout_conditions.pdf}
  \caption{\textbf{Language $2\times 2$ two-choice accuracy controls and factorial contrasts.} Top: overall and conflict-trial two-choice candidate accuracies across all four cells in Qwen2.5 and Llama-3.2 ($52$ shared pairs; core Alias red/blue accuracies $81.7\%$ and $79.8\%$), alongside $32$-pair versus $52$-pair finite effect replication. Bottom: presentation main effect, answer-vocabulary main effect, interaction, and within-vocabulary presentation contrasts ($95\%$ paired name-level CIs).}
  \label{fig:s5_llm}
\end{figure}

\Cref{fig:s5_cub,fig:s5_init,fig:s5_deriv,fig:s5_llm} archive all eight conditional-readout diagnostic plates backing \cref{sec:exp_conditions}. In CUB ($1{,}024$ images), looking solely at texture $R$ between cross- and within-habitat pairings crosses zero, whereas subtracting matched-norm random directions gives an $R$ DiD of $+0.03445\,[0.01939, 0.04940]$ (the two comparisons ask different questions). In language ($52$ dual-tokenizer confirmation pairs alongside the $32$-pair exploratory cohort), the plates detail declared-readout versus same-forward two-readout finite and derivative estimates, per-cell two-choice candidate accuracies (including conflict trials), cohort replication from $32$ to $52$ pairs, presentation and vocabulary main effects, interactions, within-pair presentation contrasts, and initial-cohort derivative checks.

%% file: supplementary/S11_to_S14_experiments.tex
\section{Extended Gating Curves, Functional Grouping, and Strict Strength Matching}
\label{supp:s6}

\begin{figure}[p]
  \centering
  \includegraphics[page=1,width=0.86\textwidth]{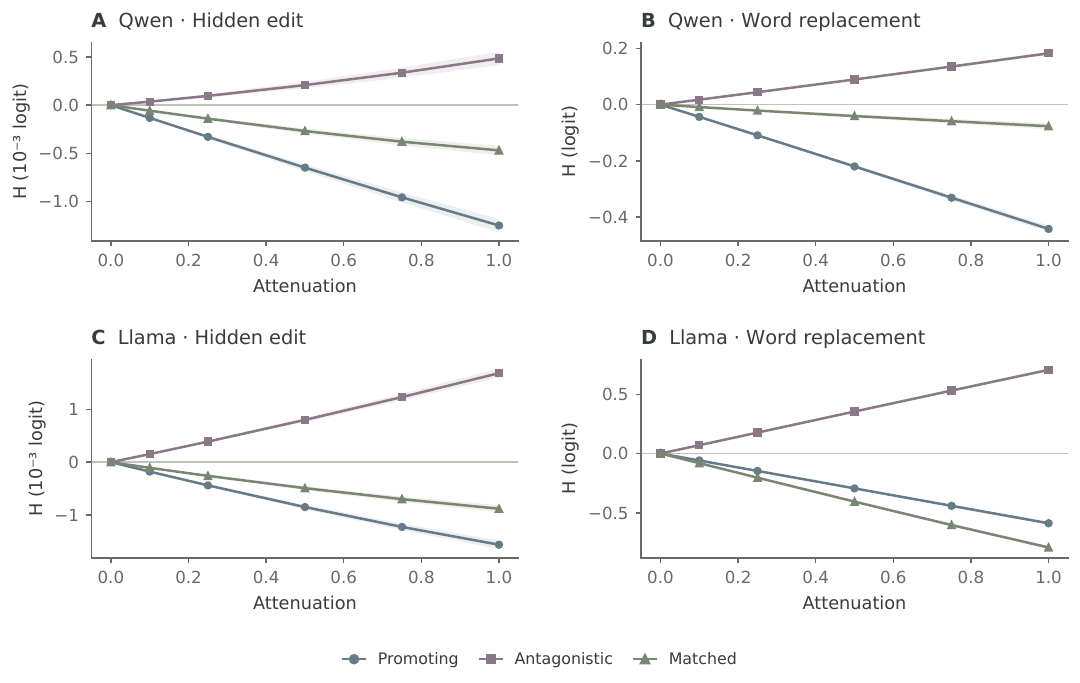}\\[0.35em]
  \includegraphics[page=2,width=0.86\textwidth]{S6_gating_and_groups.pdf}
  \caption{\textbf{Language attention-head six-dose curves under increment attenuation (top) and physical output scaling (bottom).} Evaluated on $256$ held-out name pairs across six doses ($\lambda \in \{0, 0.1, 0.25, 0.5, 0.75, 1.0\}$) for both hidden edits ($10^{-3}$ logit) and real lexical relation word replacements (raw logit) in Qwen2.5-1.5B and Llama-3.2-1B ($95\%$ name-level CIs).}
  \label{fig:s6_llm_dose}
\end{figure}

\begin{figure}[p]
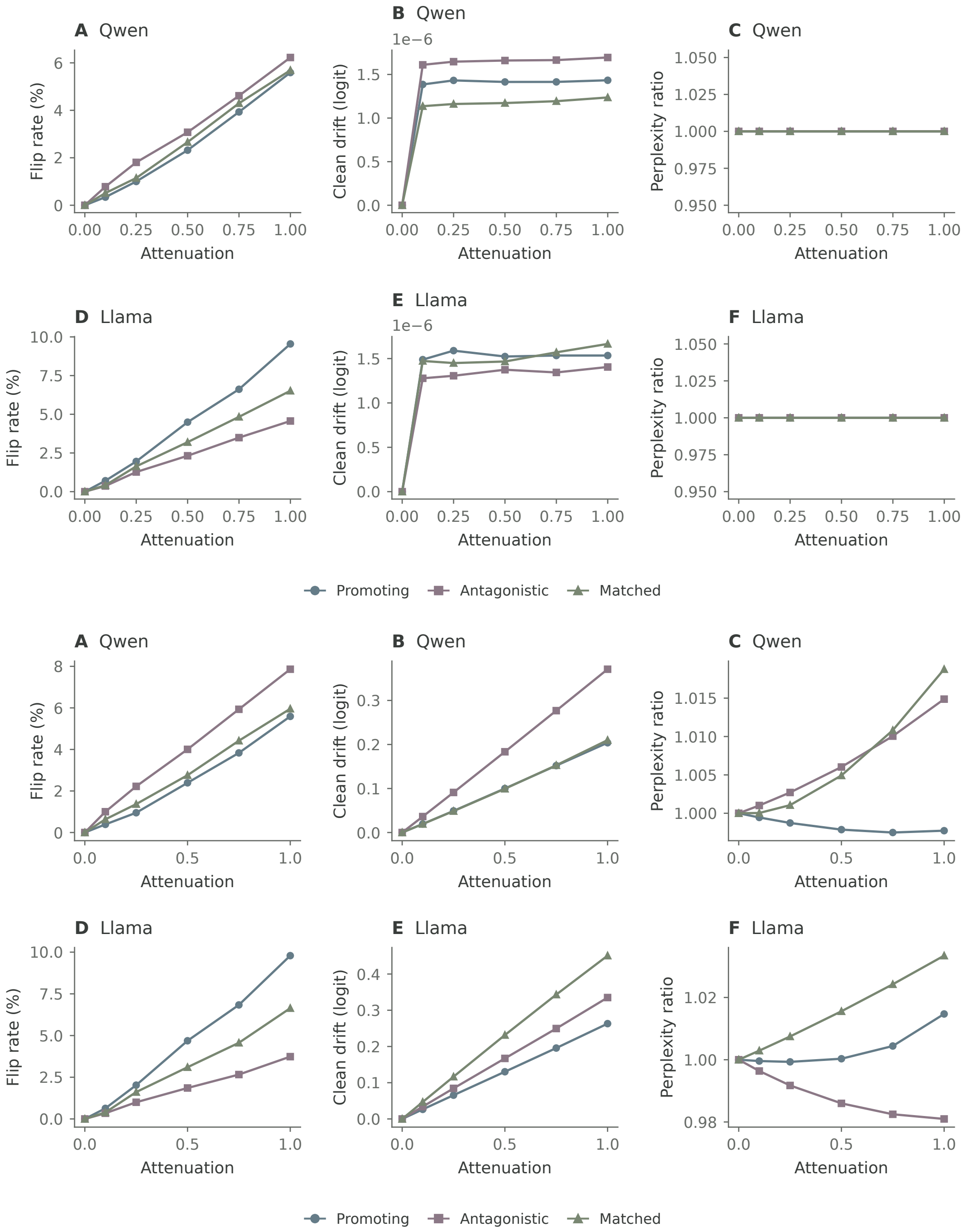

  \centering
  \includegraphics[page=3,width=0.86\textwidth]{S6_gating_and_groups.pdf}\\[0.35em]
  \includegraphics[page=4,width=0.86\textwidth]{S6_gating_and_groups.pdf}
  \caption{\textbf{Clean-operating-point stability and flip-rate diagnostics in frozen language models ($256$ pairs).} Margin flip rates, absolute clean baseline drift, and clean-prefix perplexity ratios across six doses under increment attenuation (top; drift $\le 1.7\times 10^{-6}$ logit, perplexity ratio $1.000$) versus physical output scaling (bottom; appreciable clean drift motivating increment attenuation as the primary intervention).}
  \label{fig:s6_llm_drift}
\end{figure}

\begin{figure}[p]
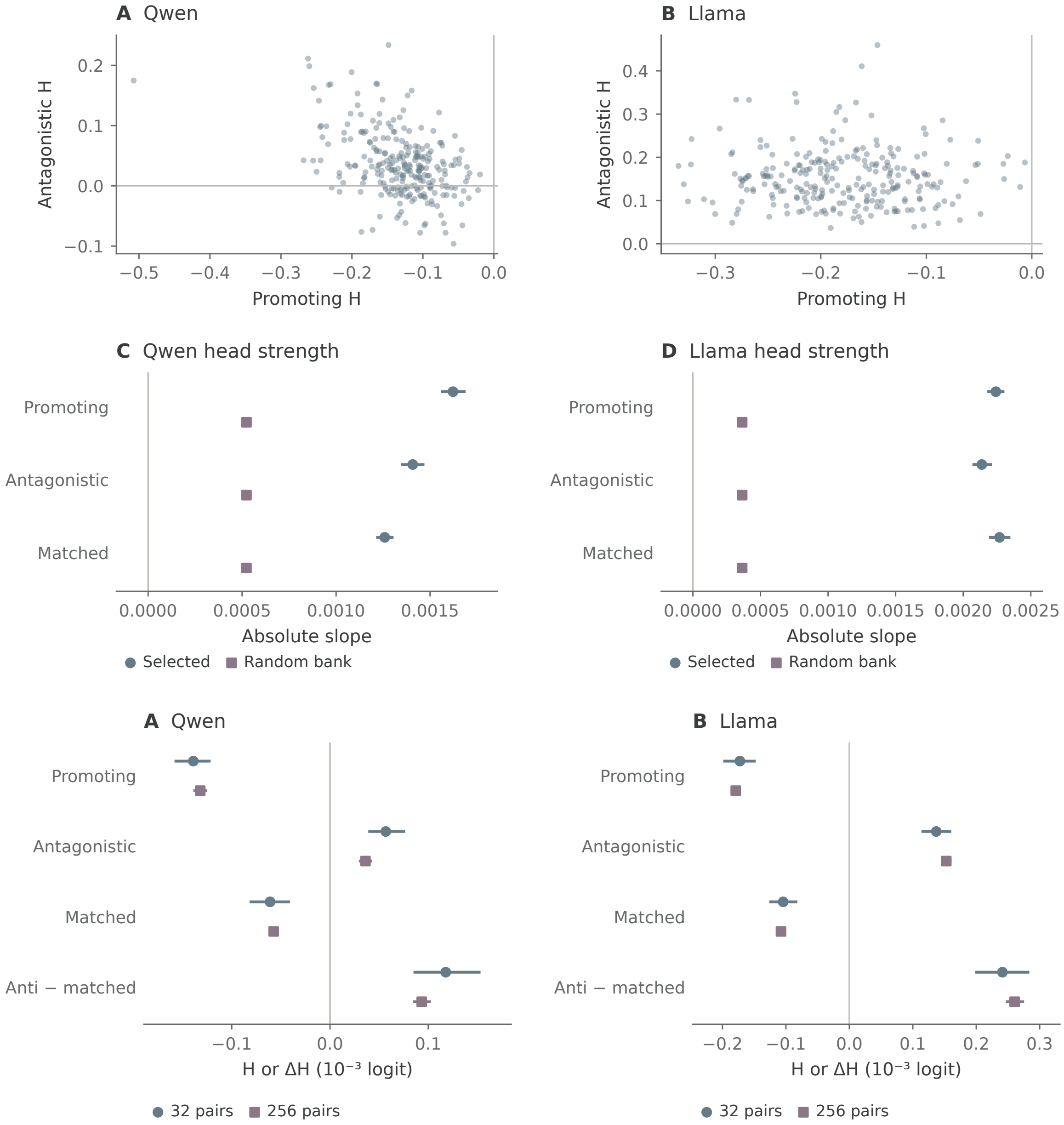

  \centering
  \includegraphics[page=5,width=0.94\textwidth]{S6_gating_and_groups.pdf}\\[0.35em]
  \includegraphics[page=6,width=0.94\textwidth]{S6_gating_and_groups.pdf}
  \caption{\textbf{Per-name-pair gating distributions, strength banks, and cohort replication in language models.} Top: per-name-pair paired $H$ values at primary dose $\lambda_0 = 0.1$ ($10^{-3}$ logit) and selected-head development strengths against the $128$-group unmatched random bank. Bottom: replication of primary gating estimates and antagonistic-minus-matched contrasts across the $32$-pair initial and $256$-pair confirmation cohorts.}
  \label{fig:s6_llm_pairs}
\end{figure}

\begin{figure}[p]
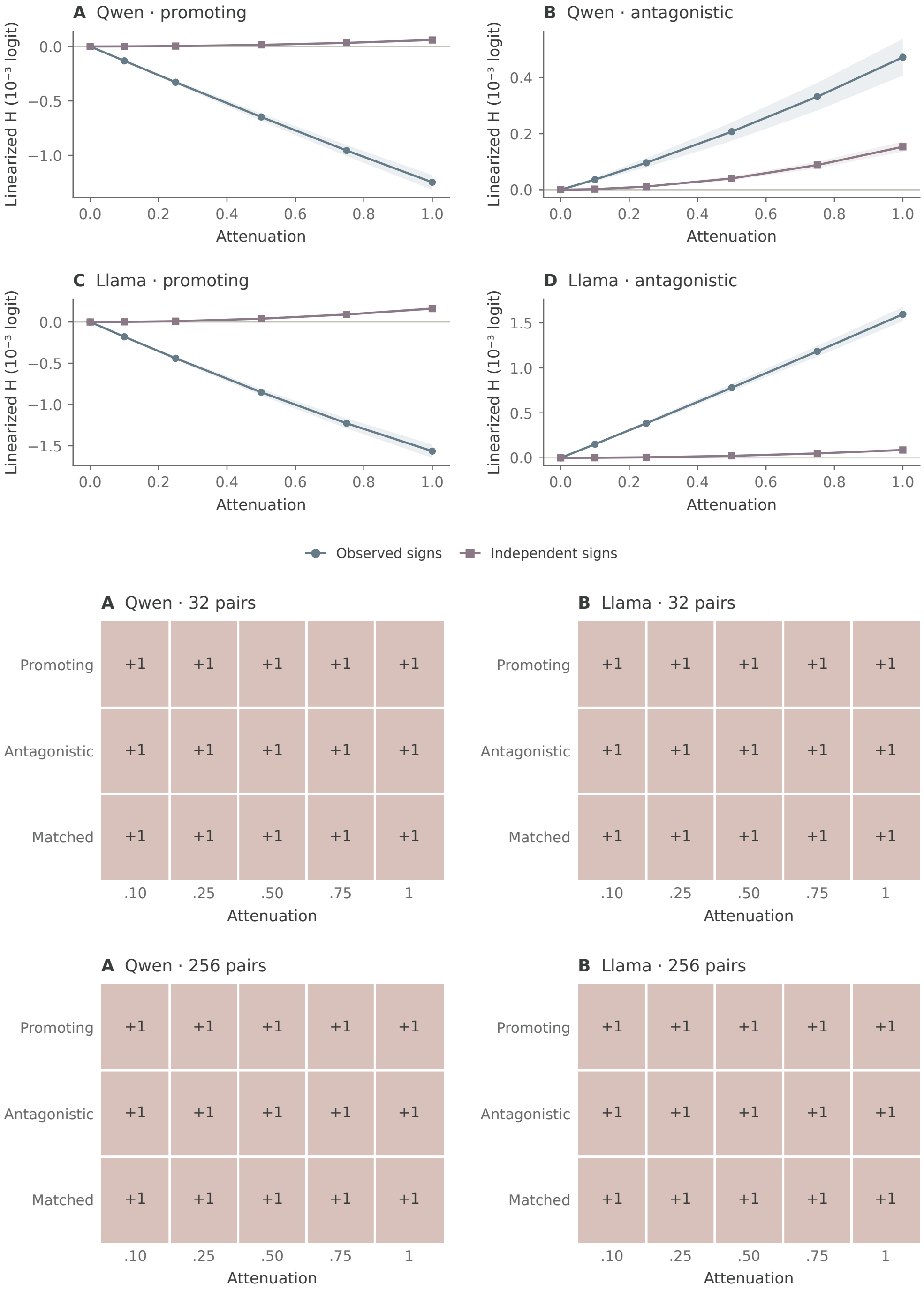

  \centering
  \includegraphics[page=7,width=0.80\textwidth]{S6_gating_and_groups.pdf}\\[0.35em]
  \includegraphics[page=5,width=0.80\textwidth]{S6_initial_controls.pdf}\\[0.35em]
  \includegraphics[page=6,width=0.80\textwidth]{S6_initial_controls.pdf}
  \caption{\textbf{Algebraic randomized-sign controls and hidden-versus-lexical sign concordance.} Top: algebraic dose responses comparing observed atomic signs against independently randomized signs ($256$ pairs). Middle and bottom: sign concordance between hidden-edit and lexical-replacement mean $H$ ($+1$ concordant, $-1$ opposite, $0$ zero mean) across the $32$-pair initial (middle) and $256$-pair confirmation (bottom) cohorts.}
  \label{fig:s6_alg_conc}
\end{figure}

\begin{figure}[p]
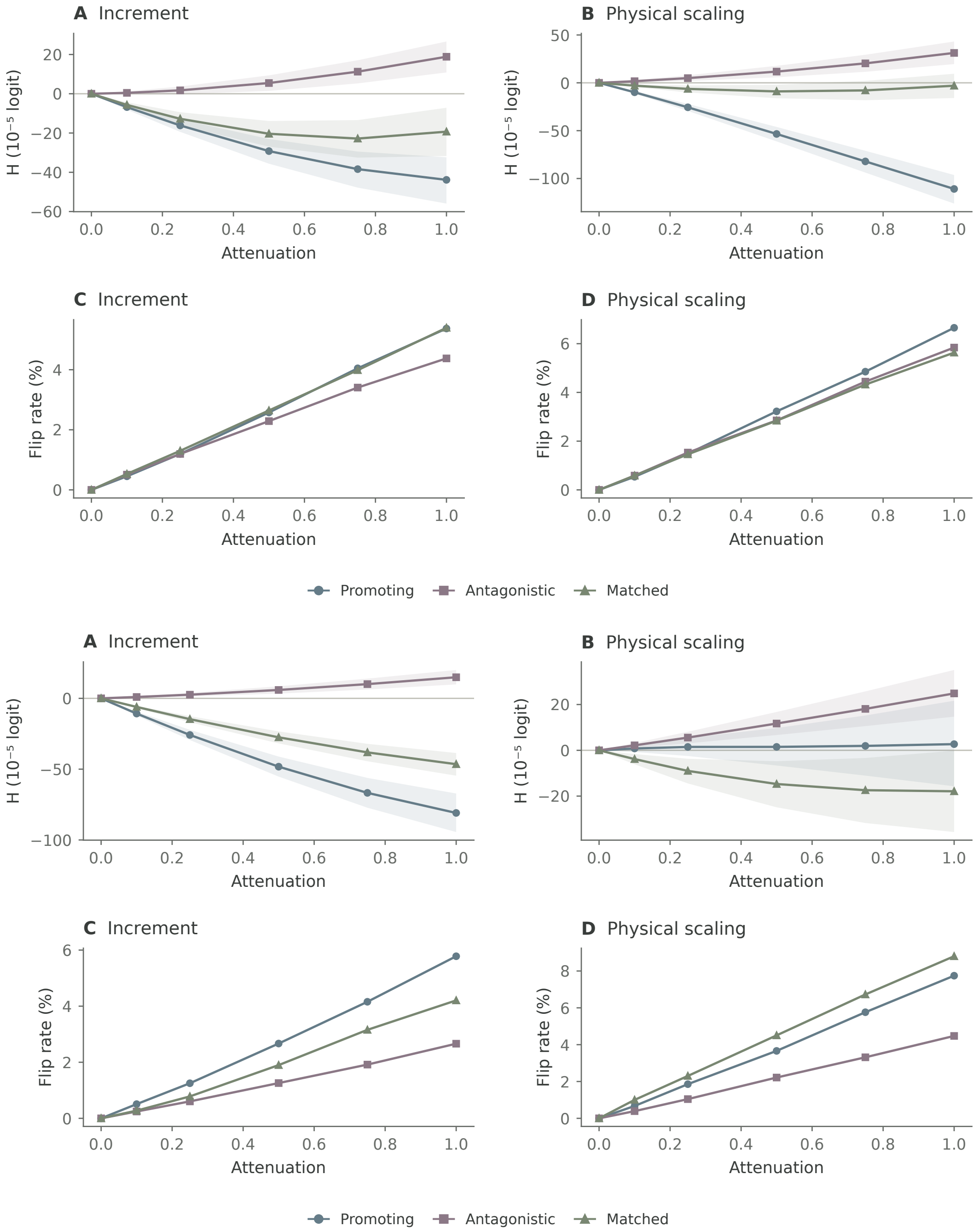

  \centering
  \includegraphics[page=8,width=0.86\textwidth]{S6_gating_and_groups.pdf}\\[0.35em]
  \includegraphics[page=9,width=0.86\textwidth]{S6_gating_and_groups.pdf}
  \caption{\textbf{Original ViT-B/16 contiguous-index gating on $1{,}024$ CUB confirmation images.} Top: original $32$-image-selected contiguous Block~6 groups evaluated across doses on $1{,}024$ confirmation images (increment attenuation and physical scaling). Bottom: $512$-image fallback contiguous selector on the same $1{,}024$ images (where $0/24$ strength-eligible contiguous candidates had positive mean development slope).}
  \label{fig:s6_vit_cont}
\end{figure}

\begin{figure}[p]
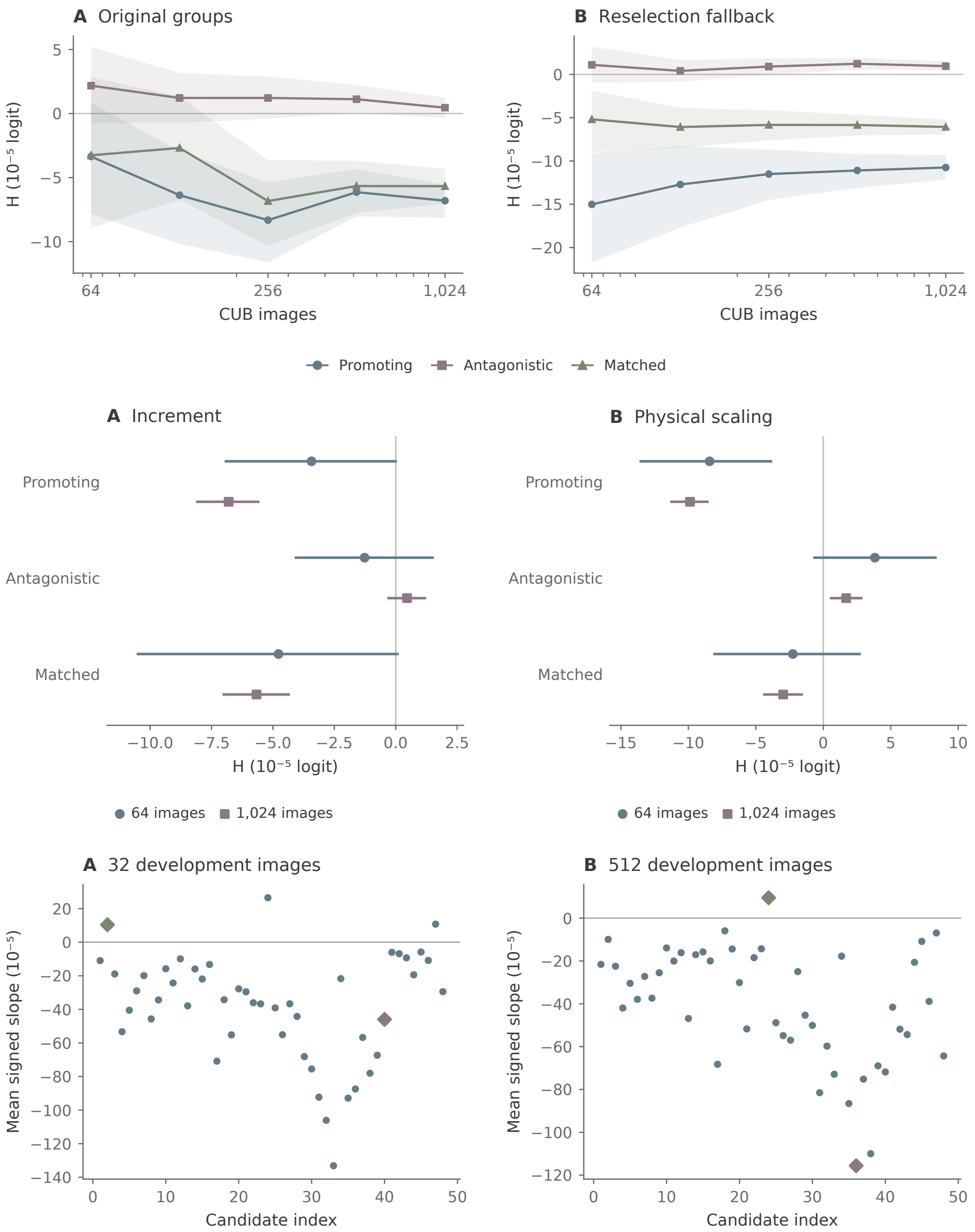

  \centering
  \includegraphics[page=10,width=0.86\textwidth]{S6_gating_and_groups.pdf}\\[0.35em]
  \includegraphics[page=11,width=0.86\textwidth]{S6_gating_and_groups.pdf}\\[0.35em]
  \includegraphics[page=12,width=0.86\textwidth]{S6_gating_and_groups.pdf}
  \caption{\textbf{ViT-B/16 contiguous-group stability, $64$-versus-$1{,}024$ comparison, and development slope landscape.} Top: nested sample-size curves for the original and fallback contiguous selectors. Middle: $64$-image initial versus $1{,}024$-image confirmation estimates for the original contiguous selector. Bottom: signed development slopes across all $48$ contiguous ViT-B/16 MLP candidates on $32$ versus $512$ development images (diamonds mark selected promoting/antagonistic candidates; on $512$ images, $0/24$ strength-eligible contiguous candidates are positive).}
  \label{fig:s6_drift_dev}
\end{figure}

\begin{figure}[p]
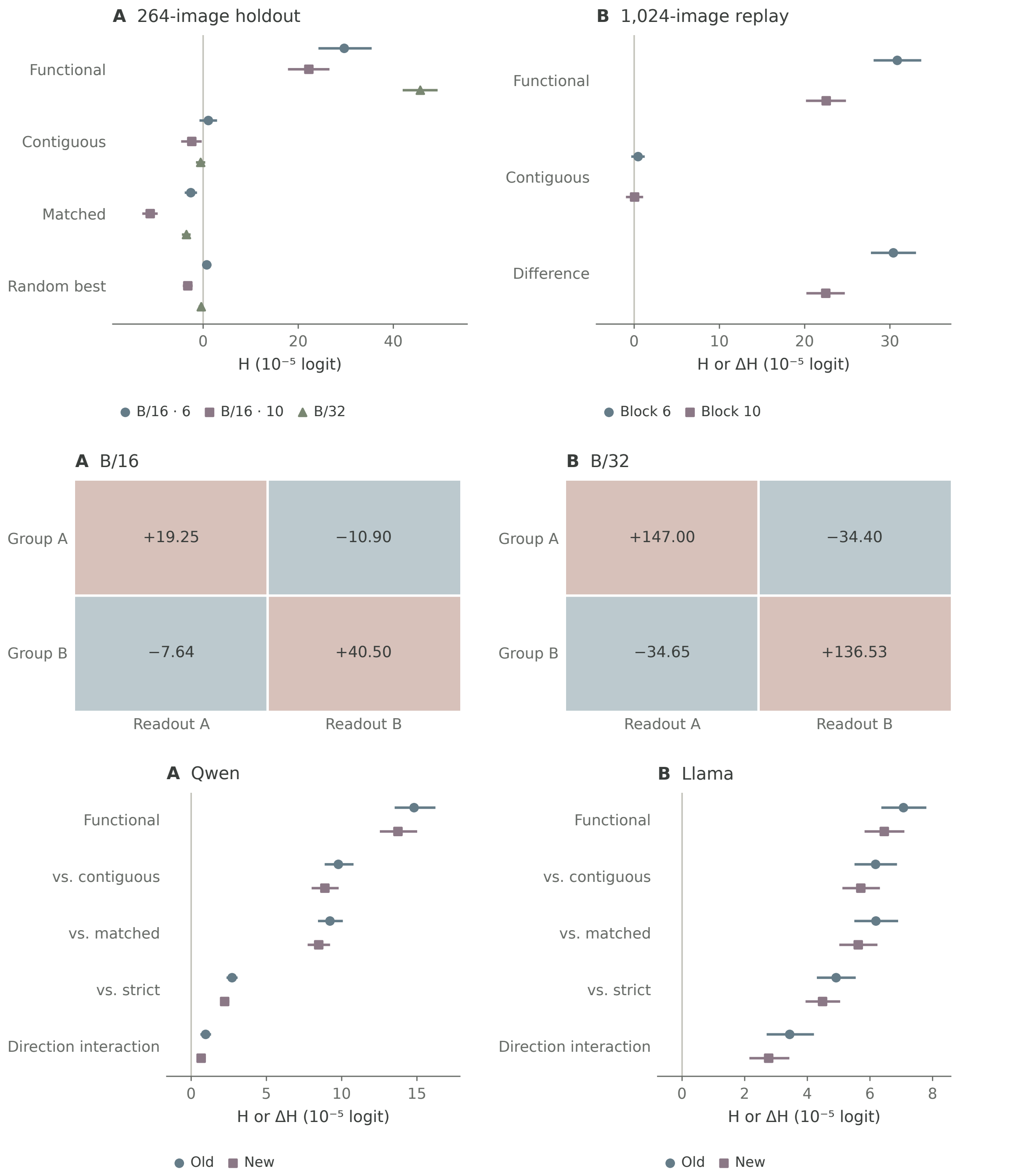

  \centering
  \includegraphics[page=13,width=0.88\textwidth]{S6_gating_and_groups.pdf}\\[0.35em]
  \includegraphics[page=14,width=0.88\textwidth]{S6_gating_and_groups.pdf}\\[0.35em]
  \includegraphics[page=15,width=0.88\textwidth]{S6_gating_and_groups.pdf}
  \caption{\textbf{Functional MLP grouping controls across vision and language.} Top: four grouping controls on the $264$-image reserve (ViT-B/16 Blocks~6 \& 10 and CUB-finetuned ViT-B/32) and $1{,}024$-image original-weight replay. Middle: same-forward-pass cross-readout matrices in original ImageNet ViT-B/16 (interaction $+78.30$) and CUB-finetuned ViT-B/32 ($+352.59 \times 10^{-5}$ logit). Bottom: complete old and new $256$-pair MLP functional gating contrasts in Qwen2.5 and Llama-3.2.}
  \label{fig:s6_functional}
\end{figure}

\begin{figure}[p]
  \centering
  \includegraphics[page=1,width=0.85\textwidth]{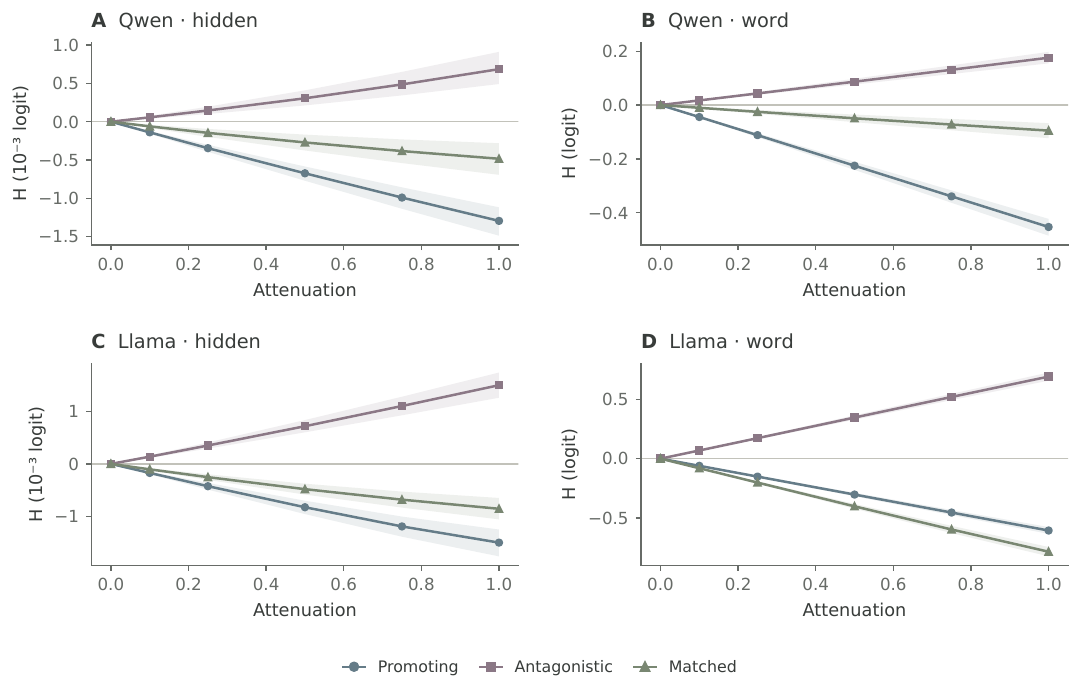}\\[0.35em]
  \includegraphics[page=2,width=0.85\textwidth]{S6_initial_controls.pdf}
  \caption{\textbf{Initial $32$-pair language cohort six-dose gating archive.} Increment attenuation (top) and physical output scaling (bottom) under hidden edits and real lexical replacements in the initial $32$-pair exploratory language cohort.}
  \label{fig:s6_init_dose}
\end{figure}

\begin{figure}[p]
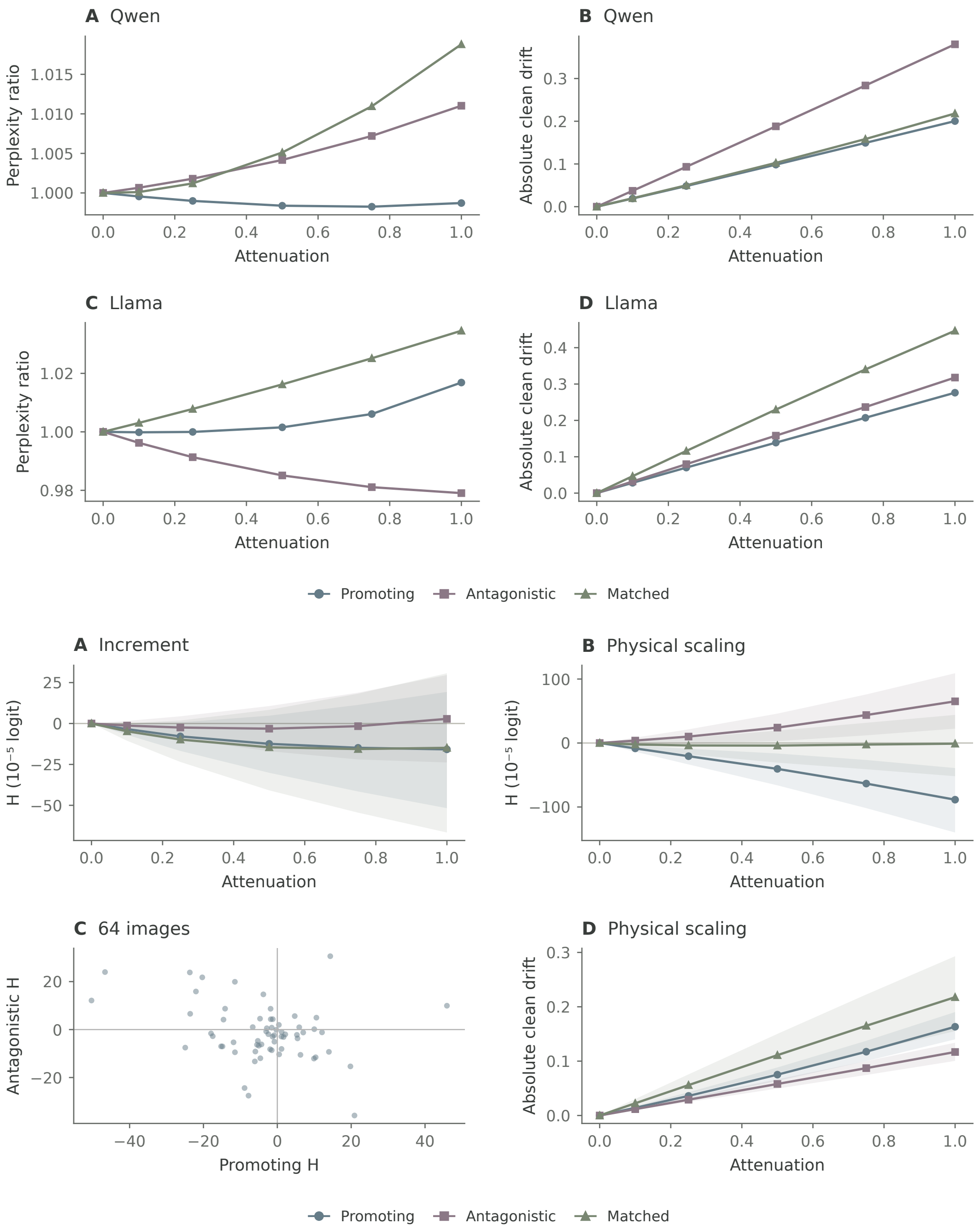

  \centering
  \includegraphics[page=3,width=0.86\textwidth]{S6_initial_controls.pdf}\\[0.35em]
  \includegraphics[page=4,width=0.86\textwidth]{S6_initial_controls.pdf}
  \caption{\textbf{Initial-cohort drift checks in language ($32$ pairs) and ViT-B/16 ($64$ images).} Top: physical-scaling clean perplexity and baseline margin drift in the initial $32$-pair language cohort. Bottom: six-dose curves, per-image primary effects ($10^{-5}$ logit), and physical baseline drift in the initial $64$-image CUB ViT-B/16 cohort.}
  \label{fig:s6_init_drift}
\end{figure}

This section archives all full gating tables (\cref{tab:s6_mlp_llm,tab:s6_vit}) and all twenty-one gating and functional-grouping graphic plates (\cref{fig:s6_llm_dose,fig:s6_llm_drift,fig:s6_llm_pairs,fig:s6_alg_conc,fig:s6_vit_cont,fig:s6_drift_dev,fig:s6_functional,fig:s6_init_dose,fig:s6_init_drift}): six-dose curves ($\lambda \in \{0, 0.1, 0.25, 0.5, 0.75, 1.0\}$) for both increment attenuation and physical output scaling under hidden edits and real lexical relation replacements, flip rates, baseline drift, and clean-prefix perplexity, per-name-pair scatter clouds and algebraic randomized-sign checks, ViT-B/16 contiguous-index development and holdout diagnostics across $32, 64, 512,$ and $1{,}024$ images, functional MLP grouping controls including CUB-finetuned ViT-B/32, and initial $32$-pair / $64$-image control sheets.

\begin{table}[htbp]
\centering
\caption{\textbf{Dual-LLM last-layer MLP functional grouping on $256$ replay and $256$ new zero-overlap name pairs.} Gating $H$ is in $10^{-5}$ logit ($10\%$ edit-increment attenuation at $+0.1$ edit dose); $\Delta R$ and coarse-minus-fine contrast $B$ are dimensionless. Tight $\pm 1\%$ strength matching ($99.07\%\text{--}100.98\%$) was run as a post-hoc sensitivity check in Qwen2.5 and locked prior to holdout in Llama-3.2.}
\label{tab:s6_mlp_llm}
\scriptsize
\setlength{\tabcolsep}{3.5pt}
\resizebox{\textwidth}{!}{%
\begin{tabular}{@{}l r r r r@{}}
\toprule
\textbf{Metric} & \textbf{Qwen Old ($256$)} & \textbf{Qwen New ($256$)} & \textbf{Llama Old ($256$)} & \textbf{Llama New ($256$)} \\
\midrule
Functional group $H$ ($10^{-5}$ logit) & $+14.81\,[13.52, 16.23]$ & $+13.74\,[12.54, 15.02]$ & $+7.07\,[6.36, 7.80]$ & $+6.46\,[5.83, 7.10]$ \\
Minus best contiguous block & $+9.78\,[8.87, 10.79]$ & $+8.88\,[8.00, 9.80]$ & $+6.18\,[5.51, 6.86]$ & $+5.71\,[5.12, 6.32]$ \\
Minus $64$-bin stratified matched & $+9.22\,[8.43, 10.08]$ & $+8.47\,[7.74, 9.23]$ & $+6.19\,[5.50, 6.90]$ & $+5.63\,[5.02, 6.24]$ \\
Minus tight $\pm 1\%$ matched & $+2.72\,[2.35, 3.08]$ & $+2.22\,[1.93, 2.54]$ & $+4.92\,[4.30, 5.55]$ & $+4.49\,[3.94, 5.05]$ \\
Selection $\times$ edit interaction & $+0.96\,[0.61, 1.32]$ & $+0.66\,[0.38, 0.97]$ & $+3.44\,[2.70, 4.21]$ & $+2.77\,[2.15, 3.43]$ \\
\midrule
Coarse four-path $\Delta R$ & $-0.0192\,[-0.0545, +0.0162]$ & $-0.0315\,[-0.0592, -0.0025]$ & $-0.0700\,[-0.1041, -0.0360]$ & $-0.0479\,[-0.0822, -0.0123]$ \\
Fine functional-split $\Delta R$ & $-0.0943\,[-0.1231, -0.0652]$ & $-0.1068\,[-0.1339, -0.0794]$ & $-0.1824\,[-0.2082, -0.1568]$ & $-0.1706\,[-0.1943, -0.1471]$ \\
Coarse$-$fine contrast $B$ & $+0.0751\,[0.0545, 0.0965]$ & $+0.0753\,[0.0564, 0.0948]$ & $+0.1124\,[0.0790, 0.1467]$ & $+0.1227\,[0.0876, 0.1595]$ \\
$B$ minus tight $\pm 1\%$ matched & $+0.0702\,[0.0548, 0.0864]$ & $+0.0699\,[0.0563, 0.0851]$ & $+0.1364\,[0.1061, 0.1680]$ & $+0.1372\,[0.1061, 0.1702]$ \\
Intra-$M$ selected $\times$ remaining & $+0.0844\,[0.0672, 0.1027]$ & $+0.0836\,[0.0668, 0.1018]$ & $+0.0983\,[0.0740, 0.1238]$ & $+0.1039\,[0.0786, 0.1304]$ \\
\bottomrule
\end{tabular}}
\end{table}

\begin{table}[htbp]
\centering
\caption{\textbf{Original ImageNet ViT-B/16 functional MLP grouping ($384/3{,}072$ units, $10\%$ attenuation, $10^{-5}$ logit) across $264$ reserve images and $1{,}024$-image original-weight replay.} CUB-finetuned ViT-B/32 on the same $264$ images yields functional $+45.67$ versus best contiguous $-0.51$ (paired $+46.19\,[42.61, 49.68]$) and cross-readout interaction $+352.59\,[319.71, 386.80]$.}
\label{tab:s6_vit}
\scriptsize
\setlength{\tabcolsep}{4pt}
\resizebox{\textwidth}{!}{%
\begin{tabular}{@{}l r r r r@{}}
\toprule
\textbf{Group or paired contrast} & \textbf{Block~10 ($264$ holdout)} & \textbf{Block~6 ($264$ holdout)} & \textbf{Block~6 ($1{,}024$ replay)} & \textbf{Block~10 ($1{,}024$ replay)} \\
\midrule
Functional group & $+22.23\,[17.84, 26.57]$ & $+29.68\,[24.25, 35.44]$ & $+30.86\,[28.09, 33.68]$ & $+22.53\,[20.15, 24.84]$ \\
Best contiguous block & $-2.39\,[-4.63, -0.30]$ & $+1.10\,[-0.77, +2.93]$ & $+0.46\,[-0.34, +1.24]$ & $+0.05\,[-0.97, +1.05]$ \\
Strength-matched random mean & $-11.13\,[-12.79, -9.56]$ & $-2.59\,[-3.87, -1.28]$ & --- & --- \\
Best in random partitions & $-3.22\,[-4.27, -2.20]$ & $+0.77\,[-0.18, +1.74]$ & --- & --- \\
\textbf{Functional $-$ best contiguous} & $\mathbf{+24.62\,[20.21, 29.04]}$ & $\mathbf{+28.57\,[23.58, 33.70]}$ & $\mathbf{+30.41\,[27.77, 33.06]}$ & $\mathbf{+22.47\,[20.20, 24.71]}$ \\
\textbf{Functional $-$ strength-matched} & $\mathbf{+33.36\,[28.75, 38.01]}$ & $\mathbf{+32.27\,[27.34, 37.46]}$ & --- & --- \\
\bottomrule
\end{tabular}}
\end{table}

\FloatBarrier
\section{Extended Pruning Recovery, Surgery, Window Controls, and Whole-Network Archive}
\label{supp:s7}

\begin{figure}[htbp]
  \centering
  \includegraphics[page=2,width=0.88\textwidth]{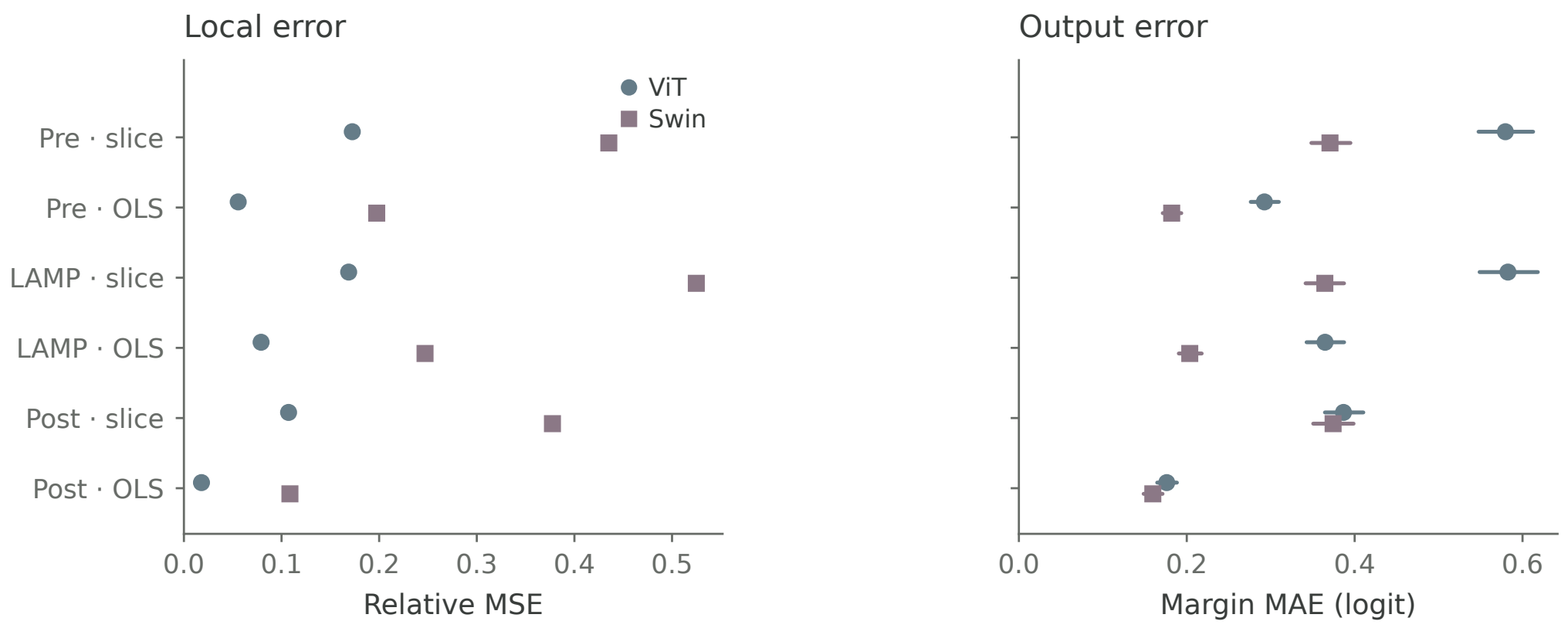}
  \caption{\textbf{Initial $1{,}000$-image six-arm summary at the middle Transformer MLP cuts.} Relative local MSE and class-margin MAE across all six arms (Pre-GELU, LAMP, Post-GELU under direct slicing vs.\ OLS) in the initial $1{,}000$-image study.}
  \label{fig:s7_init_arms}
\end{figure}

\begin{figure}[htbp]
  \centering
  \includegraphics[page=6,width=0.92\textwidth]{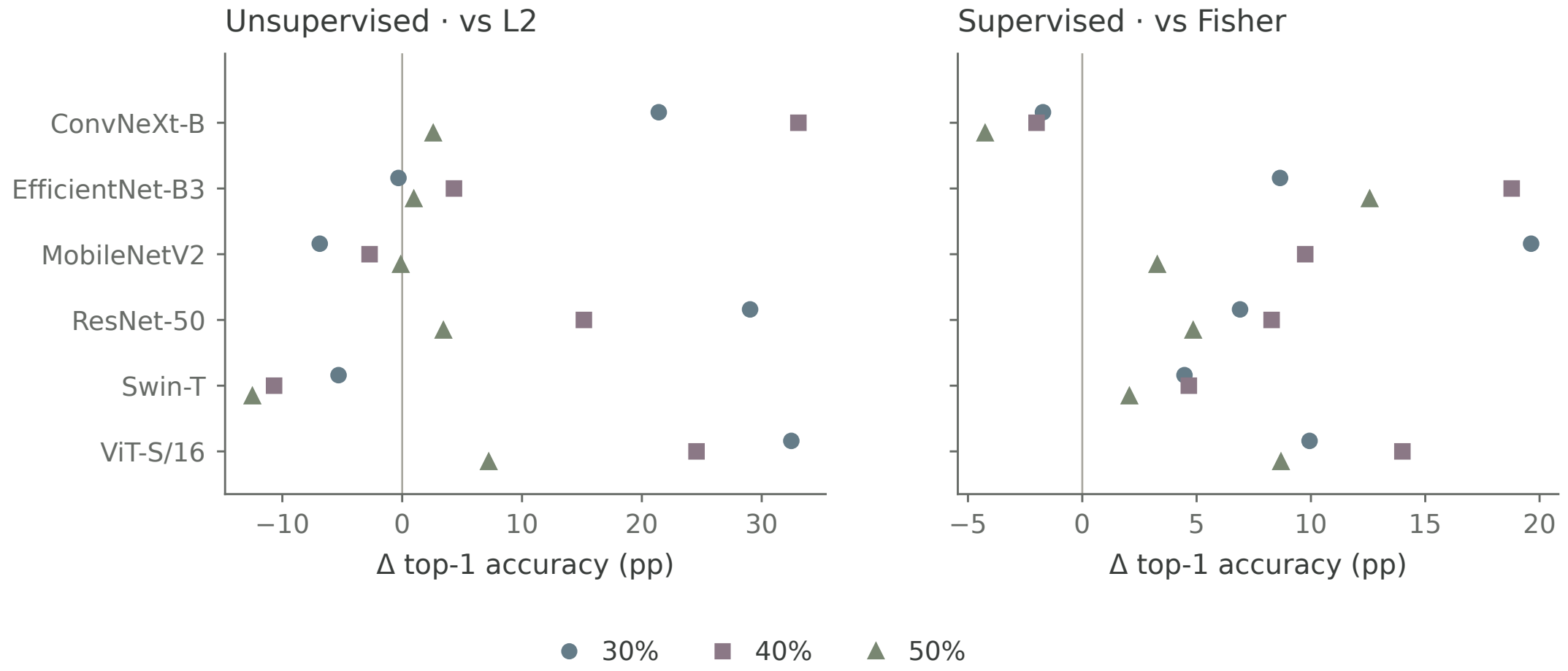}
  \caption{\textbf{Historical six-architecture whole-network accuracy gains across three removal budgets.} Visual companion to \cref{tab:s7b_whole}, showing unlabeled ($M_u$ vs.\ $L_2$) and task-conditioned ($M_s$ vs.\ Fisher) accuracy shifts across six architecture families ($3{,}200$ ImageNet images).}
  \label{fig:s7_whole_bars}
\end{figure}

\begin{figure}[p]
  \centering
  \includegraphics[page=1,width=0.83\textwidth]{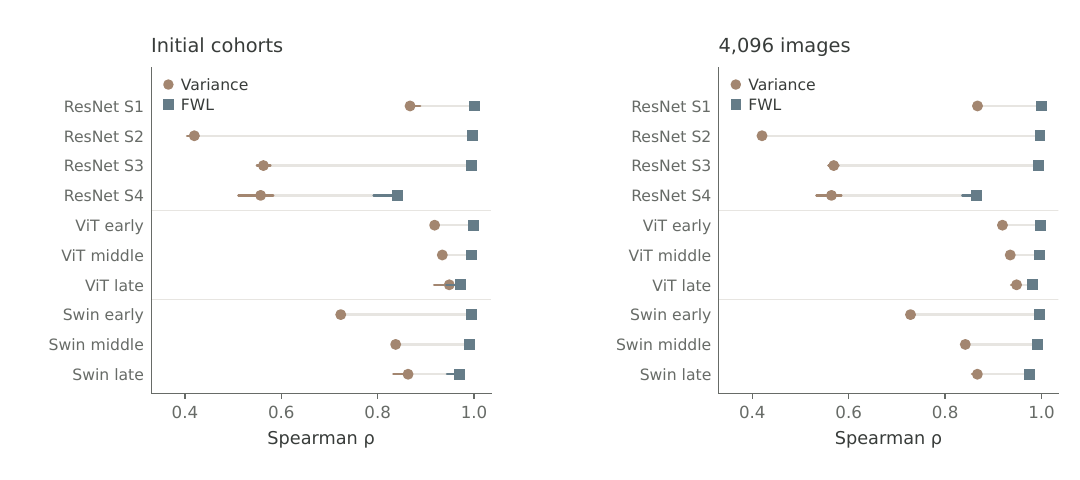}\\[0.35em]
  \includegraphics[page=5,width=0.83\textwidth]{S7_rankings_and_controls.pdf}\\[0.35em]
  \includegraphics[page=1,width=0.83\textwidth]{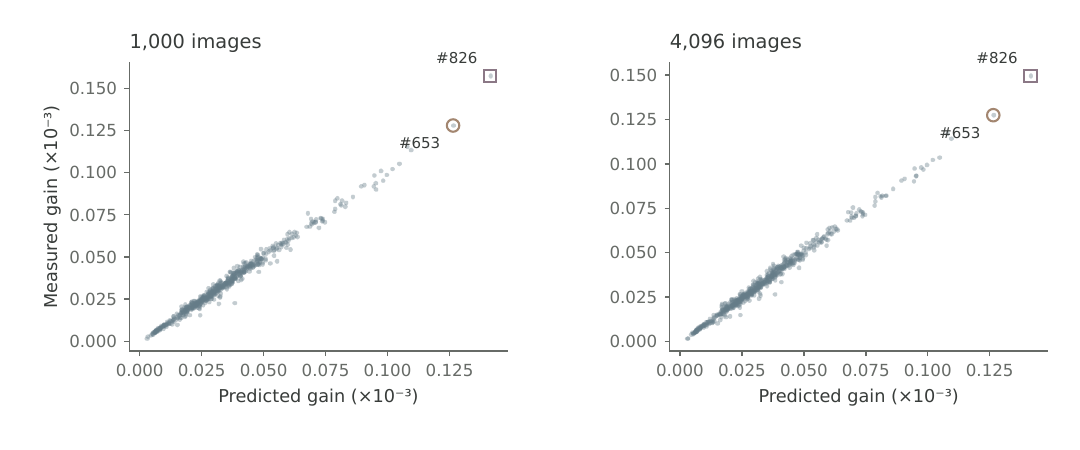}
  \caption{\textbf{Initial-versus-final pruning rankings (top), Swin window stratification (middle), and Swin candidate-gain clouds (bottom).} Top: initial ($2{,}048/1{,}000$ images) versus final ($4{,}096$ images) candidate rankings. Middle: Swin-T middle-block OLS MSE comparing global shared OLS ($0.016910$), nine real shifted windows ($0.017068$), and equal-capacity random stratification ($0.017425$). Bottom: per-channel MSE reductions ($10^{-3}$ per output coordinate) in the initial ($1{,}000$ images) versus final ($4{,}096$ images) Swin-T middle cut, highlighting variance-selected ($\#653$) and FWL-selected ($\#826$) channels.}
  \label{fig:s7_controls}
\end{figure}

\begin{figure}[p]
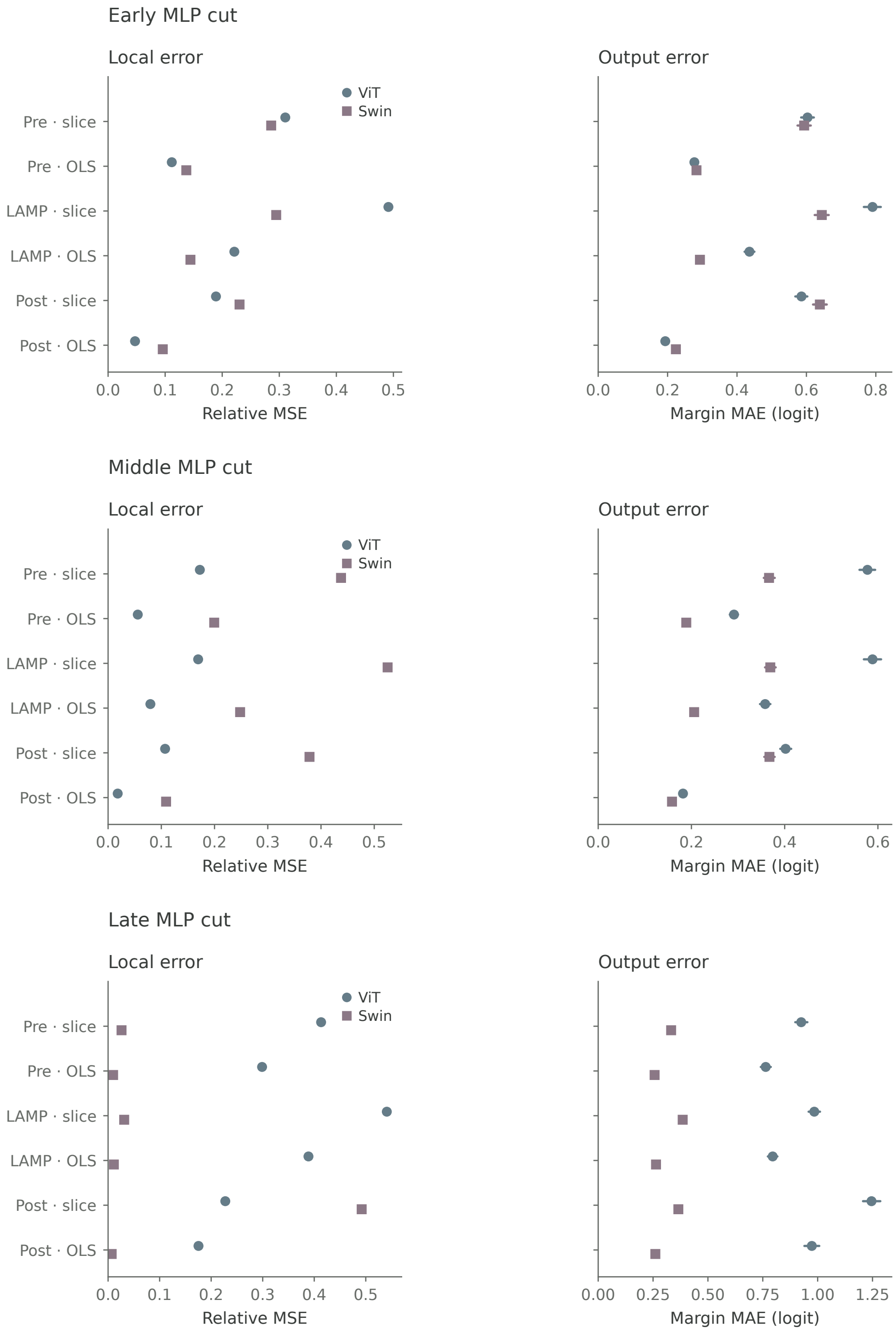

  \centering
  \includegraphics[page=2,width=0.82\textwidth]{S7_rankings_and_controls.pdf}\\[0.35em]
  \includegraphics[page=3,width=0.82\textwidth]{S7_rankings_and_controls.pdf}\\[0.35em]
  \includegraphics[page=4,width=0.82\textwidth]{S7_rankings_and_controls.pdf}
  \caption{\textbf{Six-arm relative local MSE and class-margin MAE across early (top), middle (middle), and late (bottom) Transformer MLP cuts ($4{,}096$ ImageNet images).} Compares Pre-GELU, LAMP, and Post-GELU selectors paired with direct slicing versus OLS weight folding ($95\%$ paired image bootstrap CIs).}
  \label{fig:s7_six_arms}
\end{figure}

\begin{figure}[p]
  \centering
  \includegraphics[page=1,width=0.84\textwidth]{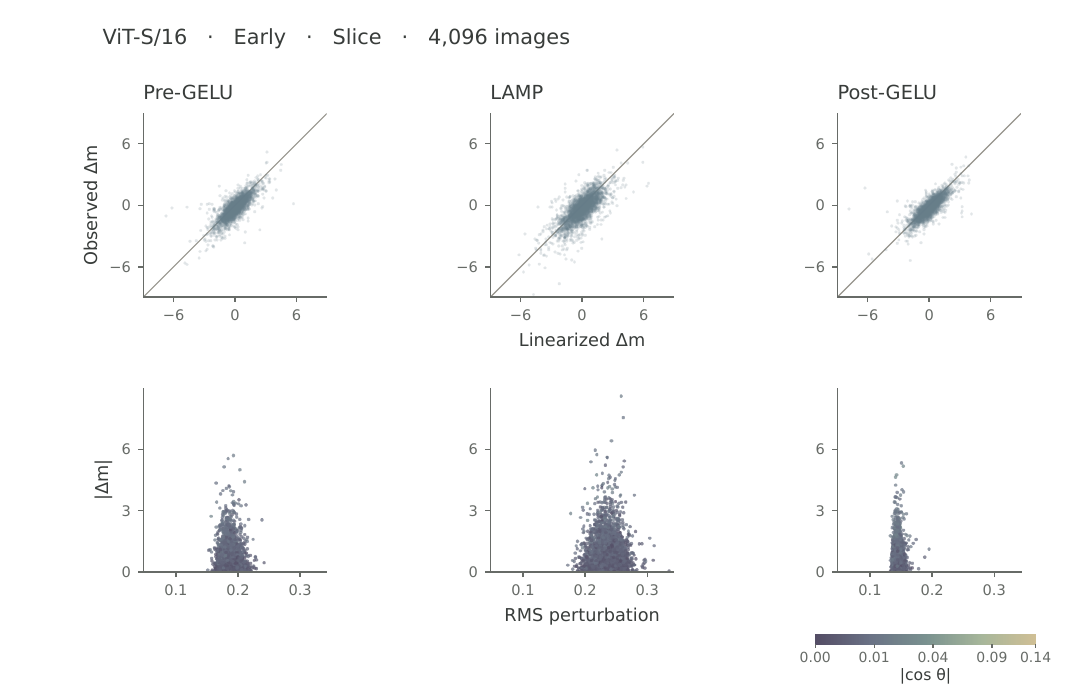}\\[0.35em]
  \includegraphics[page=2,width=0.84\textwidth]{S7_early_surgery.pdf}
  \caption{\textbf{Per-image physical surgery at the ViT-S/16 early MLP cut ($4{,}096$ ImageNet images).} Direct slicing (top) versus OLS folding (bottom) across Pre-GELU, LAMP, and Post-GELU selectors. Color encodes $|\cos\theta|$ between local perturbation and clean margin gradient on a unified square-root scale over $[0, 0.14]$ shared across all surgery sheets.}
  \label{fig:s7_early_vit}
\end{figure}

\begin{figure}[p]
  \centering
  \includegraphics[page=1,width=0.84\textwidth]{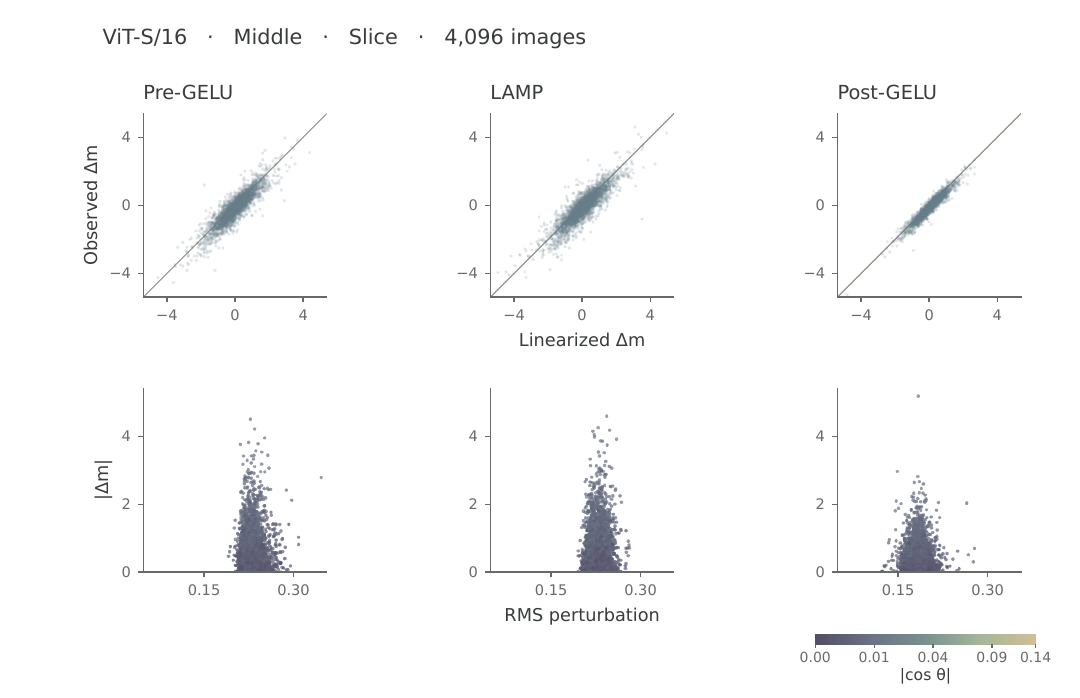}\\[0.35em]
  \includegraphics[page=2,width=0.84\textwidth]{S7_middle_surgery.pdf}
  \caption{\textbf{Per-image physical surgery at the ViT-S/16 middle MLP cut ($4{,}096$ ImageNet images).} Direct slicing (top) versus OLS folding (bottom) across Pre-GELU, LAMP, and Post-GELU selectors on the identical square-root $|\cos\theta|$ scale.}
  \label{fig:s7_middle_vit}
\end{figure}

\begin{figure}[p]
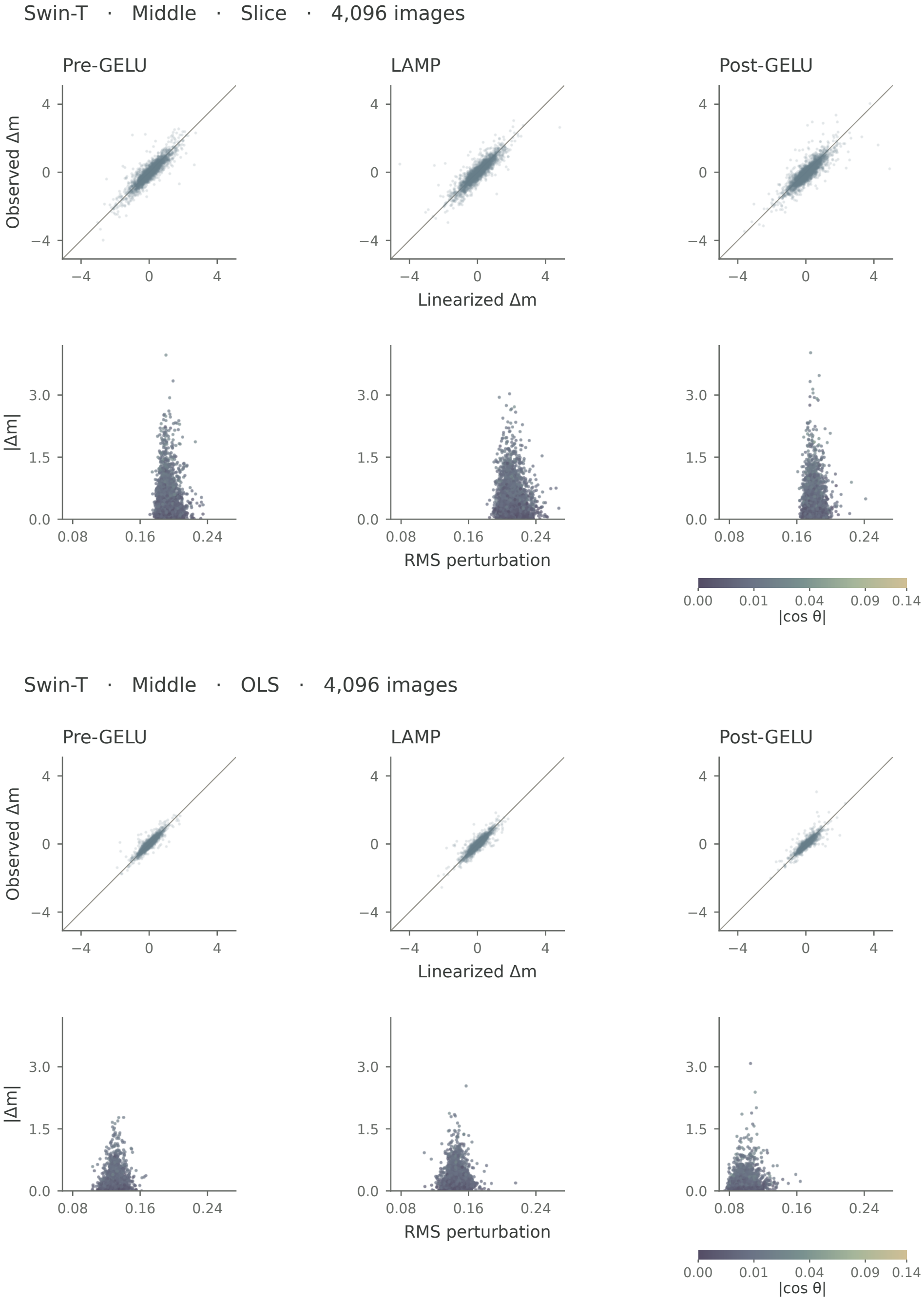

  \centering
  \includegraphics[page=3,width=0.84\textwidth]{S7_middle_surgery.pdf}\\[0.35em]
  \includegraphics[page=4,width=0.84\textwidth]{S7_middle_surgery.pdf}
  \caption{\textbf{Per-image physical surgery at the Swin-T middle MLP cut ($4{,}096$ ImageNet images).} Direct slicing (top) versus OLS folding (bottom) on the identical square-root $|\cos\theta|$ scale, visually illustrating how post-GELU direct slicing compresses RMS perturbation while increasing $|\cos\theta|$, leaving $|\Delta m|$ nearly unchanged until paired with OLS folding.}
  \label{fig:s7_middle_swin}
\end{figure}

\begin{figure}[p]
  \centering
  \includegraphics[page=1,width=0.84\textwidth]{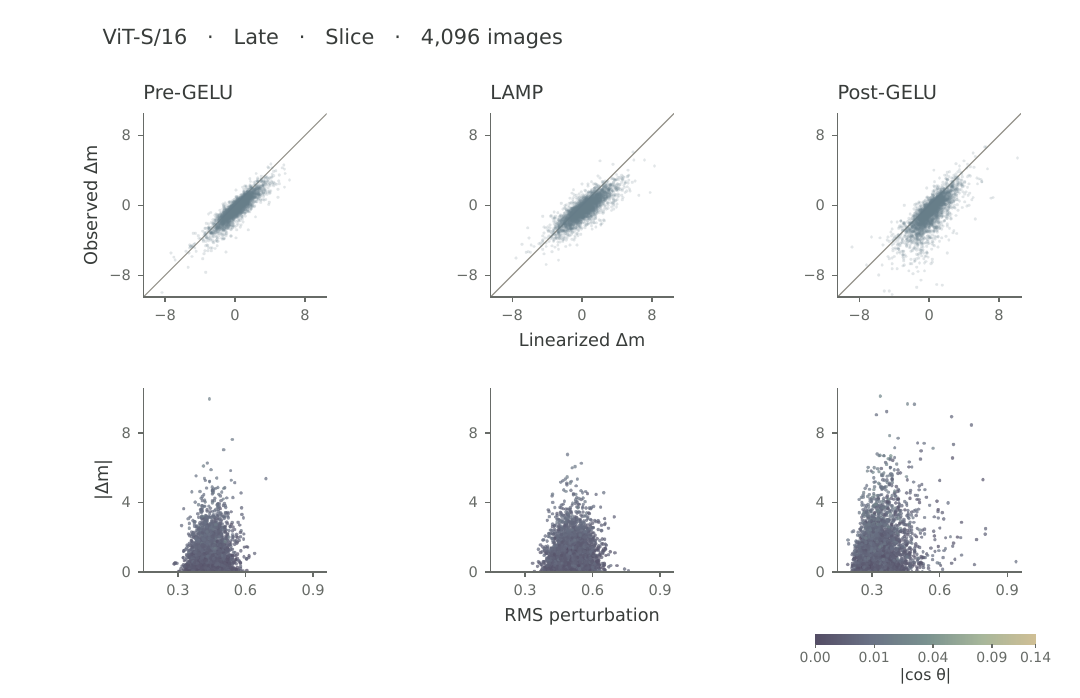}\\[0.35em]
  \includegraphics[page=2,width=0.84\textwidth]{S7_late_surgery.pdf}
  \caption{\textbf{Per-image physical surgery at the ViT-S/16 late MLP cut ($4{,}096$ ImageNet images).} Direct slicing (top) versus OLS folding (bottom) on the identical square-root $|\cos\theta|$ scale.}
  \label{fig:s7_late_vit}
\end{figure}

\begin{figure}[p]
  \centering
  \includegraphics[page=1,width=0.84\textwidth]{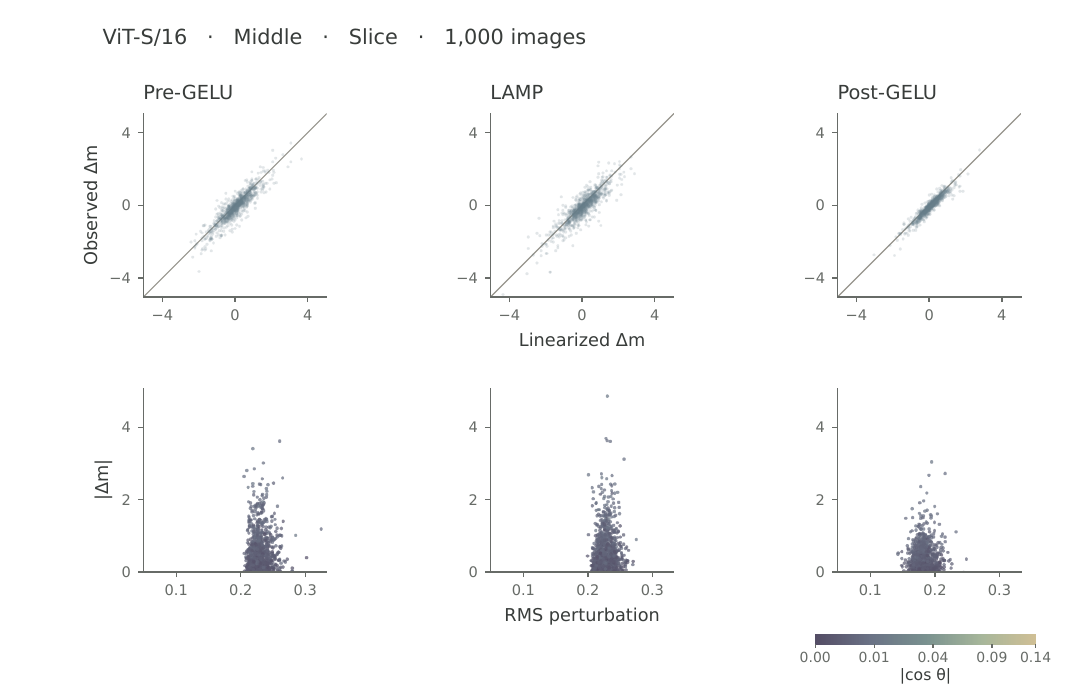}\\[0.35em]
  \includegraphics[page=2,width=0.84\textwidth]{S7_historical_surgery.pdf}
  \caption{\textbf{Historical $1{,}000$-image physical surgery at the ViT-S/16 middle MLP cut.} Direct slicing (top) versus OLS folding (bottom) on the identical square-root $|\cos\theta|$ scale, matching the $4{,}096$-image confirmation cloud in \cref{fig:s7_middle_vit}.}
  \label{fig:s7_hist_vit}
\end{figure}

\begin{figure}[p]
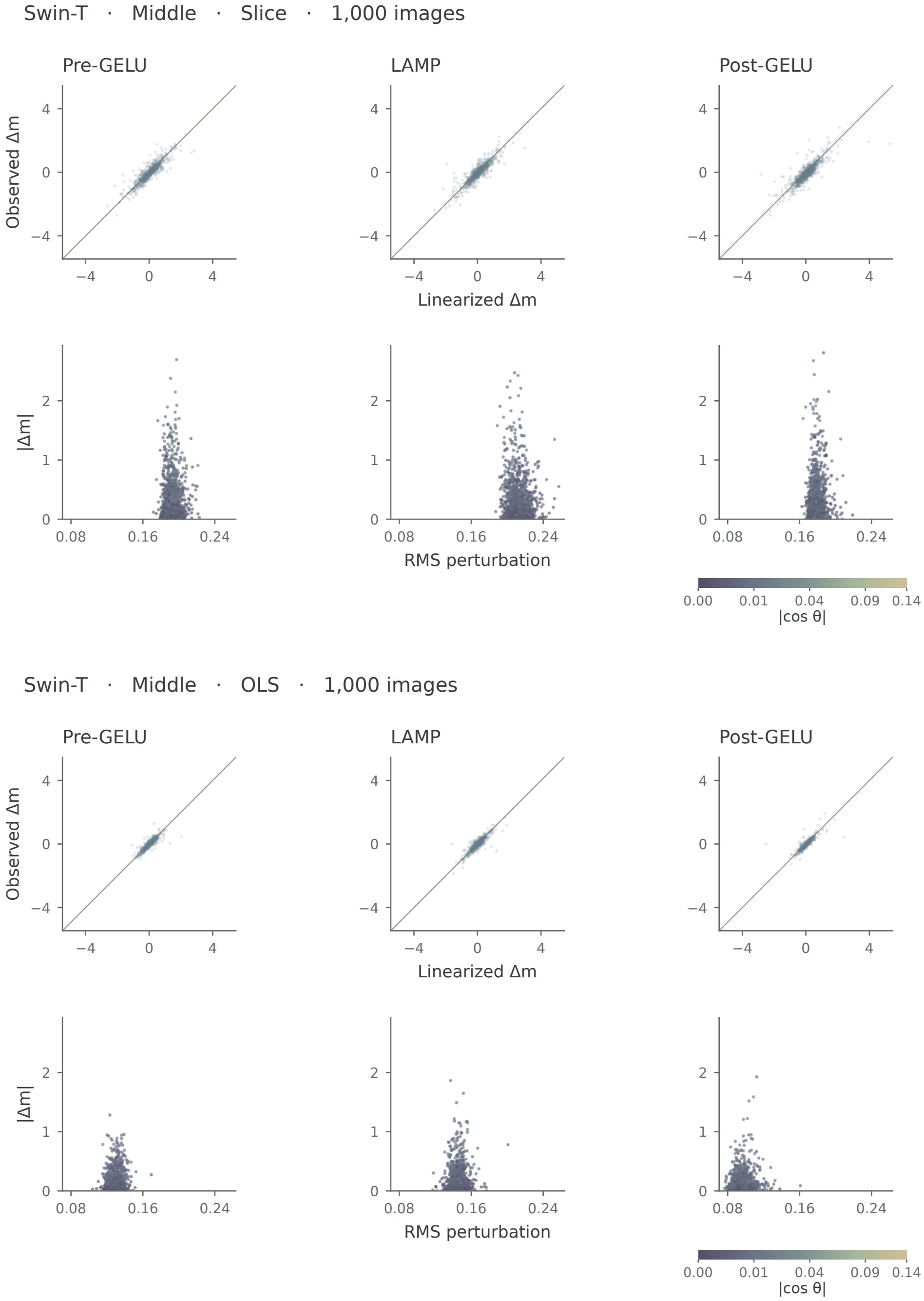

  \centering
  \includegraphics[page=3,width=0.84\textwidth]{S7_historical_surgery.pdf}\\[0.35em]
  \includegraphics[page=4,width=0.84\textwidth]{S7_historical_surgery.pdf}
  \caption{\textbf{Historical $1{,}000$-image physical surgery at the Swin-T middle MLP cut.} Direct slicing (top) versus OLS folding (bottom) on the identical square-root $|\cos\theta|$ scale, matching the $4{,}096$-image confirmation cloud in \cref{fig:s7_middle_swin}.}
  \label{fig:s7_hist_swin}
\end{figure}

\Cref{tab:s7_ten_cuts} records exact held-out Spearman correlations across all $10$ pruning cuts ($4{,}096$ ImageNet images), \cref{tab:s7_pearson} lists observed-versus-linearized margin Pearson correlations across all six arms and depths, and \cref{tab:s7b_whole} reproduces the verified historical six-family whole-network accuracy archive ($3{,}200$ images from \citet{shu2026pruning}, with $257$ images overlapping the local $4{,}096$-image cohort). \Cref{fig:s7_init_arms,fig:s7_whole_bars,fig:s7_controls,fig:s7_six_arms,fig:s7_early_vit,fig:s7_early_swin,fig:s7_middle_vit,fig:s7_middle_swin,fig:s7_late_vit,fig:s7_late_swin,fig:s7_hist_vit,fig:s7_hist_swin} provide the complete set of $24$ non-duplicate pruning plates: initial-vs-final rankings, Swin-T shifted-window stratification (global OLS $0.016910$ $<$ real shifted windows $0.017068$ $<$ random stratification $0.017425$), Swin candidate-gain clouds, whole-network accuracy gains, six-arm MSE/MAE curves across early, middle, and late cuts, and every per-image physical-surgery scatter cloud across early, middle, late ($4{,}096$ images), and historical middle ($1{,}000$ images) cuts for both ViT-S/16 and Swin-T.

\begin{table}[htbp]
\centering
\caption{\textbf{All ten local pruning cuts on $4{,}096$ held-out ImageNet images.} Family-adjusted paired bootstrap CIs are $97.5\%$ for the two primary middle Transformer cuts and $98.75\%$ for the remaining cuts.}
\label{tab:s7_ten_cuts}
\scriptsize
\begin{tabular}{@{}l r r r r l@{}}
\toprule
\textbf{Model \& cut} & \textbf{Images} & \textbf{Variance $\rho$} & \textbf{FWL $\rho$} & \textbf{Gain ($\Delta\rho$)} & \textbf{Family CI for $\Delta\rho$} \\
\midrule
ResNet-50 \texttt{layer1.0.conv2} & 4,096 & 0.8671 & 1.0000 & $+0.1329$ & $[0.1329, 0.1329]$ \\
ResNet-50 \texttt{layer2.0.conv2} & 4,096 & 0.4201 & 0.9967 & $+0.5766$ & $[0.5688, 0.5824]$ \\
ResNet-50 \texttt{layer3.0.conv2} & 4,096 & 0.5689 & 0.9936 & $+0.4248$ & $[0.4129, 0.4362]$ \\
ResNet-50 \texttt{layer4.0.conv2} & 4,096 & 0.5641 & 0.8650 & $+0.3009$ & $[0.2678, 0.3178]$ \\
\midrule
ViT-S/16 early MLP & 4,096 & 0.9189 & 0.9985 & $+0.0796$ & $[0.0787, 0.0807]$ \\
ViT-S/16 middle MLP & 4,096 & 0.9351 & 0.9959 & $+0.0608$ & $[0.0594, 0.0623]$ \\
ViT-S/16 late MLP & 4,096 & 0.9482 & 0.9806 & $+0.0324$ & $[0.0274, 0.0373]$ \\
\midrule
Swin-T early MLP & 4,096 & 0.7282 & 0.9954 & $+0.2672$ & $[0.2629, 0.2709]$ \\
Swin-T middle MLP & 4,096 & 0.8419 & 0.9918 & $+0.1499$ & $[0.1475, 0.1526]$ \\
Swin-T late MLP & 4,096 & 0.8669 & 0.9747 & $+0.1078$ & $[0.1006, 0.1136]$ \\
\bottomrule
\end{tabular}
\end{table}

\begin{table}[htbp]
\centering
\caption{\textbf{Pearson correlations between observed and linearized class-margin changes across all physical-surgery scatter clouds (\cref{fig:s7_early_vit,fig:s7_early_swin,fig:s7_middle_vit,fig:s7_middle_swin,fig:s7_late_vit,fig:s7_late_swin,fig:s7_hist_vit,fig:s7_hist_swin}).}}
\label{tab:s7_pearson}
\scriptsize
\begin{tabular}{@{}r l l r r r r r r@{}}
\toprule
\textbf{Images} & \textbf{Cut} & \textbf{Model} & \textbf{Pre slice} & \textbf{LAMP slice} & \textbf{Post slice} & \textbf{Pre OLS} & \textbf{LAMP OLS} & \textbf{Post OLS} \\
\midrule
4,096 & Early & ViT-S/16 & 0.781 & 0.733 & 0.800 & 0.848 & 0.833 & 0.918 \\
4,096 & Early & Swin-T & 0.836 & 0.762 & 0.846 & 0.897 & 0.894 & 0.887 \\
4,096 & Middle & ViT-S/16 & 0.897 & 0.881 & 0.959 & 0.953 & 0.935 & 0.979 \\
4,096 & Middle & Swin-T & 0.898 & 0.881 & 0.858 & 0.916 & 0.901 & 0.890 \\
4,096 & Late & ViT-S/16 & 0.890 & 0.827 & 0.753 & 0.914 & 0.873 & 0.859 \\
4,096 & Late & Swin-T & 0.961 & 0.942 & 0.948 & 0.973 & 0.969 & 0.975 \\
\midrule
1,000 & Init.\ mid & ViT-S/16 & 0.904 & 0.875 & 0.951 & 0.951 & 0.933 & 0.967 \\
1,000 & Init.\ mid & Swin-T & 0.899 & 0.886 & 0.844 & 0.878 & 0.885 & 0.841 \\
\bottomrule
\end{tabular}
\end{table}

\begin{table}[htbp]
\centering
\caption{\textbf{Historical six-architecture whole-network structured-pruning accuracy archive across three removal budgets.} Re-verified from $540$ published records in \citet{shu2026pruning} ($3{,}200$ ImageNet validation images; $257$ overlap our $4{,}096$-image local cohort). Unlabeled ($M_u$) and task-conditioned ($M_s$) accuracies are compared against $L_2$, Fisher, and best competing baselines. Whole-network multi-layer Swin accuracy behavior remains an open question beyond single-cut local linearization.}
\label{tab:s7b_whole}
\scriptsize
\setlength{\tabcolsep}{4pt}
\resizebox{\textwidth}{!}{%
\begin{tabular}{@{}l r r r l r r r l r@{}}
\toprule
\textbf{Architecture} & \textbf{Budget} & \textbf{$M_u$ (\%)} & \textbf{vs.\ $L_2$} & \textbf{Best other} & \textbf{vs.\ best} & \textbf{$M_s$ (\%)} & \textbf{vs.\ Fisher} & \textbf{Best other} & \textbf{vs.\ best} \\
\midrule
ConvNeXt-B & 30\% & 71.69 & $+21.41$ & APoZ & $-0.81$ & 65.69 & $-1.72$ & TaylorGate & $-6.84$ \\
ConvNeXt-B & 40\% & 47.09 & $+33.06$ & APoZ & $-0.07$ & 49.69 & $-2.00$ & FisherGate & $-6.28$ \\
ConvNeXt-B & 50\% & 3.03 & $+2.59$ & APoZ & $-0.75$ & 15.22 & $-4.25$ & OBD & $-8.81$ \\
EfficientNet-B3 & 30\% & 39.66 & $-0.31$ & FPGM & $-1.34$ & 61.03 & $+8.65$ & OBD & $+0.25$ \\
EfficientNet-B3 & 40\% & 19.38 & $+4.32$ & APoZ & $+2.63$ & 39.97 & $+18.78$ & OBD & $+3.31$ \\
EfficientNet-B3 & 50\% & 5.19 & $+0.97$ & APoZ & $+0.25$ & 19.88 & $+12.57$ & OBD & $+6.07$ \\
MobileNetV2 & 30\% & 15.84 & $-6.88$ & $L_1$ & $-7.63$ & 42.63 & $+19.63$ & OBD & $+5.66$ \\
MobileNetV2 & 40\% & 4.72 & $-2.72$ & $L_2$/LAMP & $-2.72$ & 16.25 & $+9.75$ & OBD & $+4.72$ \\
MobileNetV2 & 50\% & 1.19 & $-0.12$ & $L_2$/LAMP & $-0.12$ & 4.69 & $+3.28$ & GradNorm & $+2.25$ \\
ResNet-50 & 30\% & 40.94 & $+29.03$ & APoZ & $+18.00$ & 50.84 & $+6.90$ & GradNorm & $+5.93$ \\
ResNet-50 & 40\% & 18.75 & $+15.16$ & APoZ & $+13.25$ & 27.87 & $+8.28$ & GradNorm & $+3.68$ \\
ResNet-50 & 50\% & 5.41 & $+3.44$ & $L_2$/LAMP & $+3.44$ & 9.88 & $+4.85$ & GradNorm & $+3.04$ \\
Swin-T & 30\% & 59.94 & $-5.31$ & $L_2$/LAMP & $-5.31$ & 60.03 & $+4.47$ & Wanda & $-4.94$ \\
Swin-T & 40\% & 44.53 & $-10.69$ & $L_2$/LAMP & $-10.69$ & 43.03 & $+4.66$ & Wanda & $-7.56$ \\
Swin-T & 50\% & 21.22 & $-12.50$ & $L_2$/LAMP & $-12.50$ & 22.03 & $+2.06$ & Wanda & $-3.34$ \\
ViT-S/16 & 30\% & 52.53 & $+32.47$ & APoZ & $+6.97$ & 54.03 & $+9.94$ & FisherGate & $+0.72$ \\
ViT-S/16 & 40\% & 31.28 & $+24.56$ & APoZ & $+9.19$ & 36.34 & $+14.00$ & FisherGate & $-6.16$ \\
ViT-S/16 & 50\% & 9.75 & $+7.22$ & Random & $+6.22$ & 15.75 & $+8.69$ & FisherGate & $-5.56$ \\
\bottomrule
\end{tabular}}
\end{table}

\begin{figure}[p]
  \centering
  \includegraphics[page=3,width=0.84\textwidth]{S7_early_surgery.pdf}\\[0.35em]
  \includegraphics[page=4,width=0.84\textwidth]{S7_early_surgery.pdf}
  \caption{\textbf{Per-image physical surgery at the Swin-T early MLP cut ($4{,}096$ ImageNet images).} Direct slicing (top) versus OLS folding (bottom) on the identical square-root $|\cos\theta|$ scale.}
  \label{fig:s7_early_swin}
\end{figure}

\begin{figure}[p]
  \centering
  \includegraphics[page=3,width=0.84\textwidth]{S7_late_surgery.pdf}\\[0.35em]
  \includegraphics[page=4,width=0.84\textwidth]{S7_late_surgery.pdf}
  \caption{\textbf{Per-image physical surgery at the Swin-T late MLP cut ($4{,}096$ ImageNet images).} Direct slicing (top) versus OLS folding (bottom) on the identical square-root $|\cos\theta|$ scale.}
  \label{fig:s7_late_swin}
\end{figure}

\FloatBarrier
\section{OOD Corruption Phenomenon and Exact \texorpdfstring{$R = 1 + bc$}{R = 1 + bc} Decomposition}
\label{supp:s8}

\begin{figure}[htbp]
  \centering
  \includegraphics[width=0.92\textwidth]{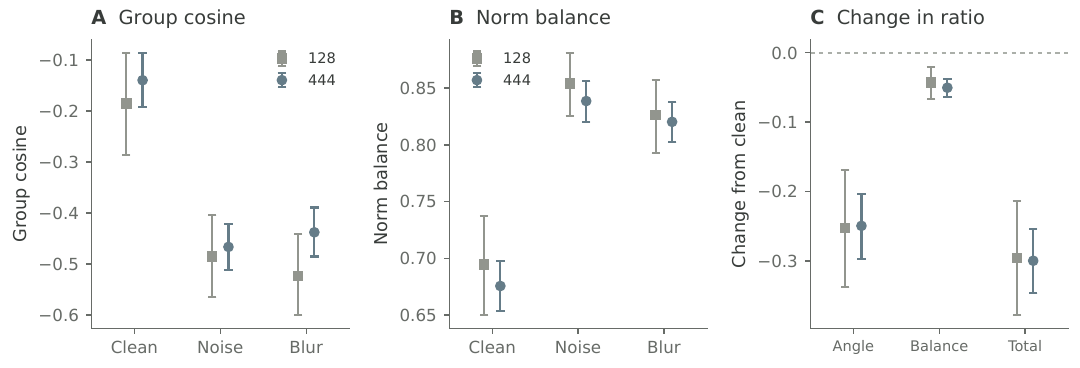}
  \caption{\textbf{Two-group corruption geometry and $R = 1 + bc$ decomposition across $128$- and $444$-image Pets cohorts.} Group cosine, norm balance, and symmetric angle-versus-balance contributions to the equal-weight max-corruption background shift ($-0.2998$ splits into $-0.2495$ angle and $-0.0503$ balance).}
  \label{fig:s8_geom}
\end{figure}

\begin{figure}[p]
  \centering
  \includegraphics[width=0.87\textwidth]{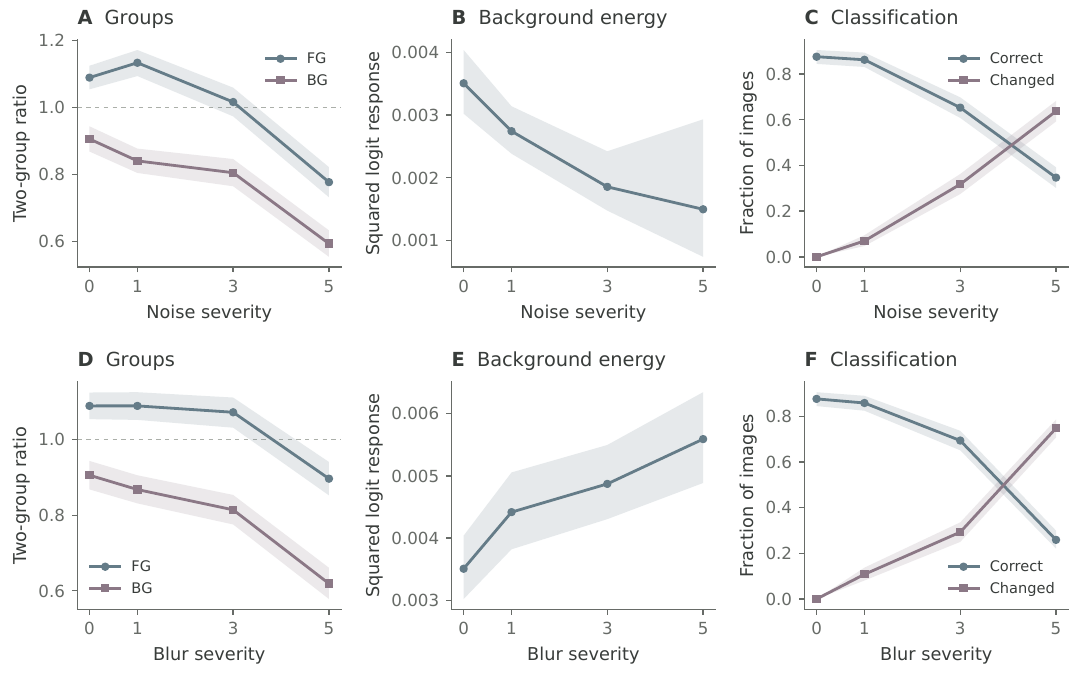}\\[0.35em]
  \includegraphics[width=0.87\textwidth]{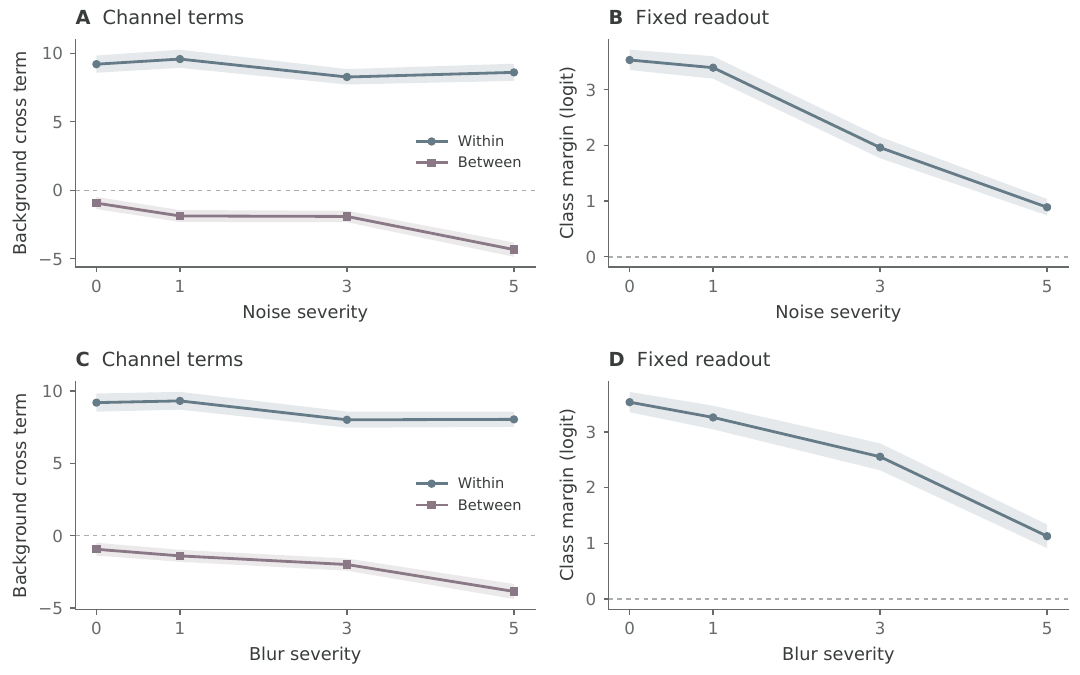}
  \caption{\textbf{OOD corruption severity curves (top) and fine-scale cross-terms with fixed clean margins (bottom).} Top: $444$-image Pets severity curves ($0, 1, 3, 5$) for two-group ratios, coherent background energy (decreasing under noise, rising under blur), and accuracy/flip rate. Bottom: $512$-channel within/between terms and fixed clean margins under noise and blur.}
  \label{fig:s8_all}
\end{figure}

We evaluated Oxford-IIIT Pets ResNet-18 across seven operating points---a shared clean baseline plus Gaussian noise and Gaussian blur at severities $1, 3,$ and $5$---holding the clean top-1 class, head-weight sign partition, and four donor edits per source image fixed (\cref{tab:s8_summary}, \cref{fig:s8_geom,fig:s8_all}). For transparent archival chronology, an initial heuristic expectation anticipated that corruption would weaken between-group background cancellation (raising $R_{\mathrm{bg}}^{(2)}$); both independent confirmation cohorts ($128$ and $444$ images) instead produced a highly consistent \emph{decrease} in $R_{\mathrm{bg}}^{(2)}$ under both noise and blur, prompting the post-hoc symmetric product decomposition $R^{(2)} = 1 + bc$ (\cref{eq:r_eq_1_plus_bc}). Under highest-severity noise (\texttt{noise5}), both relative $R_{\mathrm{bg}}^{(2)}$ and absolute coherent background energy decrease ($0.00351 \to 0.00150$); under highest-severity blur (\texttt{blur5}), relative $R_{\mathrm{bg}}^{(2)}$ drops ($0.906 \to 0.619$) even while absolute coherent background energy \emph{rises} ($0.00351 \to 0.00559$) and top-1 accuracy collapses ($87.6\% \to 25.9\%$). The geometric decomposition localizes the $-0.300$ shift at the fixed readout ($83\%$ angular opposition, $17\%$ norm equalization); upstream causal mechanisms inside earlier convolutional blocks remain open.

\begin{table}[htbp]
\centering
\caption{\textbf{Pets ResNet-18 corruption summary ($444$ confirmation images, seven operating points) and post-hoc $R=1+bc$ decomposition.}}
\label{tab:s8_summary}
\scriptsize
\setlength{\tabcolsep}{4pt}
\begin{tabular}{@{}l r r r r r r@{}}
\toprule
\textbf{Condition} & \textbf{Top-1 acc.} & \textbf{Class flip} & \textbf{FG $R^{(2)}$} & \textbf{BG $R^{(2)}$} & \textbf{BG $\Delta R^{(2)}$ vs.\ clean} & \textbf{BG coh.\ energy} \\
\midrule
Clean & $87.6\%$ & $0.0\%$ & $1.089\,[1.054, 1.125]$ & $0.906\,[0.868, 0.944]$ & $0.000$ & $0.00351$ \\
Noise level~5 ($\sigma=0.15$) & $34.7\%$ & $63.7\%$ & $0.777\,[0.732, 0.822]$ & $0.593\,[0.554, 0.633]$ & $-0.313\,[-0.365, -0.263]$ & $0.00150$ \\
Blur level~5 ($\sigma=3.2$ px) & $25.9\%$ & $74.8\%$ & $0.897\,[0.858, 0.935]$ & $0.619\,[0.578, 0.661]$ & $-0.287\,[-0.343, -0.232]$ & $0.00559$ \\
\midrule
\multicolumn{7}{@{}p{\dimexpr\textwidth-2\tabcolsep\relax}@{}}{%
\textbf{Equal-weight max-corruption BG $R = 1 + bc$ decomposition:}
In the $444$-image cohort, total shift $-0.2998\,[-0.347, -0.254]$ splits into angle $-0.2495\,[-0.2975, -0.2037]$ ($83.2\%$, $c$: $-0.140 \to -0.452$) plus norm balance $-0.0503\,[-0.0632, -0.0379]$ ($16.8\%$, $b$: $0.676 \to 0.829$).
In the $128$-image cohort, total shift $-0.2959\,[-0.378, -0.213]$ splits into angle $-0.2527$ plus balance $-0.0432$.} \\
\bottomrule
\end{tabular}
\end{table}

\FloatBarrier
\section{Source-Coordinate Inversion Observation Boundaries}
\label{supp:s9}

\begin{figure}[htbp]
  \centering
  \includegraphics[width=0.92\textwidth]{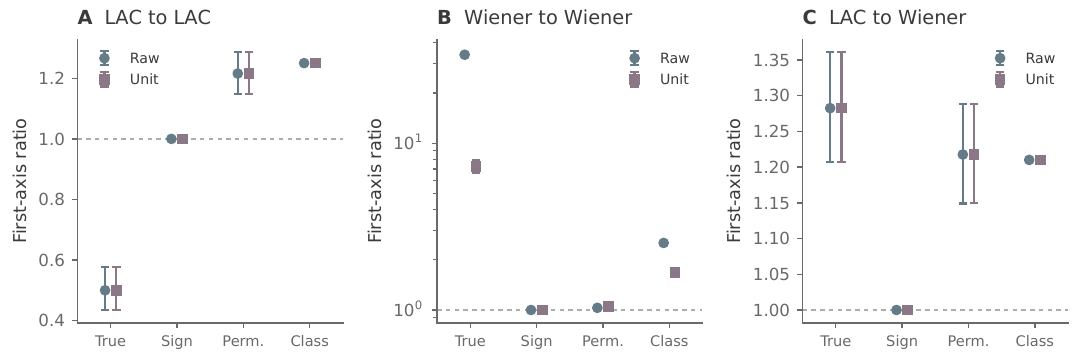}
  \caption{\textbf{Cross-operator first-shared-axis controls ($444$ Pets images).} Raw- and unit-amplitude first-shared-axis interference ratios for $\text{LAC}\to\text{LAC}$, $\text{Wiener}\to\text{Wiener}$, and cross-projected $\text{LAC}\to\text{Wiener}$ against sign-flip, weight-permutation, and wrong-class nulls.}
  \label{fig:s9_axis}
\end{figure}

\begin{figure}[p]
  \centering
  \includegraphics[width=0.86\textwidth]{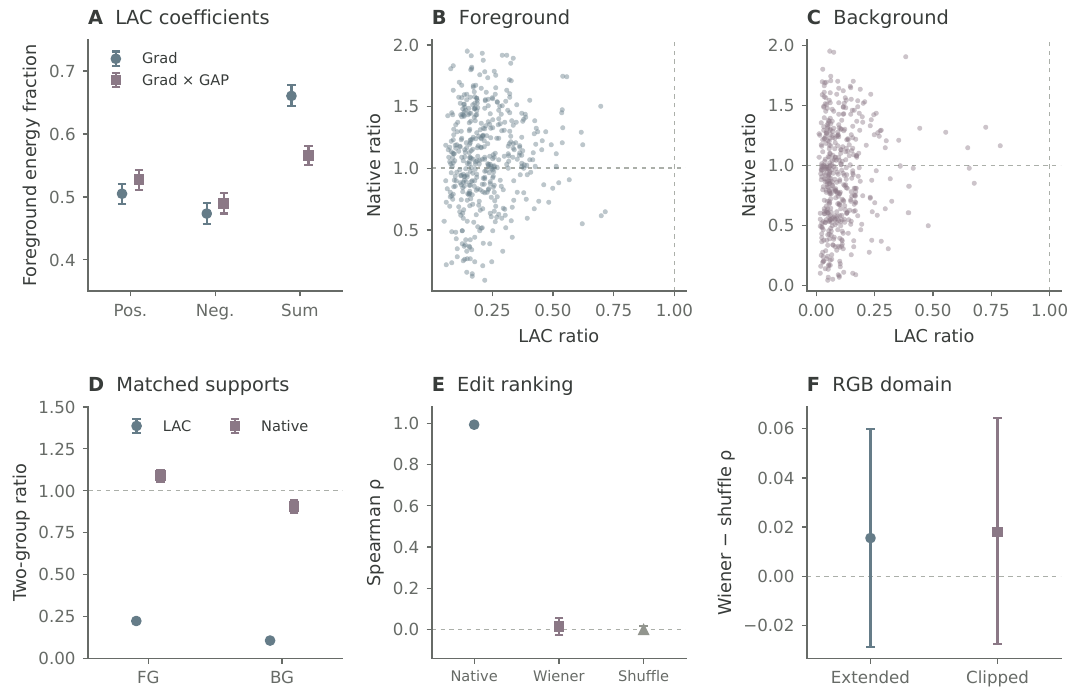}\\[0.35em]
  \includegraphics[width=0.86\textwidth]{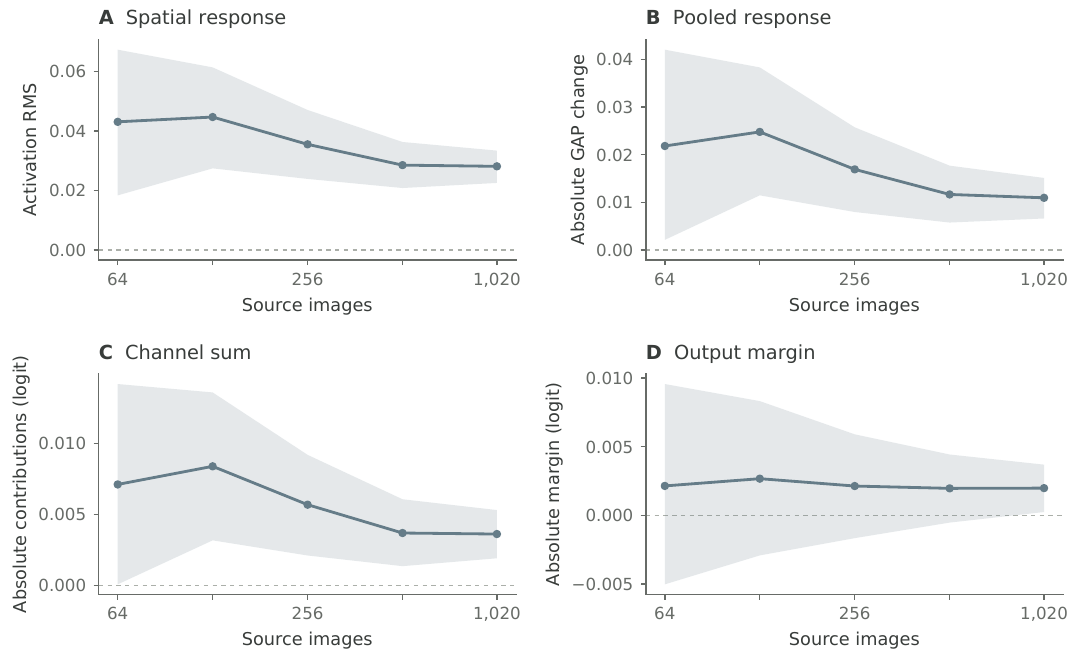}
  \caption{\textbf{Source-coordinate inversion boundaries (top) and CUB four-stage native cascade (bottom).} Top: Pets $444$-image LAC coefficient switch, LAC-versus-native scatter, support comparison, and signed pixel-edit ranking (native $\Psi$ vs.\ whole-field Wiener). Bottom: four-stage native response cascade of de-shared channels (Eq.~15 of \citet{shu2026source}) across nested CUB prefixes up to $1{,}020$ eligible images.}
  \label{fig:s9_main}
\end{figure}

\begin{figure}[p]
  \centering
  \includegraphics[width=0.77\textwidth]{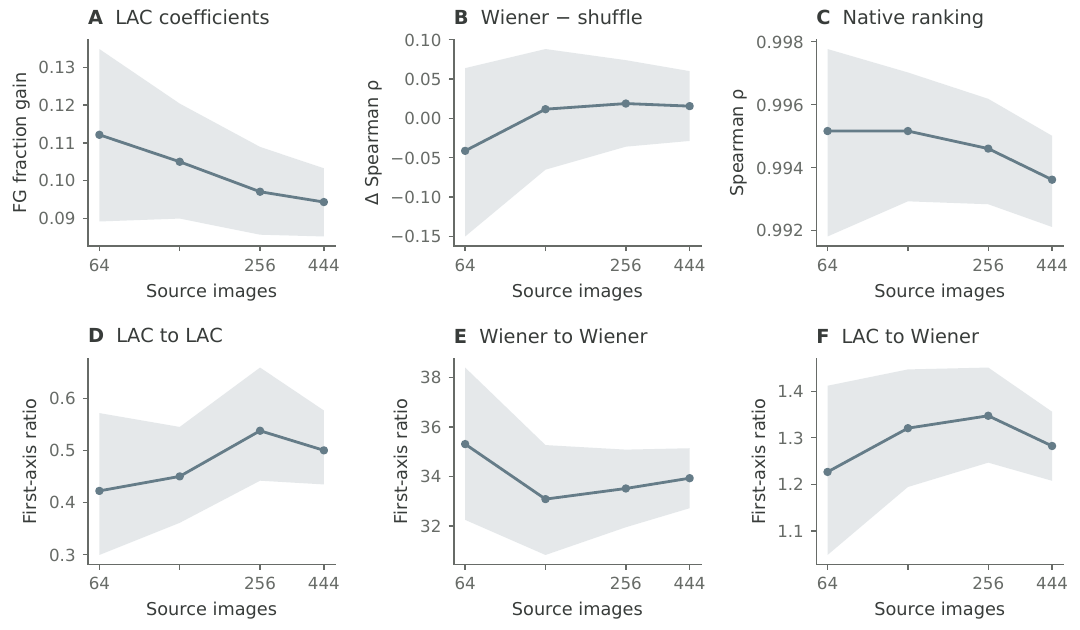}\\[0.35em]
  \includegraphics[width=0.77\textwidth]{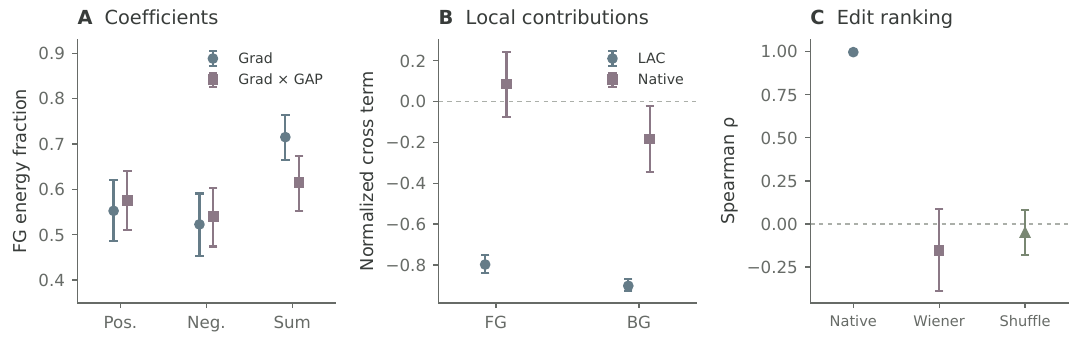}\\[0.35em]
  \includegraphics[width=0.77\textwidth]{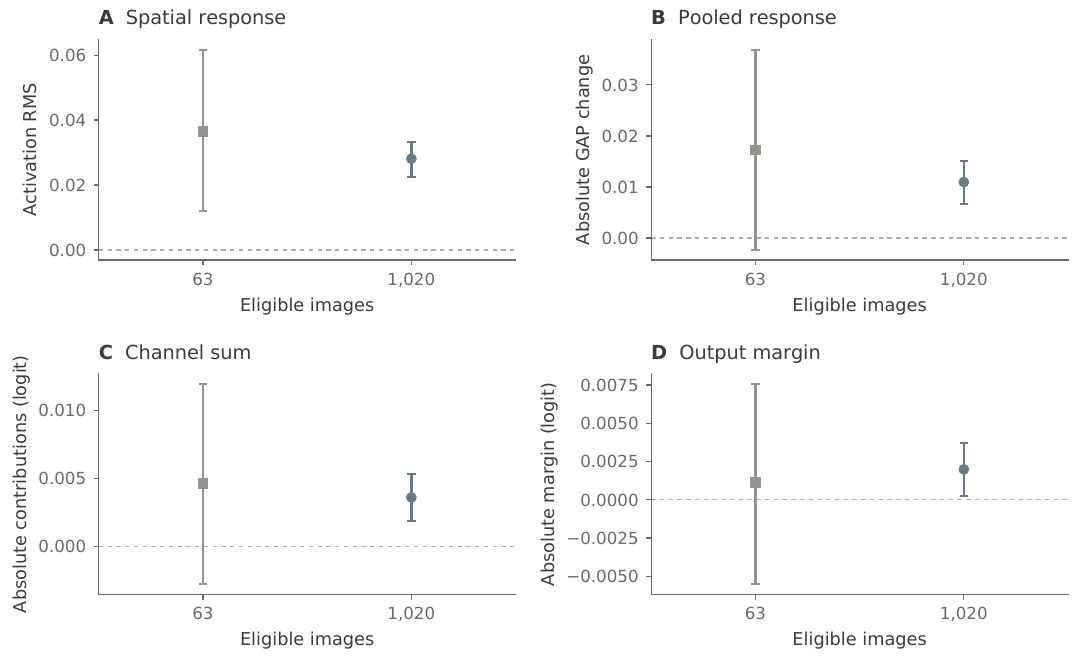}
  \caption{\textbf{Inversion sample-size stability and cohort comparisons.} Top: nested Pets prefixes ($64$ to $444$ images) for LAC shift, Wiener-minus-shuffle ranking, native $\Psi$ ranking, and raw first-axis ratios. Middle: original $16$-source Pets exploratory inversion measurements. Bottom: original $63$-source versus final $1{,}020$-source CUB regional sensitivity across all four native endpoints.}
  \label{fig:s9_extra}
\end{figure}

\Cref{fig:s9_axis,fig:s9_main,fig:s9_extra} archive all empirical boundary tests backing \cref{ax:inversion_boundary}, \cref{thm:commutation}, and \cref{sec:disc_inversion}:
\begin{enumerate}
  \item \textbf{LAC coefficient sensitivity and weak native correspondence ($444$ Pets images):} Changing LAC coefficients from $\nabla_z \odot z$ to $\nabla_z$ increases full-map foreground energy share from $56.62\%$ to $66.06\%$ (paired $+9.43\,[8.52, 10.33]$~pp), yet per-image rank correlations between LAC and native $R^{(2)}$ remain weak ($0.114$ foreground; $0.088$ background, CI crossing zero).
  \item \textbf{Shared-axis de-shared residual four-stage cascade (Eq.~15 of \citet{shu2026source}; $1{,}020$ eligible CUB images):} Residual-selected channels outperform strength-matched random controls at all four native endpoints: $7\times 7$ spatial RMS $+0.02812$, GAP response $+0.01096$, unsigned channel weight sum $+0.00362$, and matched-minus-mismatched absolute margin-restoration magnitude (signed sum within each group followed by the absolute value of the restored margin change, minus strength-matched random controls: $+0.00199\,[0.00026, 0.00373]$, while the de-shared versus full-map difference $+0.00122\,[-0.00102, +0.00345]$ remains unresolved). Calibrating signed group predictions on the $32$-image development split (frozen projection slope $0.03187$) yields a mean per-image group-level signed Pearson correlation of $+0.0485\,[0.0114, 0.0859]$ across the $1{,}020$ held-out images.
  \item \textbf{Theorem~\ref{thm:commutation} commutation defect ($444$ Pets images):} Native covariance-weighted gradients $\Psi = \Sigma_X \nabla_x m_q$ rank eight signed pixel edits at Spearman $0.9936$, whereas whole-field Wiener achieves $0.0139\,[-0.0272, +0.0556]$ (Wiener minus shuffled $+0.0155\,[-0.0287, +0.0600]$, mean normalized commutation defect $\|\Dcomm\|_2/\|\Psi\|_2 = 1.0001\,[0.9992, 1.0011]$).
  \item \textbf{Cross-operator first-shared-axis boundary (archived H-12 check):} On $444$ Pets images, $\text{LAC} \to \text{LAC}$ yields $R_b = 0.500\,[0.435, 0.578]$ (passing sign-flip, permutation, and wrong-class nulls), whereas $\text{Wiener} \to \text{Wiener}$ is constructive ($33.93$ in Pets; $25.58$ in CUB), and cross-projecting LAC weights onto the same-image Wiener first axis yields $R_b = 1.282\,[1.207, 1.361]$ despite strong unsigned axis cosine ($0.585$ versus $0.102$ mismatched).
\end{enumerate}

%% file: references.bib
@article{shu2026adjoint,
  author = {Shu, Kaixiang},
  title = {Adjoint Inversion Reveals Holographic Superposition and Destructive Interference in {CNN} Classifiers},
  journal = {arXiv preprint arXiv:2604.27529},
  year = {2026}
}

@article{shu2026source,
  author = {Shu, Kaixiang},
  title = {From Preimage Search to Source-Grounded Feature Inversion},
  journal = {arXiv preprint arXiv:2607.12526},
  year = {2026}
}

@article{shu2026pruning,
  author = {Shu, Kaixiang},
  title = {Unified Response Geometry for Structured Pruning},
  journal = {arXiv preprint arXiv:2609.18239v2},
  year = {2026}
}

@misc{shu2026shared,
  author = {Shu, Kaixiang},
  title = {Shared Relations, Specific Choices: Conditional Sensitivity and Competing Pathways in Language Models},
  howpublished = {Preprint manuscript},
  year = {2026}
}

@article{rumelhart1986learning,
  author = {Rumelhart, David E. and Hinton, Geoffrey E. and Williams, Ronald J.},
  title = {Learning Representations by Back-Propagating Errors},
  journal = {Nature},
  volume = {323},
  number = {6088},
  pages = {533--536},
  year = {1986}
}

@article{elhage2022superposition,
  author = {Elhage, Nelson and Hume, Tristan and Olsson, Catherine and Schiefer, Nicholas and Henighan, Tom and Kravec, Shauna and Hatfield-Dodds, Zac and Lasenby, Robert and Drain, Dawn and Chen, Carol and Grosse, Roger and McCandlish, Sam and Kaplan, Jared and Amodei, Dario and Wattenberg, Martin and Olah, Christopher},
  title = {Toy Models of Superposition},
  journal = {Transformer Circuits Thread},
  year = {2022}
}

@inproceedings{park2024linear,
  author = {Park, Kiho and Choe, Yo Joong and Veitch, Victor},
  title = {The Linear Representation Hypothesis and the Geometry of Large Language Models},
  booktitle = {Proceedings of the 41st International Conference on Machine Learning},
  series = {Proceedings of Machine Learning Research},
  volume = {235},
  pages = {39643--39666},
  year = {2024}
}

@inproceedings{hernandez2024linearity,
  author = {Hernandez, Evan and Sharma, Arnab Sen and Haklay, Tal and Meng, Kevin and Wattenberg, Martin and Andreas, Jacob and Belinkov, Yonatan and Bau, David},
  title = {Linearity of Relation Decoding in Transformer Language Models},
  booktitle = {International Conference on Learning Representations},
  year = {2024}
}

@inproceedings{feng2024binding,
  author = {Feng, Jiahai and Steinhardt, Jacob},
  title = {How do Language Models Bind Entities in Context?},
  booktitle = {International Conference on Learning Representations},
  year = {2024}
}

@inproceedings{meng2022locating,
  author = {Meng, Kevin and Bau, David and Andonian, Alex and Belinkov, Yonatan},
  title = {Locating and Editing Factual Associations in {GPT}},
  booktitle = {Advances in Neural Information Processing Systems},
  volume = {35},
  pages = {17359--17372},
  year = {2022}
}

@inproceedings{wang2023ioi,
  author = {Wang, Kevin Ro and Variengien, Alexandre and Conmy, Arthur and Shlegeris, Buck and Steinhardt, Jacob},
  title = {Interpretability in the Wild: A Circuit for Indirect Object Identification in {GPT-2} Small},
  booktitle = {International Conference on Learning Representations},
  year = {2023}
}

@inproceedings{conmy2023automated,
  author = {Conmy, Arthur and Mavor-Parker, Augustine and Lynch, Aengus and Heimersheim, Stefan and Garriga-Alonso, Adri{\`a}},
  title = {Towards Automated Circuit Discovery for Mechanistic Interpretability},
  booktitle = {Advances in Neural Information Processing Systems},
  volume = {36},
  pages = {16318--16352},
  year = {2023}
}

@inproceedings{geiger2021causal,
  author = {Geiger, Atticus and Lu, Hanson and Icard, Thomas and Potts, Christopher},
  title = {Causal Abstractions of Neural Networks},
  booktitle = {Advances in Neural Information Processing Systems},
  volume = {34},
  pages = {9574--9586},
  year = {2021}
}

@inproceedings{vig2020investigating,
  author = {Vig, Jesse and Gehrmann, Sebastian and Belinkov, Yonatan and Qian, Sharon and Nevo, Daniel and Singer, Yaron and Shieber, Stuart},
  title = {Investigating Gender Bias in Language Models Using Causal Mediation Analysis},
  booktitle = {Advances in Neural Information Processing Systems},
  volume = {33},
  pages = {12388--12401},
  year = {2020}
}

@article{goldowskydill2023pathpatching,
  author = {Goldowsky-Dill, Nicholas and MacLeod, Chris and Sato, Lucas and Arora, Aryaman},
  title = {Localizing Model Behavior with Path Patching},
  journal = {arXiv preprint arXiv:2304.05969},
  year = {2023}
}

@inproceedings{zhang2024patching,
  author = {Zhang, Fred and Nanda, Neel},
  title = {Towards Best Practices of Activation Patching in Language Models: Metrics and Methods},
  booktitle = {International Conference on Learning Representations},
  year = {2024}
}

@article{mcgrath2023hydra,
  author = {McGrath, Thomas and Rahtz, Matthew and Kramar, Janos and Mikulik, Vladimir and Legg, Shane},
  title = {The Hydra Effect: Emergent Self-Repair in Language Model Computations},
  journal = {arXiv preprint arXiv:2307.15771},
  year = {2023}
}

@inproceedings{mcdougall2024copy,
  author = {McDougall, Callum and Conmy, Arthur and Rushing, Cody and McGrath, Thomas and Nanda, Neel},
  title = {Copy Suppression: Comprehensively Understanding a Motif in Language Model Attention Heads},
  booktitle = {Proceedings of the 7th BlackboxNLP Workshop: Analyzing and Interpreting Neural Networks for NLP},
  pages = {337--357},
  year = {2024}
}

@inproceedings{geva2023dissecting,
  author = {Geva, Mor and Bastings, Jasmijn and Filippova, Katja and Globerson, Amir},
  title = {Dissecting Recall of Factual Associations in Auto-Regressive Language Models},
  booktitle = {Proceedings of the 2023 Conference on Empirical Methods in Natural Language Processing},
  pages = {12216--12235},
  year = {2023}
}

@article{belinkov2022probing,
  author = {Belinkov, Yonatan},
  title = {Probing Classifiers: Promises, Shortcomings, and Advances},
  journal = {Computational Linguistics},
  volume = {48},
  number = {1},
  pages = {207--219},
  year = {2022}
}

@inproceedings{ravichander2021probing,
  author = {Ravichander, Abhilasha and Belinkov, Yonatan and Hovy, Eduard},
  title = {Probing the Probing Paradigm: Does Probing Accuracy Entail Task Relevance?},
  booktitle = {Proceedings of the 16th Conference of the European Chapter of the Association for Computational Linguistics},
  pages = {3363--3377},
  year = {2021}
}

@inproceedings{bau2017network,
  author = {Bau, David and Zhou, Bolei and Khosla, Aditya and Oliva, Aude and Torralba, Antonio},
  title = {Network Dissection: Quantifying Interpretability of Deep Visual Representations},
  booktitle = {Proceedings of the IEEE Conference on Computer Vision and Pattern Recognition},
  pages = {3319--3327},
  year = {2017}
}

@inproceedings{kim2018interpretability,
  author = {Kim, Been and Wattenberg, Martin and Gilmer, Justin and Cai, Carrie and Wexler, James and Vi{\'e}gas, Fernanda and Sayres, Rory},
  title = {Interpretability Beyond Feature Attribution: Quantitative Testing with Concept Activation Vectors ({TCAV})},
  booktitle = {Proceedings of the 35th International Conference on Machine Learning},
  pages = {2668--2677},
  year = {2018}
}

@inproceedings{li2023iti,
  author = {Li, Kenneth and Patel, Oam and Vi{\'e}gas, Fernanda and Pfister, Hanspeter and Wattenberg, Martin},
  title = {Inference-Time Intervention: Eliciting Truthful Answers from a Language Model},
  booktitle = {Advances in Neural Information Processing Systems},
  volume = {36},
  pages = {41451--41530},
  year = {2023}
}

@inproceedings{tan2024steering,
  author = {Tan, Daniel and Chanin, David and Lynch, Aengus and Paige, Brooks and Kanoulas, Dimitrios and Garriga-Alonso, Adri{\`a} and Kirk, Robert},
  title = {Analysing the Generalisation and Reliability of Steering Vectors},
  booktitle = {Advances in Neural Information Processing Systems},
  volume = {37},
  pages = {139179--139212},
  year = {2024}
}

@article{turner2023activation,
  author = {Turner, Alexander Matt and Thiergart, Lisa and Leech, Gavin and Udell, David and Vazquez, Juan J. and Mini, Ulisse and MacDiarmid, Monte},
  title = {Steering Language Models With Activation Engineering},
  journal = {arXiv preprint arXiv:2308.10248},
  year = {2023}
}

@article{zou2023representation,
  author = {Zou, Andy and Phan, Long and Chen, Sarah and Campbell, James and Guo, Phillip and Ren, Richard and Pan, Alexander and Yin, Xuwang and Mazeika, Mantas and Dombrowski, Ann-Kathrin and Goel, Shashwat and Li, Nathaniel and Byun, Michael J. and Wang, Zifan and Mallen, Alex and Basart, Steven and Koyejo, Sanmi and Song, Dawn and Fredrikson, Matt and Kolter, J. Zico and Hendrycks, Dan},
  title = {Representation Engineering: A Top-Down Approach to {AI} Transparency},
  journal = {arXiv preprint arXiv:2310.01405},
  year = {2023}
}

@article{braun2025unreliability,
  author = {Braun, Joschka and Eickhoff, Carsten and Krueger, David and Bahrainian, Seyed Ali and Krasheninnikov, Dmitrii},
  title = {Understanding (Un)Reliability of Steering Vectors in Language Models},
  journal = {arXiv preprint arXiv:2505.22637},
  year = {2025}
}

@article{torop2026inverted,
  author = {Torop, Max and Masoomi, Aria and Dy, Jennifer},
  title = {Inverted Detection and Control in Steering Vectors},
  journal = {arXiv preprint arXiv:2608.02957},
  year = {2026}
}

@article{li2026vectorfields,
  author = {Li, Jiaqian and Li, Yanshu and Huang, Kuan-Hao},
  title = {Steering Vector Fields for Context-Aware Inference-Time Control in Large Language Models},
  journal = {arXiv preprint arXiv:2602.01654},
  year = {2026}
}

@article{gong2026coablation,
  author = {Gong, Zhiren and Lu, He and Wang, Tiantong and Zhang, Yichi and Wang, Yixin and Zeng, Zihao and Xiao, Ming and Yuen, Chau and Lim, Wei Yang Bryan},
  title = {Conditional Co-Ablation: Recovering Self-Repair Backups in Transformer Circuits},
  journal = {arXiv preprint arXiv:2607.01940},
  year = {2026}
}

@article{simonyan2013deep,
  author = {Simonyan, Karen and Vedaldi, Andrea and Zisserman, Andrew},
  title = {Deep Inside Convolutional Networks: Visualising Image Classification Models and Saliency Maps},
  journal = {arXiv preprint arXiv:1312.6034},
  year = {2013}
}

@inproceedings{zeiler2014visualizing,
  author = {Zeiler, Matthew D. and Fergus, Rob},
  title = {Visualizing and Understanding Convolutional Networks},
  booktitle = {European Conference on Computer Vision},
  pages = {818--833},
  year = {2014}
}

@inproceedings{mahendran2015understanding,
  author = {Mahendran, Aravindh and Vedaldi, Andrea},
  title = {Understanding Deep Image Representations by Inverting Them},
  booktitle = {Proceedings of the IEEE Conference on Computer Vision and Pattern Recognition},
  pages = {5188--5196},
  year = {2015}
}

@inproceedings{dosovitskiy2016inverting,
  author = {Dosovitskiy, Alexey and Brox, Thomas},
  title = {Inverting Visual Representations with Convolutional Networks},
  booktitle = {Proceedings of the IEEE Conference on Computer Vision and Pattern Recognition},
  pages = {4829--4837},
  year = {2016}
}

@article{olah2017feature,
  author = {Olah, Chris and Mordvintsev, Alexander and Schubert, Ludwig},
  title = {Feature Visualization},
  journal = {Distill},
  volume = {2},
  number = {11},
  pages = {e7},
  year = {2017}
}

@inproceedings{selvaraju2017grad,
  author = {Selvaraju, Ramprasaath R. and Cogswell, Michael and Das, Abhishek and Vedantam, Ramakrishna and Parikh, Devi and Batra, Dhruv},
  title = {{Grad-CAM}: Visual Explanations from Deep Networks via Gradient-Based Localization},
  booktitle = {Proceedings of the IEEE International Conference on Computer Vision},
  pages = {618--626},
  year = {2017}
}

@inproceedings{sundararajan2017axiomatic,
  author = {Sundararajan, Mukund and Taly, Ankur and Yan, Qiqi},
  title = {Axiomatic Attribution for Deep Networks},
  booktitle = {Proceedings of the 34th International Conference on Machine Learning},
  volume = {70},
  pages = {3319--3328},
  year = {2017}
}

@inproceedings{adebayo2018sanity,
  author = {Adebayo, Julius and Gilmer, Justin and Muelly, Michael and Goodfellow, Ian and Hardt, Moritz and Kim, Been},
  title = {Sanity Checks for Saliency Maps},
  booktitle = {Advances in Neural Information Processing Systems},
  volume = {31},
  year = {2018}
}

@inproceedings{ancona2018towards,
  author = {Ancona, Marco and Ceolini, Enea and {\"O}ztireli, Cengiz and Gross, Markus},
  title = {Towards Better Understanding of Gradient-Based Attribution Methods for Deep Neural Networks},
  booktitle = {International Conference on Learning Representations},
  year = {2018}
}

@inproceedings{nash2019inverting,
  author = {Nash, Charlie and Kushman, Nate and Williams, Christopher K. I.},
  title = {Inverting Supervised Representations with Autoregressive Neural Density Models},
  booktitle = {Proceedings of the 22nd International Conference on Artificial Intelligence and Statistics},
  pages = {1620--1629},
  year = {2019}
}

@article{haufe2014interpretation,
  author = {Haufe, Stefan and Meinecke, Frank and G{\"o}rgen, Kai and D{\"a}hne, Sven and Haynes, John-Dylan and Blankertz, Benjamin and Bie{\ss}mann, Felix},
  title = {On the Interpretation of Weight Vectors of Linear Models in Multivariate Neuroimaging},
  journal = {NeuroImage},
  volume = {87},
  pages = {96--110},
  year = {2014}
}

@book{wiener1949extrapolation,
  author = {Wiener, Norbert},
  title = {Extrapolation, Interpolation, and Smoothing of Stationary Time Series: With Engineering Applications},
  publisher = {MIT Press},
  year = {1949}
}

@inproceedings{jain2019attention,
  author = {Jain, Sarthak and Wallace, Byron C.},
  title = {Attention is not Explanation},
  booktitle = {Proceedings of the 2019 Conference of the North American Chapter of the Association for Computational Linguistics: Human Language Technologies},
  pages = {3543--3556},
  year = {2019}
}

@inproceedings{wiegreffe2019attention,
  author = {Wiegreffe, Sarah and Pinter, Yuval},
  title = {Attention is not not Explanation},
  booktitle = {Proceedings of the 2019 Conference on Empirical Methods in Natural Language Processing},
  pages = {11--20},
  year = {2019}
}

@inproceedings{morris2023text,
  author = {Morris, John and Kuleshov, Volodymyr and Shmatikov, Vitaly and Rush, Alexander},
  title = {Text Embeddings Reveal (Almost) As Much As Text},
  booktitle = {Proceedings of the 2023 Conference on Empirical Methods in Natural Language Processing},
  pages = {12448--12460},
  year = {2023}
}

@inproceedings{xiao2021noise,
  author = {Xiao, Kai and Engstrom, Logan and Ilyas, Andrew and Madry, Aleksander},
  title = {Noise or Signal: The Role of Image Backgrounds in Object Recognition},
  booktitle = {International Conference on Learning Representations},
  year = {2021}
}

@article{geirhos2020shortcut,
  author = {Geirhos, Robert and Jacobsen, J{\"o}rn-Henrik and Michaelis, Claudio and Zemel, Richard and Brendel, Wieland and Bethge, Matthias and Wichmann, Felix A.},
  title = {Shortcut Learning in Deep Neural Networks},
  journal = {Nature Machine Intelligence},
  volume = {2},
  number = {11},
  pages = {665--673},
  year = {2020}
}

@inproceedings{lecun1989optimal,
  author = {LeCun, Yann and Denker, John S. and Solla, Sara A.},
  title = {Optimal Brain Damage},
  booktitle = {Advances in Neural Information Processing Systems},
  volume = {2},
  pages = {598--605},
  year = {1989}
}

@inproceedings{hassibi1993optimal,
  author = {Hassibi, Babak and Stork, David G.},
  title = {Second Order Derivatives for Network Pruning: Optimal Brain Surgeon},
  booktitle = {Advances in Neural Information Processing Systems},
  volume = {5},
  pages = {164--171},
  year = {1993}
}

@inproceedings{lee2021layer,
  author = {Lee, Jaeho and Park, Sejun and Mo, Sangwoo and Ahn, Sungsoo and Shin, Jinwoo},
  title = {Layer-Adaptive Sparsity for the Magnitude-Based Pruning},
  booktitle = {International Conference on Learning Representations},
  year = {2021}
}

@inproceedings{frantar2023sparsegpt,
  author = {Frantar, Elias and Alistarh, Dan},
  title = {{SparseGPT}: Massive Language Models Can Be Accurately Pruned in One-Shot},
  booktitle = {Proceedings of the 40th International Conference on Machine Learning},
  volume = {202},
  pages = {10323--10337},
  year = {2023}
}

@inproceedings{sun2024simple,
  author = {Sun, Mingjie and Liu, Zhuang and Bair, Anna and Kolter, J. Zico},
  title = {A Simple and Effective Pruning Approach for Large Language Models},
  booktitle = {International Conference on Learning Representations},
  year = {2024}
}

@article{frisch1933partial,
  author = {Frisch, Ragnar and Waugh, Frederick V.},
  title = {Partial Time Regressions as Compared with Individual Trends},
  journal = {Econometrica},
  volume = {1},
  number = {4},
  pages = {387--401},
  year = {1933}
}

@article{lovell1963seasonal,
  author = {Lovell, Michael C.},
  title = {Seasonal Adjustment of Economic Time Series and Multiple Regression Analysis},
  journal = {Journal of the American Statistical Association},
  volume = {58},
  number = {304},
  pages = {993--1010},
  year = {1963}
}

@article{hutchinson1989stochastic,
  author = {Hutchinson, Michael F.},
  title = {A Stochastic Estimator of the Trace of the Influence Matrix for Laplacian Smoothing Splines},
  journal = {Communications in Statistics--Simulation and Computation},
  volume = {18},
  number = {3},
  pages = {1059--1076},
  year = {1989}
}

@article{qwen2024technical,
  author = {{Qwen Team}},
  title = {{Qwen2.5} Technical Report},
  journal = {arXiv preprint arXiv:2412.15115},
  year = {2024}
}

@misc{meta2024llama32,
  author = {{Meta}},
  title = {{Llama 3.2} Model Card},
  year = {2024},
  url = {https://github.com/meta-llama/llama-models/blob/main/models/llama3_2/MODEL_CARD.md}
}

@inproceedings{deng2009imagenet,
  author = {Deng, Jia and Dong, Wei and Socher, Richard and Li, Li-Jia and Li, Kai and Fei-Fei, Li},
  title = {{ImageNet}: A Large-Scale Hierarchical Image Database},
  booktitle = {Proceedings of the IEEE Conference on Computer Vision and Pattern Recognition},
  pages = {248--255},
  year = {2009}
}

@techreport{wah2011caltech,
  author = {Wah, Catherine and Branson, Steve and Welinder, Peter and Perona, Pietro and Belongie, Serge},
  title = {The {Caltech-UCSD} Birds-200-2011 Dataset},
  institution = {California Institute of Technology},
  number = {CNS-TR-2011-001},
  year = {2011}
}

@inproceedings{parkhi2012cats,
  author = {Parkhi, Omkar M. and Vedaldi, Andrea and Zisserman, Andrew and Jawahar, C. V.},
  title = {Cats and Dogs},
  booktitle = {Proceedings of the IEEE Conference on Computer Vision and Pattern Recognition},
  pages = {3498--3505},
  year = {2012}
}
